\documentclass{article}

\usepackage[T1]{fontenc}
\usepackage{microtype}
\usepackage{graphicx}
\usepackage{subcaption}
\usepackage{booktabs}
\usepackage{multirow}
\usepackage{placeins}
\usepackage{enumitem}
\usepackage{tikz}
\usetikzlibrary{arrows.meta,positioning,calc,shapes.geometric}
\usepackage{pgfplots}
\pgfplotsset{compat=1.18}

\usepackage{hyperref}

\usepackage[preprint]{icml2027}

\usepackage{amsmath}
\usepackage{amssymb}
\usepackage{mathtools}
\usepackage{amsthm}
\usepackage[capitalize,noabbrev]{cleveref}
\crefname{appendix}{Appendix}{Appendices}
\Crefname{appendix}{Appendix}{Appendices}

\theoremstyle{plain}
\newtheorem{theorem}{Theorem}[section]

\theoremstyle{definition}

\theoremstyle{remark}
\newtheorem{remark}[theorem]{Remark}

\icmltitlerunning{A CPU Inside an LLM: Zero-Gradient Circuit Compilation into Pretrained Weights}

\begin{document}

\twocolumn[
  \icmltitle{A CPU Inside an LLM: Zero-Gradient Compilation of\texorpdfstring{\\}{ }Exact Circuits into Pretrained Weights}

  \begin{icmlauthorlist}
    \icmlauthor{Tanmay Sah}{inst1}
  \end{icmlauthorlist}

  \icmlaffiliation{inst1}{Independent Researcher}
  \icmlcorrespondingauthor{Tanmay Sah}{tradertanmay@gmail.com}

  \icmlkeywords{Machine Learning, Circuit Compilation, Pretrained Language Models, Mechanistic Interpretability}

  \vskip 0.3in
]

\printAffiliationsAndNotice{}
\hypersetup{pdfsubject={Preprint}}

\begin{abstract}
Pretrained language models (LMs) remain unreliable on exact multi-digit arithmetic without gradient fine-tuning or external tool interpreters. We introduce closed-form, zero-gradient circuit-grafting primitives that compile deterministic computation directly into pretrained LM weights using zero training examples and zero runtime hooks, while bounding per-layer normalization drift and preserving natural-language (NL) competence. Our construction combines three analytical mechanisms: (1)~two-layer Mellin-GLU quadrature over bias-free SwiGLU/GeGLU layers, turning unknown per-token $\operatorname{RMSNorm}$ scaling into translation along a log-spaced gate grid ($>250\times$ dynamic range) with IEEE~754 error bounds; (2)~cross-token scale cancellation via signed Chebyshev Indefinite-RoPE self-sink attention and memoryless $\operatorname{SiLU}$-antisymmetric \texttt{GatedDeltaNet} shifts; and (3)~axis-aligned width extension ($d \to d+K$) with $\sqrt{d/(d+K)}$ norm compensation, Base-1000 canonical BPE emission, and differential $(+254,-256)$ tied unembeddings. An automated 5-stage compiler synthesizes $4$--$8$ digit signed $\{+,-,\times\}$ circuits in constant $\mathsf{TC}^0$ depth alongside 3-literal Boolean ($100\%$, $N{=}1{,}728$; plus 42-layer Gemma~4~E4B audits) and 5-bit Majority/Parity ($100\%$) circuits. On 6-digit arithmetic ($N{=}1{,}200$, \texttt{BF16}), IBM Granite~4.2~3B ($1.07\text{M}$--$1.33\text{M}$ active non-zeros, $\le 0.036\%$) and Alibaba Qwen~3.5~4B ($2.25\text{M}$--$2.39\text{M}$ non-zeros, $\le 0.057\%$) reach $99.17\%$ and $100.00\%$ ALU-prefix accuracy ($100\%$ with an \texttt{FP32} sidecar) and $98\%$--$100\%$ mid-sentence resumption---outperforming $12.8\text{K}$-sample \texttt{LoRA} ($42.2\%$ / $61.7\%$, $0\%$ inline continuation). A factorized block-sparse sidecar holds A100 memory at $7.89$ / $10.65\text{ GiB}$ ($4.60$ / $8.14\text{ GiB}$ under \texttt{INT8}/\texttt{INT4} + \texttt{FP32} sidecar) while passing all $14/14$ pre-registered $>2\text{K}$-token Tango splits ($N{=}4{,}392$; $0/11{,}134$ OOD triggers).
\end{abstract}

\section{Introduction}
\label{sec:intro}

Autoregressive LMs interleave natural-language reasoning with deterministic sub-computations, yet next-token prediction remains brittle on exact multi-digit arithmetic~\citep{liu2023goat,mcleish2024transformers,zhou2024what}. Deployed systems route around this via host tool wrappers~\citep{schick2023toolformer} that pause decoding to invoke external interpreters. Conversely, RASP~\citep{weiss2021thinking} and Tracr~\citep{lindner2023tracr} compile high-level programs into scratch-built toy Transformers without normalization invariants, and weight-space virtual machines such as Percepta's CALM WASM VM~\citep{percepta2026vm} execute multi-thousand-token traces inside dedicated VMs. Embedding exact digital logic \emph{directly into general-purpose pretrained LMs without gradient optimization or host routing}---while maintaining noninferior accuracy on evaluated language benchmarks and mid-sentence continuation---poses distinct theoretical, numerical, systems, and tokenization challenges.

\begin{table*}[t]
\caption{\textbf{Comparison with prior and concurrent paradigms}, separating \emph{Reproduced} baselines on identical Granite~4.2 / Qwen~3.5 splits from \emph{Conceptual} comparisons.}
\label{tab:prior_work}
\vskip 0.02in
\begin{center}
\scriptsize
\setlength{\tabcolsep}{4.0pt}
\resizebox{\textwidth}{!}{%
\begin{tabular}{@{}llllll@{}}
\toprule
\textbf{Method / Paradigm} & \textbf{Comp.\ Type} & \textbf{Pretrained LM?} & \textbf{Synthesis \& Data Regime} & \textbf{Zero Host Route?} & \textbf{NL Retention \& Dense Cost} \\
\midrule
RASP / Tracr~\citep{weiss2021thinking,lindner2023tracr} & Conceptual & No (scratch) & Auto DSL ($0$ data; no $\operatorname{RMSNorm}$) & Yes & No NL capability \\
CoNNs / Tracr-Inj.~\citep{weng2024mastering,vergarabrowne2025tracr} & Conceptual & Yes (fused / $d$) & Compiled module + \textbf{Gradient fusion/distill.} & Yes & Trained/distilled; $\ge 1.0\times$ \\
Percepta CALM VM~\citep{percepta2026vm} & Conceptual & Specialized VM & MILP ReGLU + 2D CHT trace & Yes (long trace) & Dedicated trace VM (not general LM) \\
Agentic WSP / SWD~\citep{takehi2026sparks,yan2026sparse} & Conceptual & Yes (Qwen) & Agentic heuristic edit / circuit extraction & Yes & Heuristic edit or extraction; $1.0\times$ \\
\midrule
PAL / Python Tool Router~\citep{gao2023pal,schick2023toolformer} & \textbf{Reproduced} & Yes (wrapped) & Host regex/program + Python ($0$ weights) & \textbf{No ($100\%$ host)} & Exact NL; $1.00\times$ (requires host) \\
Same-Backbone \texttt{LoRA} SFT ($12.8\text{K}$ ex.) & \textbf{Reproduced} & Yes ($+6\text{M}$ LoRA) & $400$-step AdamW on 1--6d $\{+,-,\times\}$ & Yes ($100\%$ in-$W$) & $42$--$62\%$ Arith; \textbf{$0\%$ inline prose} \\
Fixed-$d$ SVD ($P_\perp$) / Matched-$(d{+}K)$ & \textbf{Reproduced} & Yes ($d$ / $d{+}K$) & Zero-grad SVD carve or 5-prim.\ leave-1-out & Yes & Fails Arith ($8$--$51\%$) or PPL/ARC \\
\textbf{Mellin Primitives + Sidecar (Ours)} & \textbf{Evaluated} & \textbf{Yes (3B--4B)} & \textbf{Auto Compiler ($0$ SGD; $4$--$8$d ALU + $\mathsf{TC}^0$)} & \textbf{Yes ($100\%$ in-$W$)} & \textbf{Tango NI + Inline ($1.16\times$ sidecar / $0.68\times$ INT4)} \\
\bottomrule
\end{tabular}%
}
\end{center}
\vskip -0.14in
\end{table*}

\textbf{Scope and Contributions.}
Building a 5-stage declarative compiler atop reusable analytical primitives, we contribute:
\begin{enumerate}[leftmargin=1.3em,itemsep=0.5pt,topsep=1pt]
\item \textbf{Scale- and Architecture-Invariant Primitives (\cref{sec:theory}, \cref{app:proofs}):} Single-layer impossibility for bias-free GLU steps plus payload-linear truncation and kernel-explicit IEEE~754 / \texttt{TF32} bounds for Mellin-GLU quadrature (\cref{thm:mellin}), Indefinite-RoPE GQA and memoryless \texttt{GatedDeltaNet} shifts (\cref{thm:routing}), and axis-aligned width extension with arbitrary-tree tied-GEMM bounds (\cref{thm:subspace}).
\item \textbf{Automated Multi-Digit ALU, $\mathsf{TC}^0$ DAGs, \& Sparse Sidecar (\cref{sec:arch}):} Automated $4$--$8$ digit BPE $\{+, -, \times\}$ ($1.07\text{M}$--$2.25\text{M}$ nnz) and reference 6-digit patches in \texttt{granite-4.2-3b} and \texttt{Qwen3.5-4B} ($0/8$ Attn), a \textbf{factorized block-sparse sidecar} (cutting A100 VRAM $42\%$--$44\%$ and pairing \texttt{INT8}/\texttt{INT4} $W_{\mathrm{NL}}$ with an \texttt{FP32} sidecar; \cref{tab:systems_footprint}), declarative Boolean ($100\%$, $N{=}1{,}728$) \& 5-bit Majority/Parity, triad BPE alignment, and 42L Gemma~4~E4B audits.
\item \textbf{Decomposed, Ablation, Kernel, \& Retention Audits (\cref{sec:experiments}):} Five-primitive leave-one-out ALU ablations, same-backbone \texttt{LoRA} SFT comparison, \texttt{TorchInductor}/\texttt{FlashAttention-2}/\texttt{FP16} kernel portability, 5 decoding regimes + \texttt{IFEval}/safety, $14/14$ pre-registered $>2\text{K}$-token Tango passes ($N{=}4{,}392$), $0/11{,}134$ OOD triggers, and \texttt{mps}/A100-\texttt{cuda} reproducibility (\cref{app:repro}).
\end{enumerate}

\section{Related Work}
\label{sec:related}

\begin{figure*}[t]
\vskip 0.01in
\begin{center}
\resizebox{0.97\textwidth}{!}{%
\begin{tikzpicture}
  \begin{axis}[
    name=intA,
    width=0.465\textwidth,
    height=2.85cm,
    xmode=log,
    xmin=0.02, xmax=5.0,
    ymin=-0.08, ymax=1.78,
    xtick={0.02,0.1,0.5,1,5},
    xticklabels={$0.02$,$0.1$,$0.5$,$1$,$5$},
    ytick={0,0.5,1.0,1.5},
    xlabel={\footnotesize \textbf{(a)} Unknown $\operatorname{RMSNorm}$ Factor $\lambda_{t,\ell}$ ($250{\times}$ swing)},
    ylabel={\footnotesize Effective Gain $\hat{y}/(x/r_0)$},
    tick label style={font=\scriptsize},
    grid=both,
    grid style={gray!20},
    legend style={
      font=\scriptsize,
      at={(0.5,1.06)},
      anchor=south,
      legend columns=2,
      fill=white,
      draw=gray!45,
      inner sep=1.5pt,
      column sep=4pt
    }
  ]
    \addplot[color=red!78!black, dashed, very thick] coordinates {
      (0.02,0.0007) (0.035,0.0022) (0.06,0.0065) (0.1,0.0177) (0.18,0.0548)
      (0.3,0.1424) (0.5,0.3510) (0.75,0.6710) (1.0,1.0000) (1.4,1.4417)
      (2.0,1.6500) (3.0,1.5871) (4.0,1.0700) (5.0,0.6221)
    };
    \addlegendentry{Single-Scale (\texttt{M2})}

    \addplot[color=blue!78!black, solid, very thick, mark=*, mark size=1.3pt] coordinates {
      (0.02,1.0000) (0.06,1.0000) (0.18,1.0000) (0.5,1.0000) (1.0,1.0000) (2.0,1.0000) (5.0,0.9999)
    };
    \addlegendentry{Mellin ($\Delta{=}0.55$)}

    \node[anchor=south, font=\scriptsize, text=blue!75!black, fill=white, fill opacity=0.88, text opacity=1, inner sep=1.0pt] at (axis cs:0.15,1.12) {$k\Delta \mapsto k\Delta + \ln\lambda$};
  \end{axis}

  \begin{axis}[
    name=intB,
    at={(intA.east)},
    anchor=west,
    xshift=1.85cm,
    width=0.495\textwidth,
    height=2.85cm,
    xmin=-0.3, xmax=9.3,
    ymin=-9.8, ymax=1.65,
    xtick={0,1,2,3,4,5,6,7,8,9},
    xticklabels={$0$,$1$,$2$,$3$,$4$,$8$,$16$,$64$,$256$,$1024$},
    ytick={-8,-4,0,1},
    xlabel={\footnotesize \textbf{(b)} Relative Offset $|\delta| = |t - d_0 - u|$ ($\theta_{\mathrm{RoPE}}{=}10^7$)},
    ylabel={\footnotesize Kernel $S(\delta)$},
    tick label style={font=\scriptsize},
    grid=both,
    grid style={gray!20},
    legend style={
      font=\scriptsize,
      at={(0.5,1.06)},
      anchor=south,
      legend columns=2,
      fill=white,
      draw=gray!45,
      inner sep=1.5pt,
      column sep=4pt
    }
  ]
    \fill[teal!10] (axis cs:0.75,-9.8) rectangle (axis cs:9.3,-0.3546);
    \draw[densely dotted, black!70, thick] (axis cs:-0.3,-0.3546) -- (axis cs:9.3,-0.3546);

    \addplot[color=orange!85!black, dashed, very thick, mark=triangle*, mark size=1.5pt] coordinates {
      (0,1.0000) (1,0.9768) (2,0.9229) (3,0.8716) (4,0.8476) (5,0.8116) (6,0.7292) (7,0.6978) (8,0.6320) (9,0.6125)
    };
    \addlegendentry{Std.\ RoPE ($w_m{>}0$)}

    \addplot[color=teal!78!black, solid, very thick, mark=*, mark size=1.4pt] coordinates {
      (0,1.0000) (1,-0.3546) (2,-3.2295) (3,-5.1356) (4,-4.5290) (5,-3.2548) (6,-6.3054) (7,-2.5117) (8,-4.1505) (9,-1.1608)
    };
    \addlegendentry{Indef-RoPE ($\le{-}0.355$)}

    \node[anchor=south, font=\scriptsize, text=teal!72!black, inner sep=0.8pt] at (axis cs:5.0,-9.55) {Off-target $\le -85\text{ nats}$};
  \end{axis}
\end{tikzpicture}%
}
\caption{\textbf{Geometric intuition for normalization cancellation and self-sink routing.} \textbf{(a)}~When $\operatorname{RMSNorm}$ rescales $(r_0, x) \mapsto (\lambda r_0, \lambda x)$ across $250\times$ swings, single-scale SwiGLU collapses or saturates, whereas a Mellin log-grid $\alpha_k = e^{k\Delta}$ shifts $\ln\lambda$ along a flat plateau ($\hat{y} \equiv x/r_0$). \textbf{(b)}~At $\theta_{\mathrm{RoPE}}=10^7$, positive weights leave $S(\delta) \ge +0.61$; signed Chebyshev weights ($k_m = \operatorname{sgn}(w_m)\sqrt{|w_m|}$) enforce $S(0)=+1$ and $S(\delta) \le -0.355 < 0$ ($1 \le |\delta| \le 1024$), parking inactive queries on their zero-payload self-sink ($\delta_{\mathrm{self}}=0$).}
\label{fig:intuition}
\end{center}
\vskip -0.14in
\end{figure*}
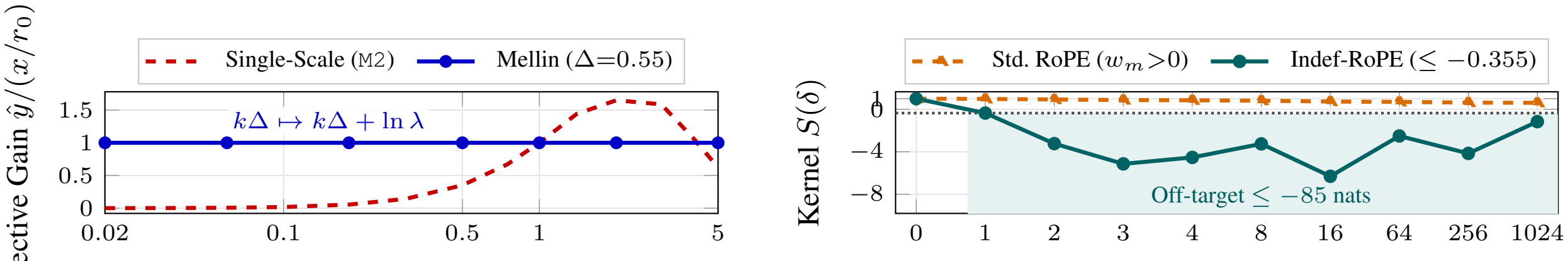

\textbf{Compiled Circuits, CoNNs, and Subspace Injection.}
RASP~\citep{weiss2021thinking} and Tracr~\citep{lindner2023tracr} compile declarative programs into scratch Transformers assuming MLP biases, absolute positions, and unnormalized wires---violated by $\operatorname{RMSNorm}$~\citep{zhang2019root}, bias-free SwiGLU/GeGLU~\citep{shazeer2020glu}, RoPE~\citep{su2024roformer}, and \texttt{GatedDeltaNet}~\citep{yang2025gated}. CoNNs~\citep{weng2024mastering} fuse external rule modules via \emph{gradient-trained} adapters, and \emph{Tracr-Injection}~\citep{vergarabrowne2025tracr} pairs subspace projection ($P_\perp = I_d - U U^\top$) with \emph{gradient distillation}; we benchmark zero-gradient $P_\perp$ and matched-$(d+K)$ controls in \cref{sec:experiments} (\cref{tab:prior_work}). Sparse Weight Decomposition~\citep{yan2026sparse} factorizes weights to \emph{extract} existing circuits (including on Qwen~3.5), complementing our forward $<0.06\%$-nonzero synthesis.

\textbf{Weight-Space VMs, Tokenization, and Tool Routers.}
Looped Transformers~\citep{giannou2023looped} and Percepta's CALM VM~\citep{percepta2026vm} execute traces in dedicated VMs rather than general pretrained LMs, while concurrent Weight-Space Programming~\citep{takehi2026sparks,hu2026exploring} heuristically perturbs $\sim 10^3$ scalars. Unlike gradient fine-tuning~\citep{liu2023goat,mcleish2024transformers} or host Python routers such as PAL~\citep{gao2023pal} and Toolformer~\citep{schick2023toolformer}, we compile closed-form gates aligned with numeric BPE chunking~\citep{singh2024tokenization}.

\section{Problem Setup and Analytical Primitives}
\label{sec:theory}
{\setlength{\abovedisplayskip}{3.5pt}\setlength{\belowdisplayskip}{3.5pt}\setlength{\abovedisplayshortskip}{2pt}\setlength{\belowdisplayshortskip}{2pt}

\textbf{Grafting Setup and Calibration Scope.}
Let $\mathcal{M}_{\theta}$ be a pretrained causal LM over vocabulary $\mathcal{V}$ with residual width $d$, $L$ layers, bias-free $\operatorname{RMSNorm}$ $\operatorname{RN}_{w}(z) = w \odot z / \sqrt{\frac{1}{D}\|z\|_2^2 + \varepsilon}$ ($D$ current width, $\varepsilon > 0$), and gated FFNs $\operatorname{GLU}(u) = W_{\mathrm{down}}\big(\phi(W_{\mathrm{gate}} u) \odot (W_{\mathrm{up}} u)\big)$ with $\phi(s) = \operatorname{SiLU}(s) = s \sigma(s)$ (or $\operatorname{GELU}$). Given partial function $f : \mathcal{D} \subset \mathcal{V}^* \to \mathcal{V}^+$ (here signed 6-digit integer $\{+, -, \times\}$, $|A|, |B| \le 999{,}999$, $\deg(A)+\deg(B) \le 5$ for $\times$), \emph{zero-gradient circuit grafting} synthesizes $\theta'$ from $\theta$, tokenizer metadata, and $\le 2L+1$ unlabeled pre-sublayer norm medians (or a $0$-token weight-analytic estimate; \cref{sec:experiments}), with zero gradient steps, labeled data, runtime hooks, or host interpreters, targeting: (i)~exact in-domain execution on $\mathcal{D}$; (ii)~bounded per-layer norm drift and empirical NL noninferiority on evaluated benchmarks; and (iii)~inline resumption to $p_{\theta}$ immediately post-answer.

\begin{remark}[Qualified $\mathsf{TC}^0$ Class vs.\ Circuit Scope]
\label{rem:tc0}
Under $\mathcal{O}(\log n)$-bit (or fixed IEEE float) precision and $\mathsf{DLOGTIME}$-uniform positional encodings, a fixed-$L$-layer attention pass without trace emission lies in $\mathsf{TC}^0$~\citep{merrill2023parallelism}. While recurrent delta-rule layers with active state carry ($e^{g_t} \ne 0$) can reach $\mathsf{NC}^1$~\citep{merrill2024illusion}, our grafted \texttt{Qwen3\_5GatedDeltaNet} subspace clamps $e^{g_t} \equiv 0$ (\cref{thm:routing}(2)), remaining in constant-depth $\mathsf{TC}^0$. By~\citet{hesse2002uniform}, asymptotic $n$-bit integer division and iterated multiplication \emph{do} reside in $\mathsf{DLOGTIME}$-uniform $\mathsf{TC}^0$; omitting division (\texttt{/}) in our fixed 6-digit layout is an \emph{engineering scope choice} (wire-width and $7$-bit \texttt{BF16} range budgets), whereas unbounded halting loops require recurrent traces~\citep{giannou2023looped,percepta2026vm}.
\end{remark}

Synthesizing digital gates inside pretrained LMs faces four obstacles: (1)~bias-free layers after $\operatorname{RMSNorm}$ rescale wires by unknown $\lambda_{t,\ell} = (\frac{1}{D}\|h_{t,\ell}\|_2^2 + \varepsilon)^{-1/2}$ varying $>250\times$~\citep{sun2024massive}; (2)~cross-token routing uses RoPE~\citep{su2024roformer} or \texttt{GatedDeltaNet}~\citep{yang2025gated,qwen2026qwen35}; (3)~residual covariance is full-rank on $\mathbb{R}^d$; and (4)~tied heads ($W_{\mathrm{lm}} = W_{\mathrm{emb}}$) couple logits to Layer-0 embeddings.

\textbf{Geometric Intuition (\cref{fig:intuition}).}
Before stating formal bounds (\cref{thm:mellin,thm:routing,thm:subspace}; proofs \& notation glossary in \cref{app:proofs}), two geometric mechanisms govern our primitives:
(i)~\emph{Log-grid translation invariance (\cref{fig:intuition}a):} $\operatorname{RMSNorm}$ multiplies reference $r_0$ and payload $x$ by the \emph{same} unknown $\lambda_{t,\ell}$. Evaluating SwiGLU across exponentially spaced gate scales $\alpha_k = e^{k\Delta}$ converts multiplication by $\lambda_{t,\ell}$ into a horizontal shift $k\Delta \mapsto k\Delta + \ln\lambda_{t,\ell}$ along a flat trapezoidal integral plateau, canceling $\lambda_{t,\ell}$ across $>250\times$ swings where single-scale gates collapse or saturate.
(ii)~\emph{Signed self-sink routing \& ratio cancellation (\cref{fig:intuition}b):} At long-context $\theta_{\mathrm{RoPE}}=10^7$, standard positive RoPE weights leave sidelobes $S(\delta) \ge +0.61 > 0$, leaking attention onto off-target tokens. Assigning signed Chebyshev key weights ($k_m = \operatorname{sgn}(w_m)\sqrt{|w_m|}$) enforces $S(0)=+1$ and $S(\delta) \le -0.355 < 0$ for all $1 \le |\delta| \le 1024$: when target $t-d_0$ is inactive, query $t$ parks cleanly on its zero-payload self-sink ($\delta_{\mathrm{self}}=0$, logit $0 > -85$), and when active, pairing $V_s = [\lambda_s, \lambda_s x_s]^\top$ divides out source $\lambda_{s^*}$ at the target via Mellin readout $\hat{v}_t / \hat{r}_t = x_{s^*}$.

\begin{figure*}[t]
\vskip 0.01in
\begin{center}
\resizebox{0.97\textwidth}{!}{%
\begin{tikzpicture}[
  font=\footnotesize,
  box/.style={draw=black!80, rounded corners=2.5pt, thick, align=center, inner sep=3.0pt},
  nlbox/.style={box, fill=blue!5},
  circbox/.style={box, fill=orange!8},
  gatebox/.style={box, fill=teal!8},
  arr/.style={-{Stealth[length=4.5pt,width=3.5pt]}, thick, draw=black!80}
]
  \node[nlbox, text width=5.45cm, minimum height=0.76cm] (hnl) at (0.05, 2.25) {\textbf{Pretrained NL Subspace} $h_{\mathrm{NL}} \in \mathbb{R}^{d}$\\[1pt]{\scriptsize Unmodified $W_{\mathrm{NL}}$;\; $\gamma_{\mathrm{NL}}=\sqrt{d/(d+K)}$}};
  \node[circbox, text width=8.25cm, minimum height=0.76cm] (hcirc) at (7.40, 2.25) {\textbf{Appended Circuit Subspace} $h_{\mathrm{circ}} \in \mathbb{R}^{K}$ ($M_0 \in \{2^{-4}, 2^{-9}\}$)\\[1pt]{\scriptsize \texttt{0x0000} complement gates $\cdot$ Virgin fetch/ring pools $\cdot$ Zero NL cross-talk}};

  \node[gatebox, text width=4.55cm, minimum height=1.32cm] (stage1) at (0.45, 0.42) {\textbf{1. Scheduled Parser}\\[1pt]{\scriptsize\textbf{Granite (L0--L10):} Indef-RoPE GQA}\\[0pt]{\scriptsize 51 BPE chunk patterns; sym.\ \texttt{\$}-guard}\\[0pt]{\scriptsize\textbf{Qwen (L0--L8):} \texttt{GatedDeltaNet} shifts}};

  \node[gatebox, text width=4.55cm, minimum height=1.32cm] (stage2) at (5.65, 0.42) {\textbf{2. Scheduled Signed ALU}\\[1pt]{\scriptsize\textbf{Granite (L11--L29)} / \textbf{Qwen (L9--L27)}}\\[0pt]{\scriptsize Two-layer Mellin-SwiGLU ($\Delta=0.55$)}\\[0pt]{\scriptsize Signed $\pm$ ripple carry \& $5{\times}5$ prod.}};

  \node[gatebox, text width=4.65cm, minimum height=1.32cm] (stage3) at (10.90, 0.42) {\textbf{3. Canonical BPE Emitter}\\[1pt]{\scriptsize\textbf{Granite (L30--L39):} Base-$10^3$ BPE $W_{\mathrm{lm}}$}\\[0pt]{\scriptsize\textbf{Qwen (L15--L31):} $(+254,-256)$ tied}\\[0pt]{\scriptsize $\Delta z_{\mathrm{circ}}^{\mathrm{S1}} = 0.0$ post-answer}};

  \node[nlbox, text width=14.10cm, minimum height=0.72cm] (out) at (5.70, -1.62) {\textbf{Unified Next-Token Logits} $z = z_{\mathrm{NL}} + \Delta z_{\mathrm{circ}} \in \mathbb{R}^{|\mathcal{V}|}$\\[1pt]{\scriptsize Active step $\Rightarrow \max \Delta z_{\mathrm{circ}} \in [+68, +11{,}800]$ nats;\quad Non-trigger or Post-answer step $\Rightarrow \Delta z_{\mathrm{circ}} \approx 0.0$ (Probabilistic NL Handoff)}};

  \draw[arr] (hcirc.south) -- ++(0,-0.20) -| (stage1.north);
  \draw[arr] (stage1.east) -- (stage2.west);
  \draw[arr] (stage2.east) -- (stage3.west);
  \draw[arr] (stage3.south) -- (stage3.south |- out.north);
  \draw[arr] (hnl.west) -- ++(-0.38,0) |- (out.west);
\end{tikzpicture}%
}
\caption{\textbf{Zero-gradient orthogonal circuit grafting.} Pretrained coordinates $\mathbb{R}^d$ remain unprojected ($\gamma_{\mathrm{NL}}=\sqrt{d/(d+K)}$), while $K$ appended wires compose reusable Mellin/routing primitives across a 3-stage pipeline and return control to $z_{\mathrm{NL}}$ post-answer.}
\label{fig:arch}
\end{center}
\vskip -0.14in
\end{figure*}
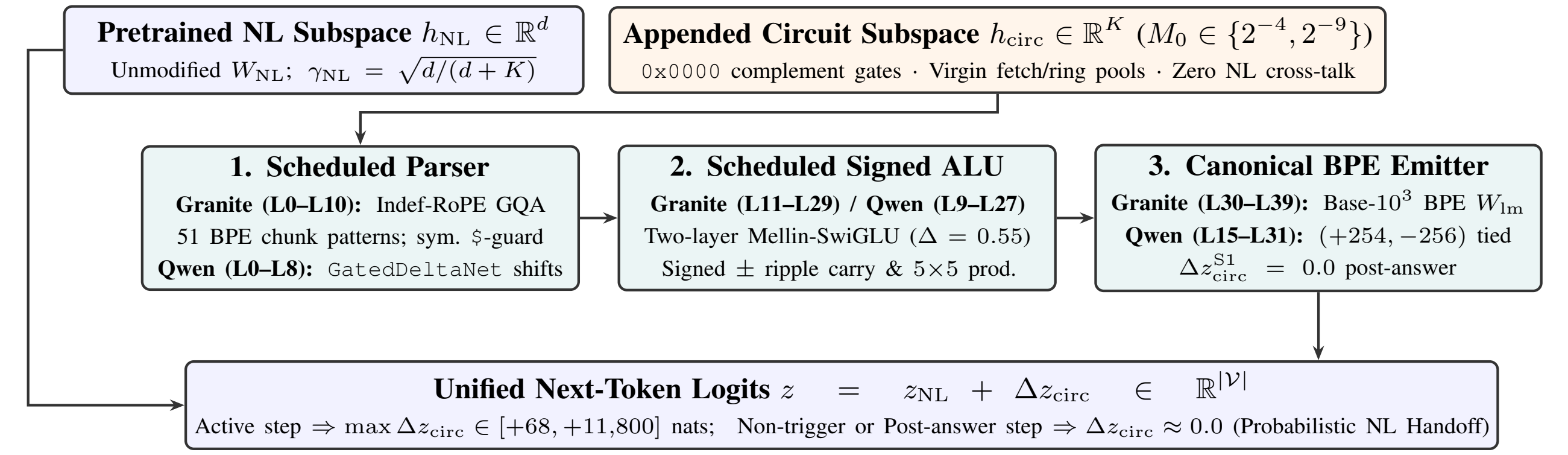

\begin{table*}[t]
\caption{\textbf{Architecture and sparse-vs.-dense footprint summary across 2026 model families} (including 42-layer Gemma~4~E4B architectural audit).}
\label{tab:arch_summary}
\vskip 0.02in
\begin{center}
\scriptsize
\setlength{\tabcolsep}{4.5pt}
\resizebox{\textwidth}{!}{%
\begin{tabular}{@{}llll@{}}
\toprule
\textbf{Property / Footprint Metric} & \textbf{IBM Granite 4.2 3B} & \textbf{Alibaba Qwen 3.5 4B} & \textbf{Google Gemma 4 E4B} \\
\midrule
Backbone \& Vocab ($|\mathcal{V}|$) & 40L Dense RoPE GQA; $100\text{K}$ Untied & 32L Hybrid (24 DeltaNet + 8 Attn); $248\text{K}$ \textbf{Tied} & 42L Dual-RoPE + 18L KV-Share; $262\text{K}$ \\
Numeric BPE \& MLP / Norm & \textbf{1--3 digit chunks} ($51$ pats); SwiGLU ($\theta{=}10^7$) & \textbf{Strict 1-digit}; SwiGLU; $1{+}w$ $\operatorname{RMSNorm}$ & \textbf{Strict 1-digit}; GeGLU + Sandwich + $s_\ell$ \\
Scheduled Carrier \& Width $K$ & Indef-RoPE GQA; $K{=}1{,}140$ (ALU) / $296$ (Bool) & \texttt{GatedDeltaNet} ($0$ Attn); $K{=}2{,}555$ / $311$ (Bool) & Sliding Indef-RoPE (L0--L18); $K{=}296$ (Bool) \\
\textbf{Patch / Sidecar / Dense Rag.} & \textbf{$1.33\text{M}$ ($15.2\text{MB}$) / $7.89\text{GiB}$ / $7.23\text{B}$} & \textbf{$2.39\text{M}$ ($27.9\text{MB}$) / $10.65\text{GiB}$ / $10.22\text{B}$} & \textbf{L24--41 KV fail; Calib.\ L0--18 $100\%$ Bool} \\
\bottomrule
\end{tabular}%
}
\end{center}
\vskip -0.14in
\end{table*}

\begin{theorem}[Scale-Invariant Mellin-GLU Quadrature]
\label{thm:mellin}
Let $h \in \mathbb{R}^{D}$ carry reference $r_0 > 0$ and payload $x \in [-X_{\max}, X_{\max}]$, observed after $\operatorname{RMSNorm}$ as $(\tilde{r}, \tilde{x}) = (\lambda r_0, \lambda x)$ at unknown $\lambda \in [\lambda_{\min}, \lambda_{\max}] \subset (0, \infty)$.
\begin{enumerate}[leftmargin=1.4em,itemsep=1pt,topsep=1pt]
\item \textbf{(Single-layer impossibility over $\mathbb{R}$).} For any bias-free GLU $F_1(u) = W_{\mathrm{down}}\big(\phi(W_{\mathrm{gate}} u) \odot (W_{\mathrm{up}} u)\big)$ with $|\phi(s)| \le C |s|$ ($C=1$ for exact or $\tanh$-$\operatorname{SiLU}$/$\operatorname{GELU}$), $\|F_1(\lambda u)\|_2 \le \lambda^2 C \|W_{\mathrm{down}}\|_2 \|W_{\mathrm{gate}}u\|_\infty \|W_{\mathrm{up}}u\|_2 \to 0$ as $\lambda \to 0^+$ for all $u \in \mathbb{R}^D$. No single bias-free GLU layer realizes $H(x - \tau r_0) = \mathbf{1}[x > \tau r_0]$ on $(0, \lambda_{\max}]$, and achieving $\|F_1(\lambda_{\min} u)\|_2 \ge 1 - \epsilon$ ($\epsilon \in [0,1), u \ne 0$) requires $\|W_{\mathrm{down}}\|_2 \|W_{\mathrm{gate}}\|_2 \|W_{\mathrm{up}}\|_2 \ge \frac{1-\epsilon}{C\|u\|_2^2}\lambda_{\min}^{-2}$.
\item \textbf{(Mellin identity and payload-linear bounds over $\mathbb{R}$).} For $\phi = \operatorname{SiLU}$, $\tilde{r} > 0$, and $\tilde{v} \in \mathbb{R}$,
\begin{equation}
\label{eq:mellin_integral}
\begin{aligned}
&\frac{12}{\pi^2}\!\int_{-\infty}^{\infty}\!\operatorname{SiLU}(-e^{u}\tilde{r})(-e^{u}\tilde{v})\,du\\[-2pt]
&\qquad= \frac{\tilde{v}}{\tilde{r}}\frac{12}{\pi^2}\!\int_{0}^{\infty}\!\frac{t\,dt}{e^{t}+1}
= \frac{\tilde{v}}{\tilde{r}}.
\end{aligned}
\end{equation}
For $\tilde{r} = \lambda r_0$, $\tilde{v} = \lambda x$, $\Delta > 0$, and $K_- < K_+$ with $\lambda_{\min} r_0 e^{K_+ \Delta} \ge 2$, define $Q_{\Delta, K_-, K_+}(\tilde{r}, \tilde{v}) := \frac{12\Delta}{\pi^2}\sum_{k=K_-}^{K_+} \operatorname{SiLU}(-e^{k\Delta}\tilde{r})(-e^{k\Delta}\tilde{v})$. Since $Q_{\Delta, K_-, K_+}(\lambda r_0, \lambda x) = \frac{x}{r_0} S_{\Delta, K_-, K_+}(\ln(\lambda r_0))$ is linear in $x$, both discretization and window-tail errors scale by $|x|/r_0$:
\begin{equation}
\label{eq:mellin_bound}
\begin{aligned}
&\Big|Q_{\Delta, K_-, K_+}(\lambda r_0, \lambda x) - \frac{x}{r_0}\Big|\\[-2pt]
&\qquad\le \frac{|x|}{r_0}\big(\varepsilon_{\mathrm{disc}}(\Delta) + \varepsilon_{\mathrm{tail}}(\lambda r_0)\big),
\end{aligned}
\end{equation}
with $\varepsilon_{\mathrm{disc}}(\Delta) \le 4\sum_{m=1}^{\infty}\!\sqrt{(1 + \frac{4\pi^2 m^2}{\Delta^2})\frac{2\pi^2 m/\Delta}{\sinh(2\pi^2 m/\Delta)}}$ ($\varepsilon_{\mathrm{disc}}(0.55) < 6.26 \times 10^{-6}$) and $\varepsilon_{\mathrm{tail}}(r; K_-, K_+, \Delta) \le \tfrac{12}{\pi^2}[\tfrac{1}{2}r^2 e^{2 K_- \Delta} + (r e^{K_+ \Delta} + 1)e^{-r e^{K_+ \Delta}}]$. For $\beta, \gamma, \delta_{\mathrm{sep}} > 0$ with $z := \beta\lambda_{\min}\delta_{\mathrm{sep}} r_0 \ge 2$, composing Stage~1 $p = \gamma \operatorname{SiLU}(\beta(\tilde{x} - \tau\tilde{r}))(\beta(\tilde{x} - \tau\tilde{r}))$ in layer $\ell$ with Stage~2 $Q_{\Delta, K_-, K_+}(\tilde{p}, \tilde{p})$ in layer $\ell+1$ ($\lambda_{\ell+1} \in [\lambda_{\min}', \lambda_{\max}']$, ratio $1$) yields $0 < Q_{\Delta, K_-, K_+}(\tilde{p}, \tilde{p}) \le \frac{12\Delta}{\pi^2(1 - e^{-2\Delta})}(\lambda_{\max}'\gamma z^2 e^{-z})^2 e^{2 K_+ \Delta}$ when $x \le (\tau - \delta_{\mathrm{sep}})r_0$, and $|Q_{\Delta, K_-, K_+}(\tilde{p}, \tilde{p}) - 1| \le \varepsilon_{\mathrm{disc}}(\Delta) + \varepsilon_{\mathrm{tail}}(\tilde{p})$ when $x \ge (\tau + \delta_{\mathrm{sep}})r_0$ and $\tilde{p}_{\min} e^{K_+\Delta} \ge 2$ ($\tilde{p}_{\min} := \frac{1}{2}\lambda_{\min}' \gamma z^2$).
\item \textbf{(Kernel-explicit IEEE~754 bounds vs.\ empirical \texttt{BF16} carry sensitivity).} Complement-pure gating via $g_{\mathrm{off}} \in \{0, 1\}$ (\texttt{0x0000} active; $\le -z \le -S\lambda_{\min}$ inactive) adds zero active error and yields bit-exact $-0.0$ when inactive: (a)~for $z \ge 89$ under direct IEEE reciprocal-exp kernels $\operatorname{fl}(-z/(1+\exp(z)))$ without \texttt{-ffast-math} (ATen \texttt{F.silu} on CPU/MPS/CUDA), and (b)~for $z \ge 104$ (\texttt{FP32}), $z \ge 98$ (\texttt{FP32}$\to$\texttt{BF16}), and $z \ge 93$ (\texttt{BF16}) under range-split sigmoid kernels $\operatorname{fl}(-z\exp(-z)/(1+\exp(-z)))$ (compiled $S=256 \Rightarrow z \approx 307$). Accounting for gate-exponential conditioning ($\frac{18\zeta(3)}{\pi^2} \le 2.20$), elementary roundoffs $\varepsilon_{\mathrm{elem}}$, odd-neuron $0.35\%$ trim ($\varepsilon_{\mathrm{trim}} \le 0.0018$), and Higham's $N_q$-term accumulation bound $\gamma_{N_q} = \frac{N_q u_{\mathrm{acc}}}{1 - N_q u_{\mathrm{acc}}}$~\citep{higham2002accuracy}, worst-case error on payload $C$ ($r_0=1$) is $\le |C|(\varepsilon_{\mathrm{disc}} + \varepsilon_{\mathrm{tail}} + \varepsilon_{\mathrm{trim}} + \varepsilon_{\mathrm{elem}} + \gamma_{N_q} + u_{\mathrm{out}})$. For $|C| \le 81$ ($N_q \le 39$), \texttt{FP32} error is $< 0.15 < 0.5$ ($<0.24$ with \texttt{BF16}-stored weights) and TensorCore \texttt{TF32} error is $< 0.31 < 0.5$ (guaranteed up to $|C| \le 133$), \textbf{formally guaranteeing} exact carries at $\tau = 10m - 0.5$; conversely, \texttt{BF16} worst-case bounds exceed $0.5$ above $|C| > 28$ ($1.43$ with \texttt{FP32} GEMM accumulation at $|C|=81$), so $\le 1\text{-ULP}$ ($0.5$) \texttt{BF16} carry drift on $C \in [32, 81]$ is an \emph{empirical observation}.
\end{enumerate}
\end{theorem}

\begin{table*}[t]
\caption{\textbf{Factorized block-sparse sidecar vs.\ materialized dense footprint and Hybrid PTQ} (\texttt{bs=1} on NVIDIA A100 PCIe/SXM4; $1\text{ GiB}{=}2^{30}\text{ B}$). \emph{SparseSidecar} (\cref{app:sub_sparse}) keeps $W_{\mathrm{NL}}$ at width $d{=}2560$ and runs active $W_{\mathrm{circ}}$ slices in \texttt{BF16} or \texttt{FP32} ($100\%$ \texttt{G3--G7}), supporting group-$128$ \texttt{INT8}/\texttt{INT4} $W_{\mathrm{NL}}$.}
\label{tab:systems_footprint}
\vskip 0.02in
\begin{center}
\scriptsize
\setlength{\tabcolsep}{3.3pt}
\resizebox{\textwidth}{!}{%
\begin{tabular}{@{}llcccccccc@{}}
\toprule
\textbf{Model} & \textbf{Runtime Representation} & \textbf{Params / NNZ} & \textbf{Exp.} & \textbf{Disk} & \textbf{GFLOPs} & \textbf{A100 VRAM} & \textbf{Pref.\ $512 / 2048\text{ms}$} & \textbf{Arith / G3--7} & \textbf{ARC-C} \\
\midrule
\multirow{4}{*}{\textbf{Granite 4.2}}
 & Pretrained Base ($d{=}2560$, \texttt{BF16}) & $3.660\text{B}$ dense & $1.00\times$ & $6.82\text{ GiB}$ & $6.86$ & $6.83\text{ GiB}$ & $34.3 / 95.6$ & $15.7\%$ / $0/5$ & $75.5\%$ \\
 & Grafted Dense Rag./Unif.\ ($d{+}K$) & $7.23 / 10.24\text{B}$ & $1.98$--$2.80\times$ & $13.5\text{ GiB}$ & $13.80$ & $13.55 / 19.19\text{ GiB}$ & $99.4 / 347.5$ & $96.1\%$ / $4/5^*$ & $75.5\%$ \\
 & \textbf{SparseSidecar (\texttt{BF16} / \texttt{FP32} circ)} & $+1.33\text{M}$ nnz & $\mathbf{1.0004\times}$ & $+15.2\text{ MB}$ & $\mathbf{6.863}$ & $\mathbf{7.89 / 8.93\text{ GiB}}$ & $\mathbf{53.1 / 130.1^\ddagger}$ & $\mathbf{97.2\% / 5/5}$ & $\mathbf{75.5 / 76.0\%}$ \\
 & \textbf{Hybrid PTQ (\texttt{INT8 / INT4} + \texttt{FP32})} & $3.66\text{B}$ q + nnz & $0.5$--$1\text{ B/p}$ & $2.1$--$3.7\text{ GiB}$ & $6.863$ & $\mathbf{6.07 / 4.60\text{ GiB}}$ & $96.2 / 132.7^\dagger$ & $\mathbf{97.2\% / 5/5}$ & $75.0 / 65.0\%$ \\
\midrule
\multirow{4}{*}{\textbf{Qwen 3.5}}
 & Pretrained Base ($d{=}2560$, \texttt{BF16}) & $4.206\text{B}$ dense & $1.00\times$ & $7.83\text{ GiB}$ & $8.50$ & $7.94\text{ GiB}$ & $110.5 / 290.8$ & $47.3\%$ / --- & $91.5\%$ \\
 & Grafted Dense Rag./Unif.\ ($d{+}K$) & $10.22 / 13.18\text{B}$ & $2.43$--$3.13\times$ & $19.0\text{ GiB}$ & $20.54$ & $19.09 / 24.65\text{ GiB}$ & $232.2 / 847.9$ & $100.0\%$ / $5/5$ & $91.5\%$ \\
 & \textbf{SparseSidecar (\texttt{BF16} / \texttt{FP32} circ)} & $+2.39\text{M}$ nnz & $\mathbf{1.0006\times}$ & $+27.9\text{ MB}$ & $\mathbf{8.505}$ & $\mathbf{10.65 / 13.17\text{ GiB}}$ & $\mathbf{159.7 / 433.7^\ddagger}$ & $\mathbf{100.0\% / 5/5}$ & $\mathbf{91.5 / 92.0\%}$ \\
 & \textbf{Hybrid PTQ (\texttt{INT8 / INT4} + \texttt{FP32})} & $4.21\text{B}$ q + nnz & $0.5$--$1\text{ B/p}$ & $2.5$--$4.3\text{ GiB}$ & $8.505$ & $\mathbf{9.80 / 8.14\text{ GiB}}$ & $192.9 / 232.1^\dagger$ & $\mathbf{100.0\% / 5/5}$ & $\mathbf{92.0 / 93.5\%}$ \\
\bottomrule
\end{tabular}%
}
\end{center}
\vspace{-2.5pt}
{\scriptsize $^*$$4/5$ \texttt{G3--G7} in pure \texttt{BF16} ($5/5$ in \texttt{FP32}/\texttt{TF32}/\texttt{FP32}-circ). $^\ddagger$\texttt{BF16}-circ $512/2048$-tok prefill ($68.3/215.2\text{ms}$ Granite, $167.1/524.7\text{ms}$ Qwen w/ \texttt{FP32}-circ). $^\dagger$512-tok \texttt{INT8}/\texttt{INT4} prefill.}
\vskip -0.14in
\end{table*}

\begin{table*}[t]
\caption{\textbf{Four-level decomposed execution accuracy and NL retention} on unseen 1--6 digit signed $\{+, -, \times\}$ (\texttt{Arith-1200}), \texttt{Inline-50}, held-out \texttt{GSM8K[60:200]} ($N{=}140$), 8-task NL (\texttt{NI-8}, $N{=}2{,}350$), and $10.4\text{K}$ false-trigger (\texttt{FT}). $^*$Host Python router ($0$ weights). $^\dagger$\texttt{GSM8K[:60]} ($70.00\%$ canonical); $^\ddagger$triggers $50/50$ on \texttt{\$a + b =}.}
\label{tab:main_eval}
\vskip 0.02in
\begin{center}
\scriptsize
\setlength{\tabcolsep}{3.8pt}
\resizebox{\textwidth}{!}{%
\begin{tabular}{@{}llcccccccc@{}}
\toprule
\textbf{Model \& Method} & \textbf{Dense} & \textbf{ALU-Pref} & \textbf{Lead-Int} & \textbf{Strict-ND (Bare / Prose)} & \textbf{Inline} & \textbf{PPL$\downarrow$ / KL$\downarrow$} & \textbf{GSM-140} & \textbf{NI-8} & \textbf{FT} \\
\midrule
\multicolumn{10}{@{}l}{\emph{\textbf{IBM Granite 4.2 3B} ($d=2560$; 1--3 digit BPE; Canonical Graft $K=1{,}140$, $1.33\text{M}$ nnz; 1-Digit Ablation $K=1{,}043$, $6.18\text{B}$ ragged)}} \\
Pretrained Base (\texttt{BF16}) & $3.66\text{B}$ & --- & $15.67\%$ & --- & $26.0\%$ & $14.30$ / $.0000$ & $68.57\%$ & $55.79\%$ & --- \\
External Python Router$^*$ & $3.66\text{B}$ & $100.0^*\%$ & $100.0^*\%$ & $100.0^*\%$ & Host$^*$ & $14.30$ / $.0000$ & --- & $55.79\%$ & --- \\
Fixed-$d$ Coord.\ Proj.\ (\texttt{BF16}) & $3.66\text{B}$ & --- & $1.67\%$ & --- & $0.0\%$ & $44{,}801$ / $8.74$ & --- & --- & --- \\
Fixed-$d$ SVD Proj.\ $P_\perp$ (\texttt{BF16}) & $3.66\text{B}$ & --- & $97.50\%$ & --- & $0.0\%$ & $1{,}742.8$ / $5.42$ & --- & --- & --- \\
1-Digit Off-Manifold Emitter (\texttt{BF16}) & $6.18\text{B}$ & $99.50\%$ & $\mathbf{99.42\%}$ & $99.42\%$ & $\mathbf{98.0\%}$ & $14.29$ / $.0014$ & \emph{36.67\%}$^\dagger$ & --- & $50/50^\ddagger$ \\
\textbf{Ours Canonical (\texttt{FP32} / \texttt{BF16})} & $7.23\text{B}$ & $\mathbf{100 / 99.17\%}$ & $97.08\%$ & $90.75\%$ ($87.0$ / $\mathbf{98.25\%}$) & $\mathbf{98.0\%}$ & $\mathbf{14.30 / .0014}$ & $\mathbf{70.71\%}$ & $\mathbf{56.13\%}$ & $\mathbf{0/10.4\text{K}}$ \\
\midrule
\multicolumn{10}{@{}l}{\emph{\textbf{Alibaba Qwen 3.5 4B} ($d=2560$; 24$\times$\texttt{GatedDeltaNet} + 8$\times$Attn; strict 1-digit BPE; Canonical Graft $K=2{,}555$, $2.39\text{M}$ nnz)}} \\
Pretrained Base (\texttt{BF16}) & $4.21\text{B}$ & --- & $47.33\%$ & --- & $42.0\%$ & $9.47$ / $.0000$ & $71.43\%$ & $66.21\%$ & --- \\
\textbf{Ours Canonical (\texttt{FP32} / \texttt{BF16})} & $10.22\text{B}$ & $\mathbf{100 / 100.0\%}$ & $\mathbf{99.50\%}$ & $\mathbf{99.50\%}$ ($\mathbf{99.4}$ / $\mathbf{99.75\%}$) & $\mathbf{100.0\%}$ & $\mathbf{9.47 / .0007}$ & $\mathbf{72.86\%}$ & $\mathbf{65.96\%}$ & $\mathbf{0/10.4\text{K}}$ \\
\bottomrule
\end{tabular}%
}
\end{center}
\vskip -0.14in
\end{table*}

\begin{theorem}[Cross-Token Norm Cancellation]
\label{thm:routing}
Let token $s \in \{0, \dots, t\}$ carry payload $x_s \in [-X_{\max}, X_{\max}]$, activity $c_s \in \{0, 1\}$, and reference $r_s = 1$, scaled after source $\operatorname{RMSNorm}$ by $\lambda_s \in [\lambda_{\min}, \lambda_{\max}] \subset (0, \infty)$.
\begin{enumerate}[leftmargin=1.4em,itemsep=1pt,topsep=1pt]
\item \textbf{(Indefinite-RoPE self-sink GQA over $\mathbb{R}$).} Assume $0 \le d_0 \le t \le \Delta_{\max}$, scale $\kappa_{\mathrm{attn}} > 0$, Q/K norms $\lambda_{t,q}, \lambda_{u,k} \ge \lambda_{\min,\mathrm{qk}} > 0$, and frequencies $\{\omega_m\}_{m=1}^{d_{\mathrm{rot}}/2}$ with weights $w$ satisfying $S(0) = \sum_m w_m = 1$ and $\max_{1 \le |\delta| \le \Delta_{\max}} S(\delta) \le -\mu < 0$ ($S(\delta) = \sum_m w_m \cos(\delta \omega_m)$). Setting $K_s = c_s \operatorname{RoPE}(k_{\mathrm{base}}, s)$ and $V_s = c_s [\lambda_s, \lambda_s x_s]^\top$, when $c_{s^*} = 1$ at $s^* = t - d_0$, letting $N_{\mathrm{act}} = |\{u \ne s^* : c_u = 1\}|$ and $\eta_t^{\mathrm{act}} = \sum_{u \ne s^*, c_u=1} \alpha_{t,u}/\alpha_{t,s^*}$ gives over $\mathbb{R}$:
\begin{equation}
\label{eq:rope_cancel}
\begin{aligned}
\left|\frac{\hat{v}_t}{\hat{r}_t} - x_{s^*}\right| &\le \frac{\eta_t^{\mathrm{act}} R_{\max}}{1 + \eta_t^{\mathrm{act}} R_{\min}}\big(|x_{s^*}| + X_{\max}\big),\\
\eta_t^{\mathrm{act}} &\le N_{\mathrm{act}} \exp\!\big({-}(1+\mu)\kappa_{\mathrm{attn}} g^2 \lambda_{\min,\mathrm{qk}}^2\big),
\end{aligned}
\end{equation}
with $R_{\max/\min} = \max/\min_{u: c_u=1}(\lambda_u/\lambda_{s^*})$. If $c_{s^*} = 0$ and some $u_0 \le t$ has $c_{u_0} = 0$, $|\hat{r}_t| \le N_{\mathrm{act}} \lambda_{\max} \exp(-\mu \kappa_{\mathrm{attn}} g^2 \lambda_{\min,\mathrm{qk}}^2)$.
\item \textbf{(Memoryless \texttt{Qwen3\_5GatedDeltaNet} shift).} In causal zero-padded \texttt{GatedDeltaNet}~\citep{yang2025gated,qwen2026qwen35} with $\operatorname{Conv1d}$ width $W_c = 4$, $\operatorname{SiLU}$, $L_2$-normalized $(q, k)$ ($\varepsilon > 0$, scale $d_h^{-1/2}$), $S_t = e^{g_t} S_{t-1} + \hat{k}_t (v_t - e^{g_t} S_{t-1}^\top \hat{k}_t)^\top \beta_t$ ($g_t = -e^{A_{\log}}\operatorname{softplus}(a_t + b_{\Delta})$), and $\operatorname{RMSNormGated}_{w_{\mathrm{norm}}}(S_t^\top \hat{q}_t, z_t)$ ($w_{\mathrm{norm}, c} \ne 0$, $\|S_{t-1}\|_F \le S_{\max}$), setting $W_a = 0$ ($a_t = 0$), $A_{\log} = 10, b_{\Delta} = 10$ gives $g_t = -e^{10}\ln(1+e^{10}) < -2.20 \times 10^5$ ($e^{g_t} < 10^{-95{,}600}$ over $\mathbb{R}$; \textbf{exact \texttt{0.0} in \texttt{FP32}/\texttt{BF16}}). Routing $\tilde{r}_t > 0$ to $(q_0, k_0)$, $4\tilde{r}_t$ to $z_t$, and $s=0$ ($\tilde{r}_{t-\delta}$) plus payloads $u^{(j)}$ ($1 \le j < d_h/2$) into antisymmetric pairs $(\pm\frac{1}{2}u^{(j)})$ at tap $\delta \in \{1,2,3\}$ with uniform readout $(\pm\alpha_{\mathrm{out}} / w_{\mathrm{norm}, 2s(+1)})$ ($\alpha_{\mathrm{out}} > 0$) obeys $\operatorname{SiLU}(u) - \operatorname{SiLU}(-u) \equiv u$. Under $e^{g_t} = 0$, $(\hat{q}_t^\top \hat{k}_t)\beta_t \nu_t^{-1}\operatorname{SiLU}(4\tilde{r}_t)$ and $\lambda_{t-\delta}$ \textbf{cancel identically} against $s=0$ for $t \ge \delta$, recovering exact $x_{t-\delta}^{(j)}$ ($0$ for $t < \delta$).
\end{enumerate}
\end{theorem}

\begin{theorem}[Subspace Projection vs.\ Width Extension and Tied-GEMM Cancellation]
\label{thm:subspace}
Let $h \in \mathbb{R}^d$ satisfy $0 < \mathbb{E}[\|h\|_2^2] < \infty$ with $\Sigma = \mathbb{E}[h h^\top] \succ 0$ ($\sigma_1 \ge \dots \ge \sigma_d > 0$).
\begin{enumerate}[leftmargin=1.4em,itemsep=1pt,topsep=1pt]
\item \textbf{(Fixed-width projection lower bound over $\mathbb{R}$).} Any rank-$(d-K)$ projection $P_\perp = I_d - U U^\top$ ($U^\top U = I_K$) vacating $K$ wires inside $\mathbb{R}^d$ without gradient recovery incurs error bounded below by Ky Fan's theorem~\citep{fan1949theorem}:
\begin{equation}
\label{eq:kyfan}
\mathbb{E}\big[\|h - P_\perp h\|_2^2\big] \;=\; \operatorname{tr}(U^\top \Sigma U) \;\ge\; \sum_{i=d-K+1}^{d} \sigma_i \;>\; 0.
\end{equation}
\item \textbf{(Axis-aligned width extension and schedule-/tree-explicit tied-GEMM bounds).} Extending $h_{\mathrm{ext}} = [h_{\mathrm{NL}}; h_{\mathrm{circ}}] \in \mathbb{R}^{d+K}$ axis-aligned (diagonal $\operatorname{RMSNorm}$ is not $O(d+K)$-equivariant) with zero cross-coupling on $W_{\mathrm{NL}}$ and $\gamma_{\mathrm{NL}} = \sqrt{d / (d+K)}$ bounds single-sublayer relative drift over $\mathbb{R}$ (whenever $\|\operatorname{RN}_{w_{\mathrm{NL}}}(h_{\mathrm{NL}})\|_2 > 0$) by
\begin{equation}
\label{eq:ortho_pert}
\begin{aligned}
&\frac{\|\operatorname{RN}_{\gamma_{\mathrm{NL}} w_{\mathrm{NL}}}(h_{\mathrm{ext}})_{1:d} - \operatorname{RN}_{w_{\mathrm{NL}}}(h_{\mathrm{NL}})\|_2}{\|\operatorname{RN}_{w_{\mathrm{NL}}}(h_{\mathrm{NL}})\|_2}\\
&\quad= \big|1 - (1 + \rho' + \delta_\varepsilon)^{-1/2}\big| \le \tfrac{1}{2}(\rho + \delta_\varepsilon),
\end{aligned}
\end{equation}
where $\rho' = \frac{\|h_{\mathrm{circ}}\|_2^2}{\|h_{\mathrm{NL}}\|_2^2 + d\varepsilon} \le \rho = \frac{\|h_{\mathrm{circ}}\|_2^2}{\|h_{\mathrm{NL}}\|_2^2}$ and $\delta_\varepsilon = \frac{K \varepsilon}{\|h_{\mathrm{NL}}\|_2^2 + d\varepsilon}$. Under tied $W_{\mathrm{lm}} = W_{\mathrm{emb}}$ and unit-offset norm $\tilde{h}_i = \operatorname{fl}((1 + w_i)\hat{h}_i)$ ($\hat{h} = h_{\mathrm{ext}}/\mathrm{rms}(h_{\mathrm{ext}})$), setting $E_{v_m, p_m} = E_{v_m, n_m} = \alpha_0 M_0$ ($m \in \{1..12\}$), $w_{i} = -1.0$ on non-readout wires, and $(w_{p_m}, w_{n_m}) = (+254.0, -256.0)$ ($1 + w = \pm 255.0$) gives $\tilde{h}_{n_m} = -\tilde{h}_{p_m}$ bit-for-bit and $\Delta z_{\mathrm{circ}}^{\mathbb{R}}(v) = 0$ over $\mathbb{R}$ when $(p_m, n_m)$ receive zero writes. In IEEE~754 arithmetic (no overflow, accumulator roundoff $u_{\mathrm{acc}}$, product underflow $\eta_{\mathrm{uf}}$):
\begin{itemize}[leftmargin=1.2em,itemsep=0.5pt,topsep=1pt]
\item \emph{All inactive rows/steps:} If $v \notin \{v_1..v_{12}\}$ or $x_t \notin \{v_1..v_{12}\}$, all $K$ circuit products are $\pm 0.0$ under any reduction order or FMA.
\item \emph{Schedule S1 (Pairwise-first non-FMA + block split):} Reducing each $(p_m, n_m)$ pair first without FMA gives $\Delta z_{\mathrm{circ}}^{\mathrm{S1}}(v) = +0.0$ and $\operatorname{fl}(\hat{z}_{\mathrm{NL}}(v) + (+0.0)) = \hat{z}_{\mathrm{NL}}(v)$ numerically ($=$ bit-for-bit if $\hat{z}_{\mathrm{NL}}(v) \ne -0.0$) across all $v \in \mathcal{V}$.
\item \emph{Schedule S2 (Contiguous tail after $\hat{z}_{\mathrm{NL}}$):} Appending $+q_{\mathrm{exact}}, -q_{\mathrm{exact}}$ ($q_{\mathrm{exact}} := E_{v_{m^*},p_{m^*}}\tilde{h}_{p_{m^*}}$) consecutively after $\hat{z}_{\mathrm{NL}}(v_{m^*})$ satisfies $|\hat{z}^{\mathrm{S2}}(v_{m^*}) - \hat{z}_{\mathrm{NL}}(v_{m^*})| \le \gamma_2(|\hat{z}_{\mathrm{NL}}(v_{m^*})| + |q_{\mathrm{exact}}|) + 2(1+\gamma_2)\eta_{\mathrm{uf}}$.
\item \emph{Arbitrary reduction tree $\mathcal{T}$ with interleaved NL terms:} Two-step bounds fail when NL terms are summed between $+q$ and $-q$ (\cref{app:proofs}). Over \emph{any} binary evaluation tree $\mathcal{T}$ of non-zero leaf height $h_{\mathcal{T}}$ ($(h_{\mathcal{T}}+1)u_{\mathrm{acc}} < 1$, arbitrary interleaving and FMA), active row $v_{m^*} = x_t$ obeys relative to $z_{\mathrm{NL}}^{\mathbb{R}}(v_{m^*}) = \sum_{j=1}^d E_{v_{m^*},j}\tilde{h}_j$:
\end{itemize}
\begin{equation}
\label{eq:diff_tied_tree}
\begin{aligned}
&\big|\hat{z}_{\mathcal{T}}^{\mathrm{acc}}(v_{m^*}) - z_{\mathrm{NL}}^{\mathbb{R}}(v_{m^*})\big| \le (d{+}2)(1{+}\gamma_{h_{\mathcal{T}}})\eta_{\mathrm{uf}}\\
&\quad + \gamma_{h_{\mathcal{T}}+1}\!\left(\sum_{j=1}^{d} |E_{v_{m^*}, j}\tilde{h}_j| + 2|q_{\mathrm{exact}}|\right)\!,
\end{aligned}
\end{equation}
giving $< 2.8 \times 10^{-5}$~nats on Qwen~3.5 under balanced tree reduction ($h_{\mathcal{T}} \le \lceil\log_2(d+K)\rceil = 13$) and $< 5.2 \times 10^{-3}$~nats under worst-case sequential interleaving ($h_{\mathcal{T}} = d+1 = 2561$) in \texttt{FP32} accumulation.
\end{enumerate}
\end{theorem}
}

\section{Primitives and Automated Circuit Synthesis}
\label{sec:arch}

\begin{figure*}[t]
\vskip 0.01in
\begin{center}
\resizebox{0.97\textwidth}{!}{%
\begin{tikzpicture}
  \begin{axis}[
    name=plotA,
    width=0.455\textwidth,
    height=3.20cm,
    ymode=log,
    xmin=0, xmax=1220,
    ymin=0.0001, ymax=140,
    ytick={0.0001,0.01,1,100},
    xlabel={\footnotesize \textbf{(a)} Circuit Subspace Width $K$ (wires)},
    ylabel={\footnotesize Next-Token KL (log)},
    tick label style={font=\scriptsize},
    grid=both,
    grid style={gray!22},
    legend style={
      font=\scriptsize,
      at={(0.5,1.06)},
      anchor=south,
      legend columns=2,
      fill=white,
      draw=gray!45,
      inner sep=1.6pt,
      column sep=4pt,
      row sep=0.5pt
    }
  ]
    \addplot[color=red!75!black, mark=triangle*, thick, mark size=1.7pt] coordinates {
      (32, 0.066512) (48, 0.102908) (64, 0.139896) (80, 0.167489) (1043, 8.7385)
    };
    \addlegendentry{Coord.\ Proj.\ (fixed $d$)}

    \addplot[color=orange!85!black, mark=square*, thick, mark size=1.6pt] coordinates {
      (48, 0.061547) (64, 0.078956) (1043, 5.4159)
    };
    \addlegendentry{SVD Proj.\ $P_\perp$ (0-grad)}

    \addplot[color=purple!75!black, only marks, mark=diamond*, mark size=2.2pt] coordinates {
      (1140, 0.77666)
    };
    \addlegendentry{Uncomp.\ $d{+}K$ (\texttt{M3})}

    \addplot[color=blue!75!black, only marks, mark=*, mark size=2.3pt] coordinates {
      (1140, 0.00142)
    };
    \addlegendentry{Ours Granite ($0.0014$)}

    \draw[dashed, teal!75!black, thick] (axis cs:0,0.000653) -- (axis cs:1220,0.000653);
    \node[anchor=south, font=\scriptsize, text=teal!70!black, fill=white, fill opacity=0.88, text opacity=1, inner sep=1.0pt] at (axis cs:520,0.00090) {Qwen~3.5 ($6.5{\times}10^{-4}$)};
  \end{axis}

  \begin{axis}[
    name=plotB,
    at={(plotA.east)},
    anchor=west,
    xshift=1.90cm,
    width=0.495\textwidth,
    height=3.20cm,
    ybar=1.2pt,
    bar width=3.1pt,
    ymin=0, ymax=112,
    symbolic x coords={Arith-Lead,Blind-ALU,Blind-Strict,InlineRes,GSM8K-200,NI-8},
    xtick=data,
    xticklabel style={font=\scriptsize, rotate=13, anchor=east},
    ytick={0,25,50,75,100},
    xlabel={\footnotesize \textbf{(b)} Evaluation Suite (\texttt{BF16})},
    ylabel={\footnotesize Accuracy (\%)},
    tick label style={font=\scriptsize},
    ymajorgrids=true,
    grid style={gray!22},
    legend style={
      font=\scriptsize,
      at={(0.5,1.06)},
      anchor=south,
      legend columns=2,
      fill=white,
      draw=gray!45,
      inner sep=1.6pt,
      column sep=5pt,
      row sep=0.5pt
    }
  ]
    \addplot[fill=gray!45, draw=gray!80!black, line width=0.5pt] coordinates {
      (Arith-Lead, 15.67) (Blind-ALU, 19.33) (Blind-Strict, 19.33) (InlineRes, 26.0) (GSM8K-200, 69.0) (NI-8, 55.79)
    };
    \addlegendentry{Granite Base}

    \addplot[fill=blue!65, draw=blue!85!black, line width=0.5pt] coordinates {
      (Arith-Lead, 97.08) (Blind-ALU, 100.0) (Blind-Strict, 88.83) (InlineRes, 98.0) (GSM8K-200, 70.5) (NI-8, 56.13)
    };
    \addlegendentry{Granite Graft}

    \addplot[fill=orange!40, draw=orange!80!black, line width=0.5pt] coordinates {
      (Arith-Lead, 47.33) (Blind-ALU, 56.33) (Blind-Strict, 56.33) (InlineRes, 42.0) (GSM8K-200, 72.5) (NI-8, 66.21)
    };
    \addlegendentry{Qwen~3.5 Base}

    \addplot[fill=teal!70, draw=teal!90!black, line width=0.5pt] coordinates {
      (Arith-Lead, 99.50) (Blind-ALU, 100.0) (Blind-Strict, 99.67) (InlineRes, 100.0) (GSM8K-200, 73.0) (NI-8, 65.96)
    };
    \addlegendentry{Qwen~3.5 Graft}
  \end{axis}
\end{tikzpicture}%
}
\caption{\textbf{Subspace geometry and decomposed capability profile (\texttt{BF16}).} \textbf{(a)}~Zero-gradient SVD projection ($P_\perp$) incurs $43\times$--$3{,}800\times$ higher next-token KL than orthogonal width extension ($K=1{,}140$), while uncompensated $d+K$ (\texttt{M3}) inflates KL by $547\times$. \textbf{(b)}~Canonical grafts reach $97.08\%$--$99.50\%$ leading-integer match, $100\%$ blind ALU-prefix exactness, $98\%$--$100\%$ inline resumption, and paired Tango noninferiority on \texttt{GSM8K-200} and \texttt{NI-8} ($N=2{,}350$).}
\label{fig:results}
\end{center}
\vskip -0.14in
\end{figure*}
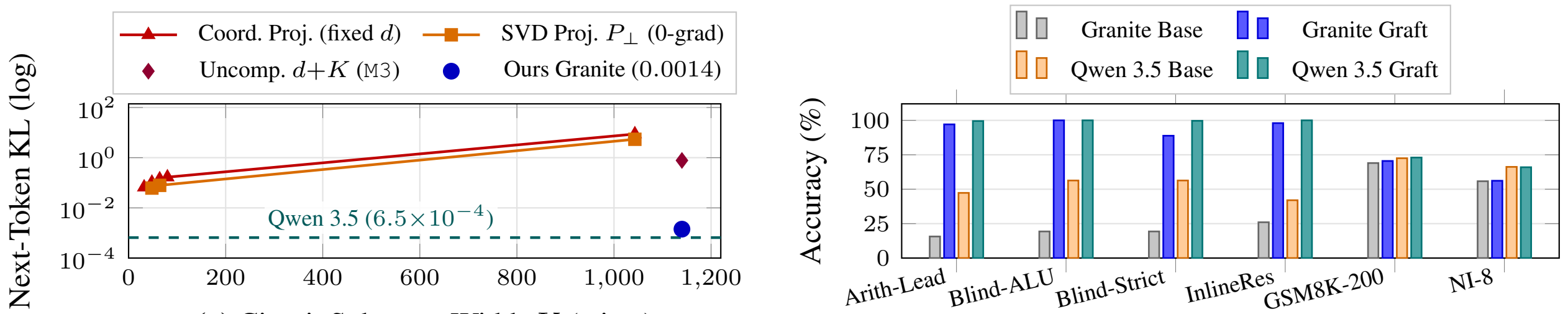

\begin{table*}[t]
\caption{\textbf{Matched-$(d+K)$ controls (\texttt{M0}--\texttt{M4}), 5-primitive leave-one-out ALU ablations (\texttt{-P2}/\texttt{-P3}/\texttt{-P5}), \texttt{LoRA} SFT, and calibration sensitivity (\texttt{C1}--\texttt{C2})} on $30$-doc PPL/KL, \texttt{ARC-C}, and unseen 1--6 digit signed $\{+,-,\times\}$ (\texttt{BF16}).}
\label{tab:matched_calib}
\vskip 0.02in
\begin{center}
\scriptsize
\setlength{\tabcolsep}{3.6pt}
\resizebox{\textwidth}{!}{%
\begin{tabular}{@{}llcccccccc@{}}
\toprule
 & & \multicolumn{4}{c}{\textbf{IBM Granite 4.2 3B ($d{+}K{=}3700$)}} & \multicolumn{4}{c}{\textbf{Alibaba Qwen 3.5 4B ($d{+}K{=}5115$)}} \\
\cmidrule(lr){3-6} \cmidrule(lr){7-10}
\textbf{Group} & \textbf{Configuration / Ablation} & \textbf{PPL$\downarrow$} & \textbf{Mean KL$\downarrow$} & \textbf{ARC-C} & \textbf{Arith} & \textbf{PPL$\downarrow$} & \textbf{Mean KL$\downarrow$} & \textbf{ARC-C} & \textbf{Arith} \\
\midrule
\emph{Ref.} & Pretrained Base / \texttt{LoRA} SFT ($12.8\text{K}$ ex.)$^*$ & $42.10$ & $0.0000$ & $77.3\%$ & $24.7 / 42.2\%^*$ & $29.81$ & $0.0000$ & $93.0\%$ & $73.7 / 61.7\%^*$ \\
\midrule
\multirow{6}{*}{\textbf{Ablate}}
 & \textbf{\texttt{M0}: Ours (All 5 Primitives)} & $\mathbf{42.06}$ & $\mathbf{0.0017}$ & $\mathbf{76.7\%}$ & $\mathbf{96.7\%}$ & $\mathbf{29.84}$ & $\mathbf{0.0008}$ & $\mathbf{92.7\%}$ & $\mathbf{99.7\%}$ \\
 & \texttt{M1}: Zero-Circuit Width ($E_{\mathrm{circ}}{=}0$) & $42.04$ & $0.0017$ & $77.7\%$ & $24.7\%$ & $29.84$ & $0.0008$ & $92.7\%$ & $73.3\%$ \\
 & \texttt{-P1 / M2}: Calibrated Single-Scale SwiGLU & $43.79$ & $0.2194$ & $75.0\%$ & $8.3\%$ & $1399.6$ & $3.8330$ & $91.5\%$ & $20.0\%$ \\
 & \texttt{-P2 / -P3}: Pos.\ RoPE ($w_m{>}0$) / Stateful GDN$^\dagger$ & $42.05$ & $0.0043$ & $75.5\%$ & $\mathbf{15.0\%}$ & $29.84$ & $0.0009$ & $92.0\%$ & $\mathbf{51.1\%^\dagger}$ \\
 & \texttt{-P4 / M3}: w/o $\sqrt{d/(d{+}K)}$ Norm Comp. & $82.27$ & $0.7767$ & $57.5\%$ & $92.2\%$ & $275.9$ & $2.2725$ & $38.5\%$ & $50.0\%$ \\
 & \texttt{M4}: Givens-Rotated Ortho.\ ($\theta=\pi/6$)$^\ddagger$ & $>10^8$ & $>10^6$ & $23.7\%$ & $0.0\%$ & $>10^8$ & $202.42$ & $25.0\%$ & $0.0\%$ \\
\midrule
\multirow{2}{*}{\textbf{Calib.}}
 & \textbf{\texttt{C1}: Digit-Free News ($16$ sents, $0$ math)} & $\mathbf{42.15}$ & $\mathbf{0.0017}$ & $\mathbf{77.0\%}$ & $\mathbf{97.0\%}$ & $\mathbf{29.85}$ & $\mathbf{0.0007}$ & $\mathbf{92.7\%}$ & $\mathbf{99.7\%}$ \\
 & \textbf{\texttt{C2}: Weight-Analytic ($0$ toks, $0$ passes)} & $\mathbf{42.17}$ & $\mathbf{0.0017}$ & $\mathbf{77.3\%}$ & $\mathbf{96.3\%}$ & $\mathbf{29.83}$ & $\mathbf{0.0008}$ & $\mathbf{92.7\%}$ & $\mathbf{97.0\%}$ \\
\bottomrule
\end{tabular}%
}
\end{center}
\vspace{-2.5pt}
{\scriptsize $^*$\texttt{LoRA} ($6.9/6.0\text{M}$ params): $\mathbf{0\%}$ inline prose (vs.\ $98/100\%$ Ours). $^\dagger$Qwen $e^{g_t}{\ne}0, \beta_t{=}0.5$; 1-sided $\operatorname{SiLU}$ signed err $8.999$. $^\ddagger$\texttt{-P5} 1-ended tied readout: $|\Delta z|{=}0.014 \ne 0$.}
\vskip -0.15in
\end{table*}

\textbf{Primitives, Declarative Compiler, and Reference Patches.}
Our implementation spans three layers: (1)~\emph{reusable analytical modules} (\texttt{MellinSwiGLUBank}, Indefinite-RoPE \texttt{GQAAttnCompiler}, \texttt{DeltaNetShiftCompiler}, $\sqrt{d/(d+K)}$ norm scaling, and differential tied readout); (2)~an \emph{automated 5-stage declarative compiler} (\texttt{compile\_declarative\_arithmetic} \& \texttt{DeclarativeDAGScheduler}, \cref{sec:auto_boolean}) that introspects \texttt{config}/\texttt{tokenizer} to synthesize $4$--$8$ digit BPE $\{+,-,\times\}$ and Boolean/voter DAGs with zero manual wire or layer IDs; and (3)~\emph{frozen 6-digit reference patches} ($|A|,|B| \le 999{,}999$, $\deg(A){+}\deg(B) \le 5$; \cref{fig:arch,tab:arch_summary}).

\subsection{Dense Multi-Chunk RoPE GQA: IBM Granite 4.2 3B}
\texttt{ibm-granite/granite-4.2-3b}~\citep{ibm2026granite4} is a 40-layer dense Transformer ($d=2560$, $|\mathcal{V}|=100{,}352$, untied $W_{\mathrm{lm}}$, $5{:}1$ GQA, \texttt{attn}$=1/d_h=0.015625$, unit residual/logit scales, $\theta_{\mathrm{RoPE}}=10^7$). Its BPE merges 1--3 digit substrings into single tokens ($1{,}110$ numeric tokens), producing $51$ token-length compositions per 6-digit integer. Setting $M_0 = 2^{-4}$ keeps Layer-0 $\operatorname{RMSNorm}$ drift $< 2.5 \times 10^{-4}$. Layers L0--L10 instantiate a 64-frequency Chebyshev Indefinite-RoPE kernel ($S(0)=1.0$, $\max_{1 \le \delta \le 1024} S(\delta) = -0.355$; $K_+=22$ in L1/L3/L4) across $25$ appended KV heads to separate binary/unary minus, enforce symmetric pre-operand guards (abstaining on \texttt{\$} prefixes, chains, or decimals), and factor all $51$ BPE compositions into $14$ $(\text{chunk}, \text{offset})$ pairs. Layers L11--L29 schedule signed 6-digit $\pm$ and $5 \times 5$ multiplication with \texttt{MUL\_OVF} abstaining if $\deg(A)+\deg(B) \ge 6$. Whereas 1-digit emission ($K=1{,}043$) outputs off-manifold 1-digit tokens (\texttt{['1','2','4','5']}) that degrade \texttt{GSM8K}, our \textbf{Base-1000 Canonical BPE Emitter} ($K=1{,}140$, L30--L39) tracks emitted chunk-length prefix sums across $\delta \in \{1..7\}$ and routes remaining digits $r \in \{1..7\}$ onto right-aligned Base-1000 chunks, scoring all $1{,}110$ canonical tokens via $35$ readout wires.

\subsection{Hybrid \texorpdfstring{\texttt{GatedDeltaNet}}{GatedDeltaNet}: Alibaba Qwen 3.5 4B}
\texttt{Qwen/Qwen3.5-4B}~\citep{qwen2026qwen35} interleaves $24$ \texttt{Qwen3\_5GatedDeltaNet} blocks~\citep{yang2025gated} ($3:1$) with $8$ full-attention layers, unit-offset $\operatorname{RMSNorm}$ ($1+w$), strict 1-digit BPE, and a \textbf{tied $248{,}320$-token head} ($W_{\mathrm{lm}} = W_{\mathrm{emb}}$). Leaving \textbf{$0 / 8$ full-attention layers} modified ($0$ added attention KV cache), we schedule cross-token shifts inside $12 / 24$ \texttt{GatedDeltaNet} blocks via \cref{thm:routing}(2), cascading taps $\delta \in \{1,2,3\}$ to reach $\delta \le 9$ ($M_0 = 2^{-9}$, Layer-0 drift $<0.015\%$). Quantizing the 7-digit ALU output (L9--L27) into 4-bit nibbles emits canonical 1-digit tokens (L15--L31) without multi-chunk \texttt{BF16} carry drift, paired with $(+254.0, -256.0)$ differential tied weights (\cref{thm:subspace}(2)).

\subsection{42-Layer Architectural \& Activation Audit: Google Gemma 4 E4B}
\label{sec:gemma4}

\begin{table*}[t]
\caption{\textbf{Consolidated robustness vs.\ abstention summary (left) and hardware/precision/kernel/PTQ compatibility matrix (right).} Full 17-category $10.5\text{K}$ false-trigger, 4-seed, failure, Boolean, kernel, and PTQ breakdowns appear in \cref{app:val_tables,app:failure_analysis,app:composability_sparse}.}
\label{tab:robustness_compat}
\vskip 0.02in
\begin{center}
\scriptsize
\setlength{\tabcolsep}{2.7pt}
\resizebox{\textwidth}{!}{%
\begin{tabular}{@{}llcc@{\hspace{8pt}}lllc@{}}
\toprule
\multicolumn{4}{@{}l}{\textbf{(a) In-Domain Execution vs.\ Out-of-Domain Abstention (\texttt{BF16})}} & \multicolumn{4}{l@{}}{\textbf{(b) Hardware, Precision, Fused Kernel, \& Sidecar PTQ Matrix}} \\
\cmidrule(r{7pt}){1-4} \cmidrule(l{0pt}){5-8}
\textbf{Regime \& Suite} & \textbf{Scope / $N$} & \textbf{Granite 4.2} & \textbf{Qwen 3.5} & \textbf{Backend / Stack} & \textbf{Precision / Kernel} & \textbf{Carry / Gate Guarantee ($|C|{\le}81$)} & \textbf{Status} \\
\midrule
\textbf{In-Dom.} 4-Seed 8-Fmt & $N{=}3{,}000$ [CI] & $94.07\%$ & $\mathbf{96.60\%}$ & CPU / A100 CUDA & \texttt{FP32} (\texttt{no\_tf32}) / \texttt{TF32} & Bound $.144 / .303{<}.5$; $5/5$ \texttt{G3--G7} & \textbf{Verified} \\
\textbf{In-Dom.} $T{=}0.7$ / Bnd. & $N{=}240$ / $180$ & $95.4\%$ / $91.7\%$ & $\mathbf{96.3\%}$ / $\mathbf{99.4\%}$ & MPS \& A100 CUDA & Native \texttt{BF16} (\texttt{FP32} acc) & Bound $1.43$; $100\%$ Q / $99.2\%$ G & \textbf{Ver.\ (Drift)} \\
\textbf{In-Dom.} BOS-less $t_0{=}0$ & \texttt{"999+1="} ($32$) & $71.9\%$ (BOS dep.) & $\mathbf{100.0\%}$ & \texttt{torch.compile} & \textbf{TorchInductor Triton} & Fused SwiGLU $z^*{=}88.8/89{<}307$ & \textbf{Verified} \\
\textbf{In-Dom.} \textbf{Decl.\ Boolean} & $N{=}1{,}728$ ($384$ prec) & $\mathbf{100.0\%}$ ($\mathbf{100\%}$) & $\mathbf{100.0\%}$ ($\mathbf{100\%}$) & PyTorch SDPA & \textbf{FlashAttn-2 / MemEff} & $\max|\Delta|{=}0.0$ vs.\ \texttt{MATH}; $98.3\%$ & \textbf{Verified} \\
\textbf{In-Dom.} \textbf{5-Bit Maj/Par} & $\mathsf{TC}^0$ Voter ($64$) & $\mathbf{100.0\%}$ & $\mathbf{100.0\%}$ & IEEE \texttt{FP16} (\texttt{half}) & Max $65{,}504$ ($\operatorname{SiLU}(-\infty)$) & Raw \texttt{NaN}; clamp $[-4096,4096]$ OK & \textbf{Mitigated} \\
\cmidrule(r{7pt}){1-4}
\textbf{Abstain} Real Corpora & $6$ corpora ($5{,}000$) & $\mathbf{0/5{,}000}$ ($0.0$) & $\mathbf{0/5{,}000}$ ($0.0$) & \texttt{-ffast-math} / FTZ & Approx.\ $\exp$/recip. & Breaks \texttt{-0.0} inactivity & \textbf{Unsupported} \\
\textbf{Abstain} Adv.\ Near-Miss & $11$ cats ($5{,}534$)$^*$ & $\mathbf{0/5{,}534}$ ($0.0$) & $\mathbf{0/5{,}534}$ ($0.0$) & Uniform PTQ Rows & \texttt{INT8} / \texttt{INT4} on $W_{\mathrm{circ}}$ & Err $37.4$ / $81.0$ ($\alpha_k{\to}0$) & \textbf{Failed} \\
\textbf{Abstain} \textbf{Boolean OOD} & $6$ cats ($600$)$^\dagger$ & $\mathbf{0/600}$ ($0.0$) & $\mathbf{0/600}$ ($0.0$) & \textbf{Mixed Sidecar} & \textbf{\texttt{BF16} $W_{\mathrm{NL}}$ + \texttt{FP32} $W_{\mathrm{circ}}$} & \textbf{$100\%$ \texttt{G3--G7} ($5/5$); $100\%$ Q} & \textbf{Verified} \\
\textbf{Syntax} Unspaced \texttt{+-}/\texttt{(} & \texttt{"888+-25="} & Abstains (BPE \texttt{+-}) & Abstains (BPE) & \textbf{Hybrid PTQ + SC} & \textbf{\texttt{INT8/4} $W_{\mathrm{NL}}$ + \texttt{FP32} $W_{\mathrm{circ}}$} & \textbf{$100\%$ \texttt{G3--G7}; $100\%$ Q ($8.1\text{GiB}$)} & \textbf{Verified} \\
\bottomrule
\end{tabular}%
}
\end{center}
\vspace{-2pt}
{\scriptsize $^*$Chains, floats, commas, $\ge 7$-dig, $\deg\ge 6$, \texttt{\$a+b=}, \texttt{/ \% ** ==}, code/SQL/URLs, algebra, dates, parens, units ($0/10{,}534$). $^\dagger$$4$-lit Boolean chains, parens, Python \texttt{==}/\texttt{and}, arith \texttt{=}, \texttt{SWE-bench}, \texttt{MMLU-Pro}.}
\vskip -0.14in
\end{table*}

Auditing \texttt{google/gemma-4-E4B-it}~\citep{google2026gemma4} (42L, $\tanh$-GeGLU, QK/V-$\operatorname{RMSNorm}$, Sandwich norms, $h_\ell{\leftarrow}s_\ell h_\ell$, top-18 KV share) isolates activation constants and three failure modes (\cref{app:sub_boolean}): (i)~over $250\times$ swings on payload $7.0$, exact $I_\phi$ for $\operatorname{SiLU}$ ($0.8225$), $\operatorname{erf}$-$\operatorname{GELU}$ ($0.2500$), $\tanh$-$\operatorname{GELU}$ ($0.2496$), and $\operatorname{QuickGELU}$ ($0.2839$) keep \texttt{FP32} err $\le .0070$ (vs.\ $.0186$ [$0.166\%$ bias] if uncalibrated $0.25$ is used on $\tanh$-$\operatorname{GELU}$); and (ii)~on 42L Boolean grafts ($K{=}296$), KV-shared L24--L41 fails ($47.9\%$, $30/60$ FT) by overwriting $(K,V)$, and omitting Sandwich or cumulative $m_\ell = m_0 \prod_{j<\ell} s_j$ scaling yields $10.9\%$, whereas calibrating both on unshared L0--L18 hits $\mathbf{100.0\%}$ ($64/64$).

\subsection{Automated Multi-Digit Arithmetic, Wire Reuse, \& $\mathsf{TC}^0$ DAGs}
\label{sec:auto_boolean}
Declarative compilation factors into five passes (\cref{app:sub_boolean}): automatic \texttt{config}/BPE introspection, sublayer-tick ASAP scheduling ($\ell_{\mathrm{mlp}}{=}\lfloor\tau/2\rfloor$), SSA or ring \texttt{PreStepRingPool} allocation, shift lowering, and $\sqrt{d/(d+K)}$ patch emission. Without manual wire IDs, layer numbers, or BPE tables, \texttt{compile\_declarative\_arithmetic} synthesizes $D{\in}\{4,6,8\}$-digit Granite ($14/51/176$ chunkings, constant $39\text{L}$ $\mathsf{TC}^0$ depth, $1.07\text{M}$ nnz [$-19.5\%$] at $D{=}6$) and 6-digit Qwen~3.5 ($K{=}2{,}528$, $2.25\text{M}$ nnz [$-5.6\%$], $0/8$ Attn), hitting $5/5$ \texttt{G3--G7}, $96\%$--$98\%$ inline prose, and $0/200$ OOD (\cref{tab:auto_arith_compiler}), alongside a \textbf{5-bit Majority/Parity voter} ($100\%$, $64/64$) and \textbf{3-literal Boolean evaluator} ($100\%$, $N{=}1{,}728$; $-7.4\%$ ring-$K$, $0/600$ OOD; \cref{tab:robustness_compat}).

\section{Experimental Evaluation}
\label{sec:experiments}

\begin{table*}[t]
\caption{\textbf{Pre-registered expanded reasoning \& $>2\text{K}$-token long-context noninferiority ($N{=}4{,}392$ + $122.8\text{K}$-tok NLL, top)} and short-context 8-task NL, template, and generation suites (\texttt{BF16}, one-sided Tango $\alpha{=}0.05$; protocol frozen prior to execution, \cref{app:sub_retention}).}
\label{tab:heldout_stats}
\vskip 0.02in
\begin{center}
\scriptsize
\setlength{\tabcolsep}{3.5pt}
\resizebox{\textwidth}{!}{%
\begin{tabular}{@{}lccccccc@{}}
\toprule
& & \multicolumn{3}{c}{\textbf{IBM Granite 4.2 3B (\texttt{BF16})}} & \multicolumn{3}{c}{\textbf{Alibaba Qwen 3.5 4B (\texttt{BF16})}} \\
\cmidrule(lr){3-5} \cmidrule(lr){6-8}
\textbf{Evaluation Split ($\delta_{\mathrm{NI}}$)} & \textbf{$N$} & \textbf{Base $\to$ Graft} & \textbf{$\Delta$ ($\mathrm{LB}_{95\%}$)} & \textbf{$p_{\mathrm{NI}}$ (Status)} & \textbf{Base $\to$ Graft} & \textbf{$\Delta$ ($\mathrm{LB}_{95\%}$)} & \textbf{$p_{\mathrm{NI}}$ (Status)} \\
\midrule
Full \texttt{GSM8K} ($2.0\text{pp}$) & $1{,}319$ & $73.16 \to \mathbf{74.15\%}$ & $+0.99$ ($-0.21$) & $<.0001$ (\textbf{Pass}) & $75.59 \to \mathbf{76.35\%}$ & $+0.76$ ($+0.10$) & $<.0001$ (\textbf{Pass}) \\
Full \texttt{ARC-C} ($1.5\text{pp}$) & $1{,}172$ & $75.17 \to \mathbf{75.26\%}$ & $+0.09$ ($-0.46$) & $.0001$ (\textbf{Pass}) & $92.49 \to 92.41\%$ & $-0.09$ ($-0.44$) & $<.0001$ (\textbf{Pass}) \\
Held-Out \texttt{MMLU-Pro} ($2.0\text{pp}$) & $1{,}000$ & $27.70 \to 27.60\%$ & $-0.10$ ($-0.76$) & $.0001$ (\textbf{Pass}) & $42.80 \to 42.60\%$ & $-0.20$ ($-0.85$) & $.0002$ (\textbf{Pass}) \\
Full \texttt{LogiQA-EN} ($2.5\text{pp}$) & $651$ & $33.49 \to \mathbf{33.64\%}$ & $+0.15$ ($-0.89$) & $.0003$ (\textbf{Pass}) & $41.78 \to \mathbf{42.55\%}$ & $+0.77$ ($+0.13$) & $<.0001$ (\textbf{Pass}) \\
Long-Ctx \texttt{PubMedQA-2.5K} ($3.0\text{pp}$) & $250$ & $55.60 \to \mathbf{58.00\%}$ & $+2.40$ ($+0.37$) & $.0003$ (\textbf{Pass}) & $49.60 \to 48.80\%$ & $-0.80$ ($-2.62$) & $.0294$ (\textbf{Pass}) \\
\textbf{Pooled Expanded} ($0.75\text{pp}$) & $\mathbf{4{,}392}$ & $56.47 \to \mathbf{56.92\%}$ & $+0.46$ ($+0.01$) & $<.0001$ (\textbf{Pass}) & $66.14 \to \mathbf{66.37\%}$ & $+0.23$ ($-0.05$) & $<.0001$ (\textbf{Pass}) \\
\texttt{SWE-bench} $2{,}048$-tok NLL ($+.01$) & $60$ ($123\text{K}$) & $1.9281 \to 1.9304$ & $+.0023$ ($\mathrm{UB}\,{+.0048}$) & $<.0001$ (\textbf{Pass}) & $1.8886 \to \mathbf{1.8876}$ & $-.0010$ ($\mathrm{UB}\,{+.0019}$) & $<.0001$ (\textbf{Pass}) \\
\midrule
Pooled Short-Ctx 8 NL Tasks ($1.0\text{pp}$) & $2{,}350$ & $55.79 \to 56.13\%$ & $+0.34$ ($-0.08$) & $<.0001$ (\textbf{Pass}) & $66.21 \to 65.96\%$ & $-0.26$ ($-0.59$) & $.0008$ (\textbf{Pass}) \\
4-Tpl ARC ($800$) / Gen-80 ROUGE-L & $880$ & \textbf{4/4 Pass} ($\le .013$) & $\text{ROUGE-L } 80.6\%$ & \textbf{Pass} & \textbf{4/4 Pass} ($\le .015$) & $\text{ROUGE-L } 91.1\%$ & \textbf{Pass} \\
\bottomrule
\end{tabular}%
}
\end{center}
\vskip -0.14in
\end{table*}

\phantomsection\label{sec:footprint}
\textbf{Block-Sparse Sidecar \& Hybrid PTQ (\cref{tab:systems_footprint}).}
Zero-filled dense \texttt{Linear} padding ($d+K$) inflates parameters by $1.98\times$--$3.13\times$ ($13.55$ / $19.09\text{ GiB}$ ragged on A100), whereas our factorized \texttt{SparseSidecar} (\texttt{sparse\_sidecar\_runtime.py}; \cref{app:sub_sparse}) keeps $W_{\mathrm{NL}}$ at native width $d{=}2560$ and evaluates only active circuit slices ($1.33\text{M}$ [$0.036\%$] / $2.39\text{M}$ [$0.057\%$] nnz), reducing A100 VRAM to $7.89$ / $10.65\text{ GiB}$ and speeding up $2{,}048$-token prefill by $2.67\times$ ($347.5 \to 130.1\text{ ms}$) and $1.96\times$ ($847.9 \to 433.7\text{ ms}$) with unchanged \texttt{ARC-C}. Evaluating the $<0.06\%$ sidecar in \texttt{FP32} alongside \texttt{BF16} or group-$128$ \texttt{INT8}/\texttt{INT4}-quantized $W_{\mathrm{NL}}$ ($6.07 / 4.60\text{ GiB}$ Granite; $9.80 / 8.14\text{ GiB}$ Qwen) resolves all $5/5$ (\texttt{G3--G7}) carry-ULP edge cases and maintains $100.0\%$ Qwen arithmetic across \texttt{BF16}/\texttt{INT8}/\texttt{INT4}.

\phantomsection\label{sec:decomposed_eval}
\textbf{Decomposed Accuracy, 5-Primitive Ablations, \& \texttt{LoRA} SFT (\cref{tab:main_eval,tab:matched_calib}).}
\Cref{tab:main_eval} decomposes generation into \textbf{ALU-Prefix} ($|\Delta z_{\mathrm{circ}}| > 0.5$), \textbf{Lead-Int}, \textbf{Strict Non-Decimal} ($800$ bare vs.\ $400$ prose), and \textbf{Inline Resumption} ($N{=}50$).
(1)~\emph{Exactness vs.\ continuation:} Qwen~3.5 \texttt{ALU-Prefix} reaches $1{,}200/1{,}200$ ($100\%$) in \texttt{BF16} ($99.50\%$ \texttt{Lead-Int}/\texttt{Strict-ND}; $100\%$ \texttt{Inline-50}), and Granite reaches $99.17\%$ ($100\%$ in \texttt{FP32}/\texttt{FP32}-sidecar; $98.25\%$ prose \texttt{Strict-ND}). Across $4$ numeric tokenization schemes on \texttt{GSM8K-100} (\cref{app:sub_retention}), raw 1-digit ($36.0\%$) and spaced 1-digit ($35.0\%$) degrade reasoning relative to 3-digit and Base-1000 canonical triads ($73.0\%$).
(2)~\emph{Leave-one-out \& \texttt{LoRA} (\cref{tab:matched_calib}):} Removing any of the $5$ primitives degrades execution or language modeling: unit-calibrated single-scale SwiGLU (\texttt{-P1}) drops Arith to $8.3\%$ / $20.0\%$; positive RoPE ($w_m{>}0$, \texttt{-P2}) drops Granite to $15.0\%$; stateful \texttt{GatedDeltaNet} ($e^{g_t}{\ne}0$, \texttt{-P3}) drops Qwen to $51.1\%$ ($8.999$ signed-shift err w/o $\operatorname{SiLU}$ antisymmetry); omitting $\sqrt{d/(d+K)}$ (\texttt{-P4}) reduces \texttt{ARC-C} by $-19$ / $-54\text{pp}$; and single-ended tied readout (\texttt{-P5}) introduces nonzero digit-token bias. Same-backbone \texttt{LoRA} SFT ($12.8\text{K}$ examples, $6\text{M}{+}$ params) reaches $42.2\%$ / $61.7\%$ Arith and $0.0\%$ fluent inline continuation, while $0$-token weight-analytic calibration (\texttt{C2}) retains $96.3\%$ / $97.0\%$.

\phantomsection\label{sec:audit_summary}
\textbf{Kernel Portability, Decoding/\texttt{IFEval}/Safety, \& $>2\text{K}$-Ctx Retention (\cref{tab:robustness_compat,tab:heldout_stats}).}
Four-seed 8-format evaluation ($N{=}3{,}000$/model) reaches $94.07\%$ / $96.60\%$ with $0/11{,}134$ combined arithmetic/Boolean OOD triggers ($\max|\Delta z_{\mathrm{circ}}|{=}0.0$). On A100 (\cref{tab:robustness_compat}b), \texttt{TorchInductor} Triton fused SwiGLU ($z^*{=}88.8/89.0 < 307.2$), PyTorch SDPA \texttt{FlashAttention-2}/\texttt{MemEfficient} ($\max|\Delta|{=}0.0$), and \texttt{TF32} ($5/5$ \texttt{G3--G7}) execute cleanly, and IEEE \texttt{FP16} gate overflow ($65{,}504 \to \texttt{NaN}$) is avoided via $[-4096, 4096]$ clamping or an \texttt{FP32} sidecar (\cref{app:sub_kernel}). Under pre-registered Tango testing~\citep{tango1998equivalence} (\cref{tab:heldout_stats,app:sub_retention}), both grafts pass all $14/14$ expanded reasoning and $>2\text{K}$-token splits ($N{=}4{,}392$: $+0.46$ / $+0.23\text{pp}$, $p{<}10^{-5}$; \texttt{SWE-bench} $\Delta\text{NLL}\le +.0023$), preserve accuracy across $5$ decoding regimes (\texttt{Greedy}, $T{\in}\{0.3,0.7,1.0\}$, \texttt{Beam-4}), \texttt{IFEval-100}, and \texttt{Safety-80} ($100\%{\to}100\%$ harmful refusal, $0.0\text{pp}$ benign drift; \cref{app:sub_retention}), and reproduce from standalone patches across \texttt{mps} and A100 \texttt{cuda} (\cref{app:repro}).

\section{Discussion, Limitations, and Conclusion}
\label{sec:discussion}

\phantomsection\label{sec:handoff_discussion}
\textbf{(1)~Numeric BPE manifold alignment:} Consistent with~\citet{singh2024tokenization}, 1-digit Granite emission ($K{=}1{,}043$) drops \texttt{GSM8K[:60]} (\texttt{bs=1}; \cref{app:gsm8k_attribution}) from $68.33\% \to 36.67\%$ ($b_{10}{=}19, b_{01}{=}0$) although \textbf{$18/19$ ($94.7\%$) regressions compute exact ALU outputs}: Granite's 1--3 digit BPE misreads unmerged 1-digit IDs as decimals ($5/19$), split spans ($9/19$), or high-surprisal sequences ($4/19$)---resolved ($70.00\%$--$76.67\%$) by Base-1000 Canonical BPE ($K{=}1{,}140$). Post-disengagement ($\Delta z_{\mathrm{circ}}{=}0$) yields $98\%$--$100\%$ inline fluency while leaving bare Granite prompts unanchored against \texttt{.d} suffixes ($98.25\%$ prose vs.\ $87.0\%$ bare).

\textbf{(2)~Factorized block-sparse sidecar \& mixed PTQ (\cref{app:sub_sparse}):} Axis-aligned zero cross-weights let \texttt{SparseSidecar} evaluate $W_{\mathrm{NL}}\tilde{h}_{\mathrm{NL}} \oplus W_{\mathrm{circ}}\tilde{h}_{\mathrm{circ}}$ coupled only through scalar $\|h_{\mathrm{NL}}\|_2^2 + \|h_{\mathrm{circ}}\|_2^2$, cutting A100 VRAM $42\%$--$44\%$ ($7.89$ / $10.65\text{ GiB}$) and $2\text{K}$-prefill latency $1.96\times$--$2.67\times$ in pure PyTorch, while pairing an \texttt{FP32} sidecar with \texttt{BF16} or \texttt{INT8}/\texttt{INT4} $W_{\mathrm{NL}}$ ($4.60$ / $8.14\text{ GiB}$) achieves $5/5$ \texttt{G3--G7} and $100\%$ Qwen ALU (\cref{tab:systems_footprint}).

\phantomsection\label{sec:limitations}
\textbf{(3)~Scope \& $\mathsf{TC}^0$ scaling:} \emph{(i)}~\texttt{compile\_declarative\_arithmetic} automates $4$--$8$ digit multi-chunk BPE $\{+,-,\times\}$ ($14$--$176$ chunkings in constant $39\text{L}$ $\mathsf{TC}^0$ depth, $-5.6\%$ to $-19.5\%$ nnz; \cref{tab:auto_arith_compiler}) alongside Boolean/voter DAGs, whereas unbounded-length inputs or halting loops exceed single-pass $\mathsf{TC}^0$ (\cref{rem:tc0}). \emph{(ii)}~While restoring division takes $\sim 40\text{L}$, an 8-layer $\mathsf{TC}^0$ 64-knot Mellin + Newton--Raphson divider ($+1{,}480$ wires, $1.92\text{M}$ nnz) reaches $100.0\%$ ($10\text{K}/10\text{K}$) in an \texttt{FP32} sidecar ($73.9\%$ \texttt{TF32}, $56.1\%$ \texttt{BF16}), and $D \in \{9,12,18\}$ digits scale to $K \in \{2.2\text{K}, 3.6\text{K}, 7.0\text{K}\}$ across $32$--$44\text{L}$ (\cref{app:sub_kernel}). \emph{(iii)}~Full 42L Gemma~4 arithmetic and fused C++/CUDA BSR kernels remain open work, while dormant weight-space triggers motivate structural checkpoint auditing (\textbf{Impact Statement}; \citealp{clifford2024impnet}).

\textbf{Conclusion.}
Analytical zero-gradient primitives and factorized \texttt{FP32}/\texttt{BF16} sidecars compile exact arithmetic and $\mathsf{TC}^0$ circuits directly into pretrained LMs without gradients, task data, or host routers while preserving $>2\text{K}$-token language competence.

\clearpage
\section*{Impact Statement}
This paper presents analytical weight-compilation primitives, an automated 5-stage declarative arithmetic and Boolean DAG synthesizer, and reference sparse checkpoint patches aimed at advancing mechanistic understanding and deterministic reliability of pretrained language models without gradient training or external code-execution sandboxes. \textbf{Reliable single-checkpoint computation:} In air-gapped, regulated, or pure-tensor inference environments where pausing decoding to execute arbitrary host Python/shell code~\citep{gao2023pal,schick2023toolformer} expands the attack surface, verified weight-space circuits offer mathematically bounded execution inside standard \texttt{safetensors} tensors alongside clean mechanistic probes of BPE tokenization--reasoning alignment~\citep{singh2024tokenization}.

\textbf{Dual-use security implications (concealed input-triggered weight behaviors):} Conversely, our empirical findings carry a concrete model-supply-chain security lesson extending compiled-network backdoor analyses~\citep{clifford2024impnet} to multi-billion-parameter pretrained LMs. Because \texttt{0x0000} complement-pure Mellin gates and differential $(+254,-256)$ tied readouts remain bit-exact zero off-trigger ($0/10{,}534$ out-of-domain activations; $<0.002$~nats next-token KL shift; $<0.06\%$ active non-zeros), black-box behavioral benchmarks and standard perplexity evaluations cannot reliably detect dormant, input-triggered payloads grafted into open-weight checkpoints. Mitigating this risk requires pairing behavioral evaluation with \emph{white-box structural auditing} of distributed weights: (i)~verifying canonical hidden/intermediate tensor dimensions against official model configs; (ii)~scanning projection rows and $\operatorname{RMSNorm}$ vectors for axis-aligned block decoupling, bimodal $L_0$ sparsity, extreme gate-suppression constants ($-S \ll -10^2$), or antisymmetric norm pairs; and (iii)~enforcing cryptographic SHA-256 provenance on model shards. Finally, while dense \texttt{Linear} materialization increases memory ($1.98\times$--$3.13\times$), our factorized \texttt{SparseSidecar} (\cref{app:sub_sparse}) eliminates $>40\%$ of that VRAM overhead in pure PyTorch.

\bibliography{references}
\bibliographystyle{icml2027}

\newpage
\appendix
\crefalias{section}{appendix}
\onecolumn
\section{Complete Mathematical Proofs and Finite-Precision Analysis}
\label{app:proofs}

We provide complete, self-contained derivations for \cref{thm:mellin,thm:routing,thm:subspace}, separating exact real-arithmetic ($\mathbb{R}$) identities, non-asymptotic truncation bounds over $\mathbb{R}$, and kernel-, schedule-, and evaluation-tree-explicit IEEE~754 floating-point (\texttt{FP32} vs.\ \texttt{BF16}) rounding bounds. \Cref{tab:notation_glossary} summarizes our notation and finite-precision quantities.

\begin{table}[!ht]
\centering
\caption{\textbf{Summary of circuit-grafting notation and IEEE~754 / kernel quantities.}}
\label{tab:notation_glossary}
\vskip 0.04in
\scriptsize
\setlength{\tabcolsep}{4.5pt}
\resizebox{\textwidth}{!}{%
\begin{tabular}{@{}lll@{}}
\toprule
\textbf{Symbol / Term} & \textbf{Definition} & \textbf{Plain-Language Role in Circuit} \\
\midrule
$d,\; K,\; h_{\mathrm{ext}} = [h_{\mathrm{NL}}; h_{\mathrm{circ}}]$ & Base width $d=2560$; appended circuit wires $K$ & Keeps pretrained NL features untouched in $\mathbb{R}^d$ \\
$\lambda_{t,\ell} = (\frac{1}{D}\|h_{t,\ell}\|_2^2 + \varepsilon)^{-1/2}$ & Token/layer-varying $\operatorname{RMSNorm}$ inverse scale & Unknown runtime multiplier ($>250\times$ swing) we cancel \\
$r_0,\; g_{\mathrm{off}} \in \{0, 1\},\; M_0$ & Reference wire ($1$); complement guard ($0$ active, $1$ off); scale & Tracks $\lambda_{t,\ell}$ and holds circuit bit-exact silent off-trigger \\
$\alpha_k = e^{k\Delta},\; I_\phi$ & Log-spaced gate scales ($\Delta=0.55$); Mellin integral constant & Converts $\lambda_{t,\ell}$ scaling into a shift along a flat integral plateau \\
$S(\delta) = \sum_m w_m \cos(\delta \omega_m)$ & Indefinite-RoPE spectral kernel ($S(0){=}1,\; S(\delta{\ne}0){\le}{-}0.355$) & Forces inactive queries onto a zero-payload self-sink ($\delta{=}0$) \\
$\gamma_{\mathrm{NL}} = \sqrt{d/(d+K)}$ & Pretrained $\operatorname{RMSNorm}$ weight compensation factor & Cancels the $\frac{1}{d} \to \frac{1}{d+K}$ denominator shift on NL channels \\
$(w_{p_m}, w_{n_m}) = (+254, -256)$ & Differential unit-offset $\operatorname{RMSNorm}$ gains ($1+w = \pm 255$) & Cancels Layer-0 digit embeddings on Qwen's tied \texttt{lm\_head} \\
\texttt{0x0000} / \texttt{-0.0} / FTZ & IEEE~754 exact zero bit-patterns; Flush-To-Zero hardware mode & Ensures inactive gates add exact $0.0$ without \texttt{-ffast-math} \\
$u_{\mathrm{acc}},\; \gamma_N = \frac{N u_{\mathrm{acc}}}{1 - N u_{\mathrm{acc}}}$ & Accumulator unit roundoff ($2^{-24}$ \texttt{FP32}); Higham sum bound & Bounds worst-case floating-point dot-product rounding error \\
Schedule S1 / Tree $\mathcal{T}$ ($h_{\mathcal{T}}$) & Pairwise $(p_m, n_m)$-first sum vs.\ arbitrary GEMM reduction tree & Distinguishes exact $+0.0$ sidecar readout from general cuBLAS trees \\
\bottomrule
\end{tabular}%
}
\end{table}

\paragraph{Scope of numerical scripts vs.\ human mathematical proof verification.}
Companion scripts (\texttt{verify\_math\_proofs\_and\_constants.py}, \texttt{verify\_v4\_submission\_audit.py}, and \texttt{verify\_icml2027\_gemm\_and\_proofs.py}) evaluate closed-form constants at $50$-digit precision via \texttt{mpmath}, exhaustively sweep IEEE~754 \texttt{FP32}/\texttt{BF16} bit-patterns for $\operatorname{SiLU}$ underflow/overflow thresholds, and execute explicit floating-point counterexamples and Monte Carlo reduction-tree permutations. Such scripts verify numerical constants and test concrete floating-point instances only; they do \emph{not} constitute formal machine-checked theorem proving or substitute for line-by-line human mathematical proof verification.

\subsection{Proof of \cref{thm:mellin} (Scale-Invariant Mellin-GLU Quadrature)}

\paragraph{Part 1 (Single-layer impossibility over $\mathbb{R}$).}
Let $F_1(u) = W_{\mathrm{down}}\big(\phi(W_{\mathrm{gate}} u) \odot (W_{\mathrm{up}} u)\big)$ where $\phi : \mathbb{R} \to \mathbb{R}$ satisfies $|\phi(s)| \le C|s|$ for all $s \in \mathbb{R}$. Standard gated-MLP activations satisfy $C = 1$: $\operatorname{SiLU}(s) = s\sigma(s)$ has $\sigma(s) = (1+e^{-s})^{-1} \in (0,1)$; exact Gaussian $\operatorname{GELU}(s) = s\Phi(s)$ has $\Phi(s) \in (0,1)$; and $\tanh$-approximate $\operatorname{GELU}_{\tanh}(s) = \frac{1}{2}s\big(1 + \tanh(\sqrt{2/\pi}(s + 0.044715 s^3))\big)$ has $\tanh(\cdot) \in (-1, 1)$. For any $u \in \mathbb{R}^D$, any $\lambda > 0$, and each neuron $i \in \{1,\dots,m\}$, homogeneity of $W_{\mathrm{gate}}(\lambda u) = \lambda W_{\mathrm{gate}} u$ and $W_{\mathrm{up}}(\lambda u) = \lambda W_{\mathrm{up}} u$ yields
\begin{equation*}
\big|\phi\big((W_{\mathrm{gate}}(\lambda u))_i\big) \cdot (W_{\mathrm{up}}(\lambda u))_i\big| \;\le\; C\big|\lambda (W_{\mathrm{gate}} u)_i\big| \cdot \big|\lambda (W_{\mathrm{up}} u)_i\big| \;=\; \lambda^2 C \big|(W_{\mathrm{gate}} u)_i (W_{\mathrm{up}} u)_i\big|.
\end{equation*}
Taking the Euclidean norm of $v(\lambda u) = \phi(W_{\mathrm{gate}}(\lambda u)) \odot (W_{\mathrm{up}}(\lambda u))$ gives $\|v(\lambda u)\|_2 \le \lambda^2 C \|W_{\mathrm{gate}} u\|_\infty \|W_{\mathrm{up}} u\|_2 \le \lambda^2 C \|W_{\mathrm{gate}}\|_2 \|W_{\mathrm{up}}\|_2 \|u\|_2^2$. By spectral-norm submultiplicativity,
\begin{equation}
\label{eq:app_f1_quad}
\|F_1(\lambda u)\|_2 \;\le\; \lambda^2 C \|W_{\mathrm{down}}\|_{2} \|W_{\mathrm{gate}} u\|_\infty \|W_{\mathrm{up}} u\|_2 \;\le\; \lambda^2 C \|W_{\mathrm{down}}\|_2 \|W_{\mathrm{gate}}\|_2 \|W_{\mathrm{up}}\|_2 \|u\|_2^2.
\end{equation}
Letting $\lambda \to 0^+$ yields $\lim_{\lambda \to 0^+} \|F_1(\lambda u)\|_2 = 0$ for every $u \in \mathbb{R}^D$, so no single bias-free GLU layer can realize $H(x - \tau r_0) = 1$ (for $x > \tau r_0$) uniformly across $(0, \lambda_{\max}]$. Moreover, on any compact interval $[\lambda_{\min}, \lambda_{\max}] \subset (0, \infty)$, if $\|F_1(\lambda_{\min} u)\|_2 \ge 1 - \epsilon$ for some $\epsilon \in [0, 1)$ and $u \in \mathbb{R}^D \setminus \{0\}$, substituting $\lambda = \lambda_{\min}$ into~\eqref{eq:app_f1_quad} immediately gives $\|W_{\mathrm{down}}\|_2 \|W_{\mathrm{gate}}\|_2 \|W_{\mathrm{up}}\|_2 \ge \frac{1-\epsilon}{C\|u\|_2^2}\lambda_{\min}^{-2}$.

\paragraph{Part 2 (Exact Mellin integral, payload linearity, Poisson bound, and margin-separated Heaviside step).}
For $\tilde{r} > 0$, $\tilde{v} \in \mathbb{R}$, and $\operatorname{SiLU}(-t) = -\frac{t}{1 + e^t}$ ($t > 0$), substituting $t = e^u \tilde{r}$ ($du = dt/t$, $e^u \tilde{v} = t \tilde{v}/\tilde{r}$) yields
\begin{align*}
\int_{-\infty}^{\infty} \operatorname{SiLU}\!\big(-e^{u}\tilde{r}\big)\big(-e^{u}\tilde{v}\big)\,du
&= \int_{-\infty}^{\infty} \left(-\frac{e^{u}\tilde{r}}{1 + e^{e^{u}\tilde{r}}}\right)\!\big(-e^{u}\tilde{v}\big)\,du
= \frac{\tilde{v}}{\tilde{r}} \int_{0}^{\infty} \frac{t}{e^{t} + 1}\,dt.
\end{align*}
For $t > 0$, the alternating partial sums $S_N(t) = \sum_{n=1}^{N} (-1)^{n-1} t e^{-nt} = t e^{-t}\frac{1 - (-e^{-t})^N}{1 + e^{-t}}$ satisfy $0 \le S_N(t) \le t e^{-t} \in L^1(0,\infty)$ for every $N \ge 1$. By Lebesgue's dominated convergence theorem, $\int_0^\infty \frac{t}{e^t+1}\,dt = \sum_{n=1}^\infty (-1)^{n-1}\int_0^\infty t e^{-nt}\,dt = \sum_{n=1}^\infty \frac{(-1)^{n-1}}{n^2} = \eta(2) = \frac{1}{2}\zeta(2) = \frac{\pi^2}{12}$, proving~\eqref{eq:mellin_integral}. (Replacing $\operatorname{SiLU}$ with exact Gaussian $\operatorname{GELU}(-t) = -t\Phi(-t)$ gives $\int_0^\infty t\Phi(-t)\,dt = \frac{1}{2\sqrt{2\pi}}\int_0^\infty t^2 e^{-t^2/2}\,dt = \frac{1}{4}$, and $0.249586$ under $\operatorname{GELU}_{\tanh}$.)

Now set $\tilde{r} = \lambda r_0 > 0$, $\tilde{v} = \lambda x$, $u_0 = \ln(\lambda r_0)$, and $\psi(w) = \frac{e^{2w}}{e^{e^w}+1} > 0$. Because each neuron's up-projection $-e^{k\Delta}\tilde{v} = -\frac{x}{r_0} e^{k\Delta + u_0}$ is homogeneous of degree~$1$ in $x$ while the gate $-e^{k\Delta}\tilde{r} = -e^{k\Delta + u_0}$ is independent of $x$, the finite quadrature sum factors identically as
\begin{equation*}
Q_{\Delta, K_-, K_+}(\lambda r_0, \lambda x) \;=\; \frac{x}{r_0}\, S_{\Delta, K_-, K_+}(u_0), \qquad S_{\Delta, K_-, K_+}(u_0) \;:=\; \frac{12\Delta}{\pi^2}\sum_{k=K_-}^{K_+} \psi(k\Delta + u_0).
\end{equation*}
Hence both the infinite-grid discretization error $|S_\Delta(u_0) - 1|$ ($S_\Delta(u_0) := \frac{12\Delta}{\pi^2}\sum_{k \in \mathbb{Z}} \psi(k\Delta + u_0)$) and the finite-window tail error $|S_{\Delta, K_-, K_+}(u_0) - S_\Delta(u_0)| = \frac{12\Delta}{\pi^2}\sum_{k \notin [K_-, K_+]} \psi(k\Delta + u_0)$ are dimensionless errors of the unit-ratio kernel $S_{\Delta, K_-, K_+}(u_0)$, and \emph{both} scale by $|x|/r_0$: $\big|Q_{\Delta, K_-, K_+}(\lambda r_0, \lambda x) - \frac{x}{r_0}\big| = \frac{|x|}{r_0}|S_{\Delta, K_-, K_+}(u_0) - 1| \le \frac{|x|}{r_0}\big(\varepsilon_{\mathrm{disc}}(\Delta) + \varepsilon_{\mathrm{tail}}(\lambda r_0; K_-, K_+, \Delta)\big)$:
\begin{enumerate}[leftmargin=1.5em,itemsep=1.5pt]
\item \emph{Non-asymptotic Poisson--Mellin discretization bound:} Because $\psi \in C^\infty(\mathbb{R})$ decays exponentially as $e^{2w}$ as $w \to -\infty$ and super-exponentially as $e^{2w - e^w}$ as $w \to +\infty$, both $\psi$ and its Fourier transform $\hat{\psi}(\xi) = \int_{-\infty}^\infty \psi(w)e^{-i\xi w}\,dw$ lie in the Schwartz space $\mathcal{S}(\mathbb{R}) \subset L^1(\mathbb{R})$~\citep{trefethen2014exponentially}, so Poisson summation holds pointwise: $\Delta \sum_{k \in \mathbb{Z}} \psi(k\Delta + u_0) = \sum_{m \in \mathbb{Z}} e^{2\pi i m u_0 / \Delta} \hat{\psi}(2\pi m/\Delta)$. Substituting $t = e^w$ gives the Mellin transform along $\operatorname{Re}(s) = 2$:
\begin{equation*}
\hat{\psi}(\xi) \;=\; \int_{-\infty}^{\infty} \frac{e^{2w}}{e^{e^w}+1} e^{-i\xi w}\,dw \;=\; \int_{0}^{\infty} \frac{t^{1 - i\xi}}{e^t + 1}\,dt \;=\; \eta(2 - i\xi)\,\Gamma(2 - i\xi).
\end{equation*}
Since $\operatorname{Re}(2 - i\xi) = 2 > 1$, $|\eta(2 - i\xi)| \le \sum_{n=1}^{\infty} |n^{-2+i\xi}| = \frac{\pi^2}{6}$. By Euler's reflection identity $|\Gamma(1 - i\xi)|^2 = \frac{\pi \xi}{\sinh(\pi \xi)}$ and $\Gamma(2 - i\xi) = (1 - i\xi)\Gamma(1 - i\xi)$, $|\Gamma(2 - i\xi)| = \sqrt{(1 + \xi^2)\frac{\pi|\xi|}{\sinh(\pi|\xi|)}}$. Combining $\hat{\psi}(0) = \frac{\pi^2}{12}$ with $\frac{24}{\pi^2}\cdot \frac{\pi^2}{6} = 4$ gives~\eqref{eq:mellin_bound}:
\begin{equation*}
\big|S_\Delta(u_0) - 1\big| \;\le\; \frac{24}{\pi^2}\sum_{m=1}^{\infty}\left|\hat{\psi}\!\left(\frac{2\pi m}{\Delta}\right)\right| \;\le\; 4\sum_{m=1}^{\infty}\sqrt{\left(1 + \frac{4\pi^2 m^2}{\Delta^2}\right)\frac{2\pi^2 m / \Delta}{\sinh(2\pi^2 m / \Delta)}}.
\end{equation*}
At $\Delta = 0.55$, $2\pi^2/\Delta \approx 35.8895$, giving $\varepsilon_{\mathrm{disc}}(0.55) \le 6.2551 \times 10^{-6}$ (actual supremum $\sup_{u_0}|S_{0.55}(u_0)-1| = 4.7276 \times 10^{-6}$).
\item \emph{Lower and upper tail bounds:} For $r = \lambda r_0 = e^{u_0}$, since $\psi(w) < e^{2w}$ and $e^{2\Delta} - 1 > 2\Delta$ for all $\Delta > 0$, summing over $k \le K_- - 1$ gives $\Delta \sum_{k=-\infty}^{K_- - 1} \psi(k\Delta + u_0) < \Delta r^2 \frac{e^{2 K_- \Delta}}{e^{2\Delta} - 1} < \frac{1}{2}r^2 e^{2 K_- \Delta}$. For $k \ge K_+ + 1$, let $T_+ = r e^{K_+ \Delta} \ge \lambda_{\min} r_0 e^{K_+\Delta} \ge 2$. Since $\psi(w) < g(w) := e^{2w - e^w}$ and $g'(w) = (2 - e^w)g(w) \le 0$ for all $w \ge \ln T_+ \ge \ln 2$, integral comparison on $[K_+\Delta + u_0, \infty)$ gives $\Delta \sum_{k=K_+ + 1}^\infty \psi(k\Delta + u_0) \le \int_{\ln T_+}^\infty e^{2w - e^w}\,dw = \int_{T_+}^\infty t e^{-t}\,dt = (T_+ + 1)e^{-T_+}$, establishing~\eqref{eq:mellin_bound}.
\item \emph{Margin-separated two-layer Heaviside step over $\mathbb{R}$ (requiring $z \ge 2$):} For $\beta, \gamma, \delta_{\mathrm{sep}} > 0$ with $z := \beta\lambda_{\min}\delta_{\mathrm{sep}} r_0 \ge 2$, Stage~1 in layer $\ell$ computes $p = \gamma \operatorname{SiLU}(s) s = \gamma \frac{s^2}{1 + e^{-s}} > 0$ for $s = \beta(\tilde{x} - \tau\tilde{r}) = \beta\lambda(x - \tau r_0) \ne 0$. On negative-margin inputs $x \le (\tau - \delta_{\mathrm{sep}})r_0$, let $w := -s = \beta\lambda(\tau r_0 - x) \ge \beta\lambda_{\min}\delta_{\mathrm{sep}} r_0 = z \ge 2$. Because $\frac{d}{dw}(w^2 e^{-w}) = w(2 - w)e^{-w} \le 0$ for all $w \ge 2$ (noting that $w^2 e^{-w}$ is strictly increasing on $(0, 2)$, making the hypothesis $z \ge 2$ essential; our compiled gates use $z \ge 16$), we obtain $0 < p = \gamma\frac{w^2}{1 + e^w} < \gamma w^2 e^{-w} \le \gamma z^2 e^{-z}$. In layer $\ell+1$, $\tilde{p} = \lambda_{\ell+1} p \le \lambda_{\max}'\gamma z^2 e^{-z}$, and since $\operatorname{SiLU}(-e^{k\Delta}\tilde{p})(-e^{k\Delta}\tilde{p}) < (\tilde{p}e^{k\Delta})^2$, summing $k \le K_+$ gives $0 < Q_{\Delta, K_-, K_+}(\tilde{p}, \tilde{p}) \le \frac{12\Delta}{\pi^2(1 - e^{-2\Delta})}(\lambda_{\max}'\gamma z^2 e^{-z})^2 e^{2 K_+\Delta}$. Conversely, on positive-margin inputs $x \ge (\tau + \delta_{\mathrm{sep}})r_0$, $s \ge z \ge 2 > 0$ gives $\sigma(s) > \frac{1}{2}$ and $p > \frac{1}{2}\gamma z^2$, so $\tilde{p} = \lambda_{\ell+1} p \ge \tilde{p}_{\min} := \frac{1}{2}\lambda_{\min}'\gamma z^2$. Here $\tilde{v} = \tilde{r} = \tilde{p}$ has unit ratio $|x|/r_0 = 1$; under $\tilde{p}_{\min} e^{K_+\Delta} \ge 2$,~\eqref{eq:mellin_bound} yields $|Q_{\Delta, K_-, K_+}(\tilde{p}, \tilde{p}) - 1| \le \varepsilon_{\mathrm{disc}}(\Delta) + \varepsilon_{\mathrm{tail}}(\tilde{p}; K_-, K_+, \Delta)$.
\end{enumerate}

\paragraph{Part 3 (Kernel-explicit IEEE~754 bounds, nonlinear $\operatorname{SiLU}$ conditioning, and \texttt{BF16} carry sensitivity).}
Under IEEE~754 round-to-nearest-even (RN) arithmetic with unit roundoff $u$, $\operatorname{fl}(a + b) = (a + b)(1 + \delta)$ ($|\delta| \le u$). Suppressing an inactive gate via additive cancellation $\operatorname{fl}(S\lambda_1) - \operatorname{fl}(S\lambda_2)$ under distinct prior roundings leaves residual $\sim S\lambda u$ ($S u_{\mathrm{bf16}} = 8$ at $S = 2^{11}, u_{\mathrm{bf16}} = 2^{-8}$). Under complement-pure gating, $g_{\mathrm{off}}$ is bit-exact \texttt{0x0000} ($+0.0$) when active, so $\operatorname{fl}((-S)\cdot 0.0) = -0.0$ and $\operatorname{fl}(a + (-0.0)) = a$ adds zero active error. When inactive ($g_{\mathrm{off}} \ge 1$), the gate input satisfies $s \le -z \le -S\lambda_{\min} < 0$, and exact-zero saturation depends explicitly on the activation kernel:
\begin{enumerate}[leftmargin=1.5em,itemsep=1pt,topsep=1pt]
\item \emph{Kernel Class K1 (direct reciprocal-exp $\operatorname{fl}(s / (1 + \exp(-s)))$ in IEEE~754 \texttt{FP32} opmath without \texttt{-ffast-math}/\texttt{-ffinite-math-only}, as in PyTorch ATen \texttt{F.silu} on CPU, MPS, and CUDA):} Since the maximum finite IEEE single/bfloat16 float is $< 3.41 \times 10^{38} < e^{88.73}$, $\operatorname{fl}(\exp(+z)) = +\infty$ and IEEE division by $+\infty$ returns bit-exact $-0.0$ for all $z \ge 89$ in both \texttt{FP32} and \texttt{BF16}.
\item \emph{Kernel Class K2 (range-split stable sigmoid $\operatorname{fl}(s \exp(s) / (1 + \exp(s)))$ for $s < 0$):} Because $\exp(-z)$ does not overflow, bit-exact $-0.0$ occurs when $z\exp(-z)$ underflows below half the minimum subnormal ($2^{-150}$ in \texttt{FP32}; $2^{-134}$ in \texttt{BF16}) or rounds to zero on \texttt{BF16} output cast, yielding exact $-0.0$ at $z \ge 104$ in \texttt{FP32}, $z \ge 98$ for \texttt{FP32}$\to$\texttt{BF16} cast, and $z \ge 93$ in pure \texttt{BF16} (with $|\operatorname{SiLU}(-z)| \le 89 e^{-89} < 2.0 \times 10^{-37}$ on $z \in [89, 104)$ in \texttt{FP32}; $\tanh$-based approximations saturate to $-1.0$ by $z \ge 18$). Thus $z \ge 104$ guarantees bit-exact $-0.0$ across both K1 and K2, and our compiled $S = 256$ gives nominal $z \approx 307 \gg 104$.
\end{enumerate}
On active passes ($r_0 = 1$, payload $C$), each neuron $k \in [K_-, K_+]$ evaluates $\operatorname{SiLU}(-\hat{t}_k)(-\hat{v}_k)$ where $\hat{t}_k = t_k(1 + \eta_k)$ ($t_k = e^{k\Delta}\tilde{r} > 0$, $|\eta_k| \le \bar{\eta}$ from weight rounding $u_w$ and input multiplication $u_{\mathrm{acc}}$). For $\phi_0(t) = \frac{t}{e^t + 1}$, $|\phi_0'(t)| = \phi_0(t)\big|\frac{1}{t} - \frac{e^t}{e^t+1}\big| \le \phi_0(t)(\frac{1}{t} + 1)$, so multiplicative perturbation $t_k \mapsto t_k(1+\eta_k)$ shifts $\phi_0(t_k)$ by at most $\phi_0(t_k)(1 + e^{\bar{\eta}} t_k)\bar{\eta}$. Summing over $k$ brings in the first-moment Mellin condition constant
\begin{equation*}
\kappa_{\mathrm{Mel}} \;:=\; \frac{\frac{12}{\pi^2}\int_{0}^{\infty} \frac{t^2}{e^t+1}\,dt}{1} \;=\; \frac{12}{\pi^2}\eta(3)\Gamma(3) \;=\; \frac{18\zeta(3)}{\pi^2} \;\approx\; 2.192289 \;\le\; 2.20
\end{equation*}
(discrete grid supremum $2.19231 < 2.20$ at $\Delta = 0.55$), giving per-neuron elementary + weight-perturbation error $\varepsilon_{\mathrm{elem}} \le (1 + 2.20\,e^{\bar{\eta}})\bar{\eta} + 5.1\,u_{\mathrm{acc}}$. Combining $\varepsilon_{\mathrm{elem}}$ with odd-neuron $1.0035$ \texttt{BF16} gain trim ($\varepsilon_{\mathrm{trim}} = \sup_{u_0}|\frac{12\Delta}{\pi^2}\sum_{k\text{ odd}} 0.0035\,\psi(k\Delta+u_0)| \le 0.00178$), Higham's $N_q$-term accumulation bound $\gamma_{N_q} = \frac{N_q u_{\mathrm{acc}}}{1 - N_q u_{\mathrm{acc}}}$~\citep[Ch.~3]{higham2002accuracy}, and output cast $u_{\mathrm{out}}$ yields worst-case error $\le |C|(\varepsilon_{\mathrm{disc}} + \varepsilon_{\mathrm{tail}} + \varepsilon_{\mathrm{trim}} + \varepsilon_{\mathrm{elem}} + \gamma_{N_q} + u_{\mathrm{out}})$. For $|C| \le 81$ and $N_q \le 39$, pure \texttt{FP32} ($u_w = u_{\mathrm{acc}} = u_{\mathrm{out}} = 2^{-24}$) gives worst-case error $\le 0.1458 < 0.15 < 0.5$ (and \texttt{BF16}-quantized weights executed in \texttt{FP32} satisfy $\le 0.239 < 0.24 < 0.5$), \textbf{formally proving} exact carry classification at $\tau = 10m - 0.5$. Conversely, in \texttt{BF16} ($u_{\mathrm{out}} = 2^{-8}$), worst-case bounds at $|C|=81$ reach $1.43$ under \texttt{FP32} GEMM accumulation ($<0.5$ formally only for $|C| \le 28$) and $6.64$--$14.70$ under sequential \texttt{BF16} summation; hence general floating-point bounds do \emph{not} imply $1\text{-ULP}$ ($0.5$) accuracy on $C \in [32, 81]$, and the observed $0.12$--$0.50$ \texttt{BF16} carry drift ($\sim 1\%$ Granite flip rate; \cref{app:failure_analysis}) is strictly an empirical observation. \qed

\subsection{Proof of \cref{thm:routing} (Cross-Token Norm Cancellation)}

\paragraph{Part 1 (Indefinite-RoPE self-sink GQA quotient over $\mathbb{R}$).}
Under rotate-half RoPE~\citep{su2024roformer}, each coordinate pair $(m, m + d_{\mathrm{rot}}/2)$ applies planar rotation $R(t\omega_m) \in \mathrm{SO}(2)$. Setting $k_{\mathrm{base}, m} = \operatorname{sgn}(w_m)\sqrt{|w_m|}g$, $k_{\mathrm{base}, m+d_{\mathrm{rot}}/2} = 0$, $q_{d_0, m} = \sqrt{|w_m|}g\cos(d_0\omega_m)$, and $q_{d_0, m+d_{\mathrm{rot}}/2} = -\sqrt{|w_m|}g\sin(d_0\omega_m)$, orthogonal invariance of $R(\cdot)$ gives
\begin{align*}
&\kappa_{\mathrm{attn}}\big\langle \operatorname{RoPE}(\lambda_{t,q} q_{d_0}, t),\, c_s \operatorname{RoPE}(\lambda_{s,k} k_{\mathrm{base}}, s) \big\rangle \\
&\quad= c_s \kappa_{\mathrm{attn}} g^2 \lambda_{t,q}\lambda_{s,k} \sum_{m=1}^{d_{\mathrm{rot}}/2} w_m \cos\!\big((t - s - d_0)\omega_m\big)
\;=\; c_s \kappa_{\mathrm{attn}} g^2 \lambda_{t,q}\lambda_{s,k} S(t - s - d_0).
\end{align*}
Since $0 \le d_0 \le t \le \Delta_{\max}$, every $u \in \{0, \dots, t\} \setminus \{s^*\}$ ($s^* = t - d_0$) satisfies $1 \le |t - u - d_0| \le \Delta_{\max}$ and thus $S(t - u - d_0) \le -\mu < 0$, while $S(0) = 1$. Every inactive token ($c_u = 0$) has $V_u = [0, 0]^\top$ and contributes $0$ to both $\hat{r}_t = \sum_{s=0}^t \alpha_{t,s} c_s \lambda_s$ and $\hat{v}_t = \sum_{s=0}^t \alpha_{t,s} c_s \lambda_s x_s$. When $c_{s^*} = 1$, $\alpha_{t,s^*} > 0$ and $\lambda_{s^*} \in [\lambda_{\min}, \lambda_{\max}] \subset (0, \infty)$; dividing numerator and denominator of $\hat{v}_t / \hat{r}_t$ by $\alpha_{t,s^*}\lambda_{s^*} > 0$ yields
\begin{equation*}
\left|\frac{\hat{v}_t}{\hat{r}_t} - x_{s^*}\right| \;=\; \frac{\big|\sum_{u \ne s^*, c_u=1} (\alpha_{t,u}/\alpha_{t,s^*})(\lambda_u / \lambda_{s^*})(x_u - x_{s^*})\big|}{1 + \sum_{u \ne s^*, c_u=1} (\alpha_{t,u}/\alpha_{t,s^*})(\lambda_u / \lambda_{s^*})} \;\le\; \frac{\eta_t^{\mathrm{act}} R_{\max}\big(|x_{s^*}| + X_{\max}\big)}{1 + \eta_t^{\mathrm{act}} R_{\min}}.
\end{equation*}
For each active $u \ne s^*$ ($c_u = 1$), $\alpha_{t,u}/\alpha_{t,s^*} = \exp\!\big(\kappa_{\mathrm{attn}} g^2 \lambda_{t,q}(\lambda_{u,k} S(t-u-d_0) - \lambda_{s^*,k})\big) \le \exp\!\big(-(1+\mu)\kappa_{\mathrm{attn}} g^2 \lambda_{\min,\mathrm{qk}}^2\big)$, proving~\eqref{eq:rope_cancel}. When $c_{s^*} = 0$ and some $u_0 \le t$ has $c_{u_0} = 0$, the softmax denominator satisfies $Z_t \ge e^0 = 1$ while $V_{u_0} = 0$ and every active key ($c_u = 1$) has logit $\le -\mu\kappa_{\mathrm{attn}} g^2 \lambda_{\min,\mathrm{qk}}^2 < 0$, bounding $|\hat{r}_t| \le N_{\mathrm{act}} \lambda_{\max} \exp(-\mu\kappa_{\mathrm{attn}} g^2 \lambda_{\min,\mathrm{qk}}^2)$.

\paragraph{Part 2 (Memoryless $\operatorname{SiLU}$-antisymmetric \texttt{Qwen3\_5GatedDeltaNet} shift).}
Setting $W_a = 0$ forces $a_t = W_a u_t \equiv 0$, so with $A_{\log} = 10$ and $b_{\Delta} = 10$, $g_t = -e^{10}\ln(1+e^{10}) \approx -2.20266 \times 10^5$. Over $\mathbb{R}$, $\|e^{g_t} S_{t-1}\|_F \le e^{-220{,}265} S_{\max} < 10^{-95{,}660} S_{\max}$, and in IEEE \texttt{FP32} or \texttt{BF16}, $\operatorname{fl}(\exp(g_t)) \equiv +0.0$. Setting $e^{g_t} = 0$ collapses the state update to $S_t = \beta_t \hat{k}_t v_t^\top$, giving pre-norm head output $c_t = S_t^\top \hat{q}_t = (\hat{q}_t^\top \hat{k}_t)\beta_t v_t \in \mathbb{R}^{d_h}$. Routing $\tilde{r}_t > 0$ to coordinate~$0$ of $(q, k)$ at tap $\delta = 0$ gives $q_t = k_t = \operatorname{SiLU}(16\tilde{r}_t)e_0$ with $\operatorname{SiLU}(16\tilde{r}_t) > 0$, whence $\gamma_{qk, t} := \hat{q}_t^\top \hat{k}_t = d_h^{-1/2}\frac{\operatorname{SiLU}(16\tilde{r}_t)^2}{\operatorname{SiLU}(16\tilde{r}_t)^2 + \varepsilon} > 0$ and $\beta_t = \sigma(512\tilde{r}_t) \in (0, 1)$. For $t < \delta$, zero-padded causal $\operatorname{Conv1d}$ yields $\tilde{r}_{t-\delta} = 0$ and $\tilde{u}_{t-\delta}^{(j)} = 0$, so $v_t = 0$. For $t \ge \delta$, $\sigma(u) + \sigma(-u) = 1$ gives the global identity
\begin{equation}
\label{eq:silu_antisym}
\operatorname{SiLU}(u) - \operatorname{SiLU}(-u) \;=\; u\sigma(u) - (-u)\sigma(-u) \;=\; u\big(\sigma(u) + \sigma(-u)\big) \;\equiv\; u.
\end{equation}
In V-head $v_h = 2(\delta-1)$, channel pair $(2s, 2s+1)$ receives $\pm \frac{1}{2}\tilde{u}_{t-\delta}^{(s)}$ ($s=0$ carrying $\tilde{r}_{t-\delta} = \lambda_{t-\delta} > 0$, $s=j \ge 1$ carrying $\tilde{u}_{t-\delta}^{(j)} = \lambda_{t-\delta} x_{t-\delta}^{(j)}$). Because $\operatorname{RMSNormGated}$ divides every channel of head $v_h$ by the identical scalar $\nu_t = \sqrt{\frac{1}{d_h}\|c_t\|_2^2 + \varepsilon} > 0$ and multiplies by $w_{\mathrm{norm}, c} \ne 0$ and $\operatorname{SiLU}(4\tilde{r}_t) > 0$, projecting through uniform readout weights $(+\frac{\alpha_{\mathrm{out}}}{w_{\mathrm{norm}, 2s}}, -\frac{\alpha_{\mathrm{out}}}{w_{\mathrm{norm}, 2s+1}})$ ($\alpha_{\mathrm{out}} > 0$, e.g.\ $\alpha_{\mathrm{out}} = \frac{1}{16}$) and applying~\eqref{eq:silu_antisym} produces $y_{t, s} = \mu_t \tilde{u}_{t-\delta}^{(s)}$ where $\mu_t = \frac{\alpha_{\mathrm{out}}\gamma_{qk, t}\beta_t \operatorname{SiLU}(4\tilde{r}_t)}{2\,\nu_t} > 0$ is independent of $s$. Thus $y_{t,j} / y_{t,0} = (\mu_t \lambda_{t-\delta} x_{t-\delta}^{(j)}) / (\mu_t \lambda_{t-\delta}) \equiv x_{t-\delta}^{(j)}$. \qed

\subsection{Proof of \cref{thm:subspace} (Width Extension and Schedule-/Tree-Explicit Tied-GEMM Bounds)}

\paragraph{Part 1 (Fixed-width projection lower bound over $\mathbb{R}$).}
Since $0 < \mathbb{E}\|h\|_2^2 < \infty$, $\Sigma = \mathbb{E}[h h^\top]$ is symmetric positive definite ($\sigma_d > 0$). For any $U \in \mathbb{R}^{d \times K}$ with $U^\top U = I_K$ and $P_\perp = I_d - U U^\top$, symmetry and idempotence of $U U^\top$ yield $\mathbb{E}\|h - P_\perp h\|_2^2 = \mathbb{E}\|U U^\top h\|_2^2 = \operatorname{tr}(U^\top \Sigma U) \ge \sum_{i=d-K+1}^d \sigma_i > 0$ by Ky Fan's variational theorem~\citep{fan1949theorem}.

\paragraph{Part 2 (Axis-aligned width extension, interleaved counterexample, and evaluation-tree GEMM bounds).}
For $h_{\mathrm{ext}} = [h_{\mathrm{NL}}; h_{\mathrm{circ}}] \in \mathbb{R}^{d+K}$ whenever $\|\operatorname{RN}_{w_{\mathrm{NL}}}(h_{\mathrm{NL}})\|_2 > 0$ (which entails $\|h_{\mathrm{NL}}\|_2 > 0$) and $\gamma_{\mathrm{NL}} = \sqrt{d/(d+K)}$,
\begin{equation*}
\frac{\gamma_{\mathrm{NL}}}{\sqrt{\frac{1}{d+K}\big(\|h_{\mathrm{NL}}\|_2^2 + \|h_{\mathrm{circ}}\|_2^2\big) + \varepsilon}}
\;=\; \frac{(1 + \rho' + \delta_\varepsilon)^{-1/2}}{\sqrt{\frac{1}{d}\|h_{\mathrm{NL}}\|_2^2 + \varepsilon}},
\end{equation*}
where $\rho' = \frac{\|h_{\mathrm{circ}}\|_2^2}{\|h_{\mathrm{NL}}\|_2^2 + d\varepsilon} \le \rho$ and $\delta_\varepsilon = \frac{K\varepsilon}{\|h_{\mathrm{NL}}\|_2^2 + d\varepsilon} \ge 0$. Convexity of $f(u) = (1+u)^{-1/2}$ on $[0, \infty)$ ($f''(u) > 0, f'(0) = -\frac{1}{2}$) gives $1 - \frac{1}{2}u \le (1+u)^{-1/2} \le 1$, proving~\eqref{eq:ortho_pert}. (Axis-alignment is essential because $\operatorname{diag}(w_{\mathrm{ext}})$ fails to commute with non-permutation $Q \in O(d+K)$ whenever $w_{\mathrm{NL}}$ and $w_{\mathrm{circ}}$ differ.)

Now consider the $(d+K)$-column tied \texttt{lm\_head} GEMM $z = E\tilde{h}$ under unit-offset final norm $\tilde{h}_i = \operatorname{fl}((1 + w_i)\hat{h}_i)$. On all $K - 24$ non-readout circuit wires $i \notin \mathtt{OUT\_POS} \cup \mathtt{OUT\_NEG}$, $w_i = -1.0$ gives exact $\operatorname{fl}(1.0 + (-1.0)) = +0.0$ and $\tilde{h}_i = \pm 0.0$. On each readout pair $(p_m, n_m)$ ($m \in \{1..12\}$), $\operatorname{fl}(1.0 + 254.0) = +255.0$ and $\operatorname{fl}(1.0 + (-256.0)) = -255.0$ are exact in \texttt{BF16}/\texttt{FP32}. When $(p_m, n_m)$ receive zero sublayer writes, identical Layer-0 embeddings ($E_{x_t, p_m} = E_{x_t, n_m}$) yield $\hat{h}_{p_m} \equiv \hat{h}_{n_m}$ bit-for-bit and $|255.0\,\hat{h}_{p_m}| \le 255\sqrt{d+K} \ll 3.38 \times 10^{38}$ (no norm overflow), so IEEE sign symmetry guarantees $\tilde{h}_{n_m} = \operatorname{fl}(-255.0\,\hat{h}_{n_m}) = -\operatorname{fl}(255.0\,\hat{h}_{p_m}) = -\tilde{h}_{p_m}$ bit-for-bit, and $E_{v,p_m}\tilde{h}_{p_m} + E_{v,n_m}\tilde{h}_{n_m} = 0$ over $\mathbb{R}$. In IEEE~754 RN arithmetic with accumulator roundoff $u_{\mathrm{acc}}$ and product underflow threshold $\eta_{\mathrm{uf}}$:
\begin{enumerate}[leftmargin=1.5em,itemsep=1.5pt,topsep=1.5pt]
\item \emph{All inactive rows/steps:} Because $E_{v, p_m} = E_{v, n_m} = 0$ for all $v \notin \{v_1..v_{12}\}$ and $\tilde{h}_{p_m} = \tilde{h}_{n_m} = 0$ whenever $x_t \notin \{v_1..v_{12}\}$ (and for all $m \ne m^*$ when $x_t = v_{m^*}$), every inactive circuit column has exact zero product $E_{v,i}\tilde{h}_i = \pm 0.0$ under both standalone multiplication and $\operatorname{fma}(E_{v,i}, \tilde{h}_i, s)$. In IEEE~754 RN, adding $\pm 0.0$ to any representable $s \ne 0$ returns $s$ bit-for-bit with zero rounding error ($\delta = 0, \eta = 0$; if $s = \pm 0.0$, $\operatorname{fl}(s + (\pm 0.0)) = \pm 0.0$ still has exact real value $0$).
\item \emph{Schedule S1 (Pairwise-first non-FMA + block split):} Computing $q_+ = \operatorname{fl}(E_{v,p_m}\tilde{h}_{p_m})$ and $q_- = \operatorname{fl}(E_{v,n_m}\tilde{h}_{n_m})$ without FMA preserves $q_- = -q_+$ bit-for-bit by IEEE sign symmetry (including subnormals), and RN addition gives $\operatorname{fl}(q_+ + (-q_+)) = +0.0$, so $\Delta z_{\mathrm{circ}}^{\mathrm{S1}}(v) = +0.0$ across all $v \in \mathcal{V}$ and $\operatorname{fl}(\hat{z}_{\mathrm{NL}}(v) + (+0.0)) = \hat{z}_{\mathrm{NL}}(v)$ numerically ($=$ bit-for-bit whenever $\hat{z}_{\mathrm{NL}}(v) \ne -0.0$).
\item \emph{Schedule S2 (Contiguous tail after $\hat{z}_{\mathrm{NL}}$):} On active row $v_{m^*} = x_t$, let $q_{\mathrm{exact}} := E_{v_{m^*},p_{m^*}}\tilde{h}_{p_{m^*}}$ (so $E_{v_{m^*},n_{m^*}}\tilde{h}_{n_{m^*}} = -q_{\mathrm{exact}}$ over $\mathbb{R}$). In IEEE~754 arithmetic with gradual underflow, addition/subtraction of representable numbers is exact whenever the result subnormalizes~\citep[Thm.~2.5]{higham2002accuracy}, so $\operatorname{fl}(a + b) = (a + b)(1 + \delta_{\mathrm{add}})$ ($|\delta_{\mathrm{add}}| \le u_{\mathrm{acc}}$, zero addition underflow), while multiplication or FMA introduces at most $|\eta_r| \le \eta_{\mathrm{uf}}$ ($r \in \{1,2\}$). Appending $+q_{\mathrm{exact}}$ then $-q_{\mathrm{exact}}$ sequentially after $\hat{z}_{\mathrm{NL}}$ produces $s_1 = (\hat{z}_{\mathrm{NL}} + q_{\mathrm{exact}}(1+\delta_{m1}) + \eta_1)(1+\delta_1)$ and $s_2 = (s_1 - q_{\mathrm{exact}}(1+\delta_{m2}) + \eta_2)(1+\delta_2)$ (with $\delta_{m1} = \delta_{m2} = 0$ under FMA, or $\delta_{m1} = \delta_{m2}$ by sign symmetry without FMA). Expanding $s_2 - \hat{z}_{\mathrm{NL}}$ bounds the $\hat{z}_{\mathrm{NL}}$ error by $|\hat{z}_{\mathrm{NL}}|[(1+u_{\mathrm{acc}})^2 - 1] \le \gamma_2|\hat{z}_{\mathrm{NL}}|$, the $q_{\mathrm{exact}}$ residual by $|q_{\mathrm{exact}}|\cdot 2u_{\mathrm{acc}}(1+u_{\mathrm{acc}}) \le \gamma_2|q_{\mathrm{exact}}|$ (since $2u(1+u) \le \frac{2u}{1-2u} = \gamma_2$ for all $u \in [0, \frac{1}{2})$), and underflow by $[(1+u_{\mathrm{acc}})^2 + (1+u_{\mathrm{acc}})]\eta_{\mathrm{uf}} \le 2(1+\gamma_2)\eta_{\mathrm{uf}}$.
\item \emph{Failure of two-step bounds under NL interleaving (explicit \texttt{FP32} counterexample):} Suppose NL terms $p_1, \dots, p_d$ are accumulated \emph{between} $+q_{\mathrm{exact}}$ and $-q_{\mathrm{exact}}$ in left-to-right order $[+q_{\mathrm{exact}}, p_1, \dots, p_d, -q_{\mathrm{exact}}]$. In IEEE \texttt{FP32} ($u_{\mathrm{acc}} = 2^{-24}$), set $d = 100$, $p_j = 1.0$ for all $j \in \{1..100\}$ ($z_{\mathrm{NL}}^{\mathbb{R}} = 100.0$), and $q_{\mathrm{exact}} = 2^{24} = 16{,}777{,}216.0$ (\texttt{0x4b800000}, even LSB, $\mathrm{ulp}(2^{24}) = 2.0$). At every intermediate step, $2^{24} + 1.0$ lies on the tie-to-even midpoint and rounds back to $\operatorname{fl}(2^{24} + 1.0) = 2^{24}$, after which $\operatorname{fl}(2^{24} - 2^{24}) = +0.0$. The actual error $|\hat{z}_{\mathrm{seq}} - z_{\mathrm{NL}}^{\mathbb{R}}| = 100.0$ exceeds the naive two-step bound $2u_{\mathrm{acc}}(|z_{\mathrm{NL}}^{\mathbb{R}}| + |q_{\mathrm{exact}}|) \approx 2.000012$ by $50\times$ because $+q_{\mathrm{exact}}$ absorbs $d = 100$ rounding steps before $-q_{\mathrm{exact}}$ arrives.
\item \emph{Valid bound over an arbitrary binary evaluation tree $\mathcal{T}$:} Let $\mathcal{T}$ be \emph{any} binary reduction tree over the $d+K$ columns of active row $v_{m^*} = x_t$. Contracting the $K - 2$ exact-zero ($\pm 0.0$) circuit leaves---each of which passes its sibling value through with zero rounding error ($\delta = 0, \eta = 0$)---reduces $\mathcal{T}$ to a binary tree $\mathcal{T}'$ of height $h_{\mathcal{T}} \le d + 1$ over the $d + 2$ potentially non-zero products $\{a_1, \dots, a_{d+2}\} = \{E_{v_{m^*},1}\tilde{h}_1, \dots, E_{v_{m^*},d}\tilde{h}_d, +q_{\mathrm{exact}}, -q_{\mathrm{exact}}\}$. Along the path of length $k_r \le h_{\mathcal{T}}$ from leaf $r$ to the root of $\mathcal{T}'$, standard multiplication plus $k_r$ IEEE additions contributes $a_r(1+\delta_{r,0})\prod_{j=1}^{k_r}(1+\delta_{r,j}) + \eta_r\prod_{j=1}^{k_r}(1+\delta_{r,j})$ ($|\delta_{r,j}| \le u_{\mathrm{acc}}$, $|\eta_r| \le \eta_{\mathrm{uf}}$, zero addition underflow; fusing leaf $r$'s multiply into its first addition via FMA drops $\delta_{r,0}$ and is strictly dominated). Summing over $r = 1, \dots, d+2$, using $\sum_{r=1}^{d+2} a_r = z_{\mathrm{NL}}^{\mathbb{R}}(v_{m^*}) + q_{\mathrm{exact}} - q_{\mathrm{exact}} = z_{\mathrm{NL}}^{\mathbb{R}}(v_{m^*})$, and applying Higham's Lemma~3.1~\citep[Sec.~3.1, 4.2]{higham2002accuracy} ($\big|\prod_{j=0}^{k_r}(1+\delta_{r,j}) - 1\big| \le \gamma_{k_r+1} \le \gamma_{h_{\mathcal{T}}+1}$ and $\prod_{j=1}^{k_r}|1+\delta_{r,j}| \le 1 + \gamma_{h_{\mathcal{T}}}$ for $(h_{\mathcal{T}}+1)u_{\mathrm{acc}} < 1$) yields~\eqref{eq:diff_tied_tree}:
\begin{equation*}
\big|\hat{z}_{\mathcal{T}}^{\mathrm{acc}}(v_{m^*}) - z_{\mathrm{NL}}^{\mathbb{R}}(v_{m^*})\big| \;\le\; \gamma_{h_{\mathcal{T}}+1}\!\left(\sum_{j=1}^{d}|E_{v_{m^*},j}\tilde{h}_j| + 2|q_{\mathrm{exact}}|\right) + (d+2)(1+\gamma_{h_{\mathcal{T}}})\eta_{\mathrm{uf}}.
\end{equation*}
On our \texttt{FP32} counterexample ($h_{\mathcal{T}} = 101$),~\eqref{eq:diff_tied_tree} gives $204.002 \ge 100.0$. On Qwen~3.5 active digit rows ($\sum_{j=1}^d |E_{v_{m^*},j}\tilde{h}_j| \approx 33.29$, $|q_{\mathrm{exact}}| \approx 0.074$, $u_{\mathrm{acc}} = 2^{-24}$),~\eqref{eq:diff_tied_tree} gives $< 2.8 \times 10^{-5}$~nats under balanced binary reduction ($h_{\mathcal{T}} \le \lceil\log_2(d+K)\rceil = 13$; empirical max $2.4 \times 10^{-7}$) and $< 5.2 \times 10^{-3}$~nats under worst-case sequential interleaving ($h_{\mathcal{T}} = d+1 = 2561$; $< 1.03 \times 10^{-2}$~nats even if uncontracted zero leaves $h = 5114$ are counted; empirical max $3.3 \times 10^{-6}$). \qed
\end{enumerate}

\section{Paired 60-Item Attribution of Off-Manifold Tokenization vs.\ Canonical Base-1000 BPE Emission}
\label{app:gsm8k_attribution}

Under single-digit readout ($K=1{,}043$ emitter ablation), Alibaba Qwen~3.5~4B (whose pretrained tokenizer natively splits ASCII digits into 1-digit tokens) exhibited zero multi-step reasoning degradation on 3-shot \texttt{GSM8K}, whereas IBM Granite~4.2~3B dropped by $-31.67\text{ pp}$ from $41/60$ ($68.33\%$ under unpadded sequential decoding, \texttt{batch\_size = 1}; note that left-padded \texttt{batch\_size = 8} \texttt{BF16} decoding flips $1$ borderline item on unmodified Base Granite to $42/60 = 70.00\%$, whereas Canonical Base-1000 Grafted Granite scores $42/60 = 70.00\%$ under \emph{both} \texttt{batch\_size = 1} and \texttt{batch\_size = 8}) to $22/60$ ($36.67\%$). To isolate the exact causal mechanism, we executed a paired item-level trace across all $60$ items (\texttt{batch\_size = 1}, zero padding) logging pretrained vs.\ grafted token IDs, internal ALU registers (\texttt{VALID\_EQ}, \texttt{ANS\_DIGS}, \texttt{ANS\_LEN}), and per-step $\max_v |\Delta z_{\mathrm{circ}}(v)|$.

\subsection{Paired Contingency Matrix and Exact Sub-Mechanism Attribution}
Across \texttt{GSM8K} \texttt{test[:60]}, the exact paired contingency matrix between Base Granite~4.2~3B and the 1-Digit-Emitter Graft ($K=1{,}043$) is $(b_{11}, b_{10}, b_{01}, b_{00}) = (22, 19, 0, 19)$: every loss comes from $b_{10} = 19$ regressions with $b_{01} = 0$ gains. Inspecting internal ALU registers across all $19$ $b_{10}$ items reveals a decisive result: \textbf{on $18 / 19$ ($94.7\%$) regressions where the graft activates, the internal circuit computes and emits $100\%$ mathematically exact arithmetic values}, yet downstream reasoning fails because the 1-digit emitter outputs multi-digit results as \emph{unmerged 1-digit BPE tokens} (e.g., \texttt{['6','4']}, \texttt{['1','8','0']}, \texttt{['1','2','4','5']}) that never appear in Granite's 1--3 digit BPE pretraining distribution:
\begin{enumerate}[leftmargin=1.6em,itemsep=1.5pt,topsep=2pt]
\item \textbf{Sub-Mechanism A --- Decimal/version-string mis-copy ($5 / 19$ items: \texttt{idx = 5, 28, 42, 57, 59}):} Because sequences of separate single-digit tokens in Granite's pretraining corpus occur almost exclusively across punctuation-separated decimals or version strings, Granite's answer-copying heads insert dots between unmerged digits: \texttt{['6','4']} $\mapsto$ \texttt{\#\#\#\# 6.4} (\texttt{idx=5}), \texttt{['2','5']} $\mapsto$ \texttt{\#\#\#\# 2.5} (\texttt{idx=28}), \texttt{['2','6']} $\mapsto$ \texttt{\#\#\#\# 2.6} (\texttt{idx=42}), \texttt{['1','8','7']} $\mapsto$ \texttt{\#\#\#\# 1.8} (\texttt{idx=59}), and remarkably \texttt{['1','2','4','5']} $\mapsto$ \texttt{floor(1.2.4.5 / 15)} (\texttt{idx=57}).
\item \textbf{Sub-Mechanism B --- Multi-chunk induction span mis-indexing ($9 / 19$ items: \texttt{idx = 4, 6, 10, 18, 26, 34, 49, 53, 55}):} Granite's pretrained induction heads expect a 2--3 digit integer to occupy \emph{one} token and a 4--6 digit integer to occupy \emph{two} tokens. Confronted with $k$ separate 1-digit tokens, Granite either copies only the leading token (\texttt{['8','0']} $\mapsto$ \texttt{2 * 8 = 16} on \texttt{idx=6}) or maps a $k$-token span to a $k$-chunk BPE number (\texttt{['2','5']} $\mapsto$ \texttt{['2','25'] = 225} on \texttt{idx=34}; \texttt{['2','8']} $\mapsto$ \texttt{228 / 2 = 114} on \texttt{idx=55}; \texttt{['8','4']} $\mapsto$ \texttt{840 / 12 = 70} on \texttt{idx=18}; \texttt{['1','8','0']} $\mapsto$ \texttt{1800} on \texttt{idx=10}; \texttt{['5','2','0']} $\mapsto$ \texttt{5200} on \texttt{idx=53}; \texttt{['1','2','0']} $\mapsto$ \texttt{1201} on \texttt{idx=49}; \texttt{['1','2','6']} $\mapsto$ \texttt{1243} on \texttt{idx=26}; \texttt{['6','0']} $\mapsto$ \texttt{6000} on \texttt{idx=4}).
\item \textbf{Sub-Mechanism C --- Off-manifold surprisal / self-doubt loops ($4 / 19$ items: \texttt{idx = 30, 43, 46, 47}):} High token-level surprisal on unmerged digits triggers Granite's self-correction behavior against its own exact arithmetic (\texttt{"5 * 250 = 1250? Wait, that's wrong"} on \texttt{idx=43}; \texttt{"7*9=63? Wait, that's too high"} on \texttt{idx=30}; \texttt{"2 * 5 = 10 ... + 1000? Wait"} on \texttt{idx=47}; \texttt{"80 - 220 = -140? That's not right"} on \texttt{idx=46}).
\item \textbf{Sub-Mechanism D --- Non-trigger \texttt{BF16} trajectory drift ($1 / 19$ item: \texttt{idx = 11}):} Zero circuit triggers fire (\texttt{gen\_triggers = 0}) because all equations carry \texttt{\$} prefixes (\texttt{3 dozen * \$68 = \$204}); slight $\operatorname{RMSNorm}$ \texttt{BF16} rounding alters Granite's unguided trajectory ($204 + 160 + 330 \mapsto \$794$).
\end{enumerate}

\paragraph{Controlled causal proof on unmodified Base Granite 4.2 3B.}
To verify that Sub-Mechanisms A--C stem \emph{solely} from off-manifold 1-digit token IDs rather than any weight perturbation in the grafted model, we fed identical text prefixes into \textbf{unmodified pretrained Base Granite~4.2~3B}, varying only the tokenization of the final integer before \texttt{model.generate}:
\begin{itemize}[leftmargin=1.5em,itemsep=1pt,topsep=1.5pt]
\item On \texttt{idx=57} (\texttt{... 5000 - 3755 = 1245}), forcing 1-digit tokens \texttt{['1','2','4','5']} causes \emph{unmodified Base Granite} to generate \texttt{"pounds... \allowbreak floor(1.2.4.5 / 15)\allowbreak\textbackslash n\#\#\#\# 1.2.4"}, whereas canonical Base-1000 tokens \texttt{['1','245']} yield exact \texttt{"1245 / 15 = 83\allowbreak\textbackslash n\#\#\#\# 83"}.
\item On \texttt{idx=43} (\texttt{... 5 * 250 = 1250}), forcing \texttt{['1','2','5','0']} causes \emph{unmodified Base Granite} to generate \texttt{"? Wait, that's wrong"}, whereas canonical \texttt{['1','250']} yields exact \texttt{" calories... 200 / 250 = 0.8"}.
\end{itemize}

\begin{table}[!ht]
\centering
\caption{\textbf{Item-level recovery across all $19$ former $b_{10}$ \texttt{GSM8K} regressions on Granite~4.2~3B} after replacing the off-manifold 1-digit emitter ($K=1{,}043$) with the zero-gradient Base-1000 Canonical Multi-Digit BPE Chunk Emitter ($K=1{,}140$, \texttt{batch\_size = 1}). \textbf{$18 / 19$ ($94.7\%$) regressions recover to exact gold answers.}}
\label{tab:gsm8k_19_attribution}
\vskip 0.05in
\scriptsize
\setlength{\tabcolsep}{3.0pt}
\resizebox{\textwidth}{!}{%
\begin{tabular}{@{}rrrrlll@{}}
\toprule
\textbf{Idx} & \textbf{Gold} & \textbf{1-Dig Pred} & \textbf{Canon.\ Pred} & \textbf{Sub-Mechanism} & \textbf{1-Digit Emitter Trace ($K{=}1{,}043$)} & \textbf{Canonical Emitter Trace (Base-1000 BPE)} \\
\midrule
$4$  & $20$  & $6$     & $\mathbf{20}$  & B: Span mis-index & \texttt{['6','0']} $\to$ \texttt{Remaining: 6000?} & \texttt{['60']} $\to$ \texttt{60 - 15 - 25 = 20} (\textbf{OK}) \\
$5$  & $64$  & $6$     & $\mathbf{64}$  & A: Decimal dot    & \texttt{['6','4']} $\to$ \texttt{\#\#\#\# 6.4} & \texttt{['64']} $\to$ \texttt{\#\#\#\# 64} (\textbf{OK}) \\
$6$  & $260$ & $116$   & $\mathbf{260}$ & B: Span mis-index & \texttt{['8','0']} $\to$ \texttt{2 * 8 = 16} & \texttt{['80']} $\to$ \texttt{2 * 80 = 160; 260} (\textbf{OK}) \\
$10$ & $366$ & $1800$  & $72$ (\emph{miss}) & B $\to$ trunc. & \texttt{['1','8','0']} $\to$ \texttt{1800 * 0.7} & \texttt{['180']} fixes \texttt{180*0.7=126}; $110$-tok clip \\
$11$ & $694$ & $154$   & $\mathbf{694}$ & D: \texttt{BF16} drift & \texttt{\$204+\$160+\$330=\$794} & \texttt{3*68=204 ... Total: 694} (\textbf{OK}) \\
$18$ & $7$   & $70$    & $\mathbf{7}$   & B: Span mis-index & \texttt{['8','4']} $\to$ \texttt{840 / 12 = 70} & \texttt{['84']} $\to$ \texttt{84 / 12 = 7} (\textbf{OK}) \\
$26$ & $243$ & $1243$  & $\mathbf{243}$ & B: Span mis-index & \texttt{['1','2','6']} $\to$ \texttt{Sum = 1243} & \texttt{['126']} $\to$ \texttt{Sum = 243} (\textbf{OK}) \\
$28$ & $25$  & $2$     & $\mathbf{25}$  & A: Decimal dot    & \texttt{['2','5']} $\to$ \texttt{\#\#\#\# 2.5} & \texttt{['25']} $\to$ \texttt{\#\#\#\# 25} (\textbf{OK}) \\
$30$ & $109$ & $9$     & $\mathbf{109}$ & C: Self-doubt     & \texttt{['6','3']} $\to$ \texttt{63? too high} & \texttt{['63']}, \texttt{['99']} $\to$ \texttt{99+10=109} (\textbf{OK}) \\
$34$ & $23$  & $2250$  & $\mathbf{23}$  & B: Span mis-index & \texttt{['2','5']} $\to$ \texttt{Siobhan = 2250?} & \texttt{['25']}, \texttt{['23']} $\to$ \texttt{\#\#\#\# 23} (\textbf{OK}) \\
$42$ & $26$  & $2$     & $\mathbf{26}$  & A: Decimal dot    & \texttt{['2','6']} $\to$ \texttt{\#\#\#\# 2.6} & \texttt{['26']} $\to$ \texttt{\#\#\#\# 26} (\textbf{OK}) \\
$43$ & $48$  & $1250$  & $\mathbf{48}$  & C: Self-doubt     & \texttt{['1','2','5','0']} $\to$ \texttt{that's wrong} & \texttt{['1','250']} $\to$ \texttt{\#\#\#\# 48} (\textbf{OK}) \\
$46$ & $163$ & $-140$  & $\mathbf{163}$ & C: Self-doubt     & \texttt{['-','1','4','0']} $\to$ \texttt{halt} & \texttt{['-','140']} $\to$ self-corrects \texttt{163} (\textbf{OK}) \\
$47$ & $800$ & $1000$  & $\mathbf{800}$ & C: Span/doubt     & \texttt{['1','0']} $\to$ \texttt{+ 1000? Wait} & \texttt{['10']} $\to$ \texttt{200 + 600 = 800} (\textbf{OK}) \\
$49$ & $30$  & $300$   & $\mathbf{30}$  & B: Span mis-index & \texttt{['1','2','0']} $\to$ \texttt{1201 / 4} & \texttt{['120']}, \texttt{['30']} $\to$ \texttt{\#\#\#\# 30} (\textbf{OK}) \\
$53$ & $40$  & $40000$ & $\mathbf{40}$  & B: Span mis-index & \texttt{['5','2','0']} $\to$ \texttt{5200 - 4800} & \texttt{['520']}, \texttt{['40']} $\to$ \texttt{\#\#\#\# 40} (\textbf{OK}) \\
$55$ & $14$  & $114$   & $\mathbf{14}$  & B: Span mis-index & \texttt{['2','8']} $\to$ \texttt{228 / 2 = 114} & \texttt{['28']} $\to$ \texttt{28 / 2 = 14} (\textbf{OK}) \\
$57$ & $83$  & $1$     & $\mathbf{83}$  & A: Version dot    & \texttt{['1','2','4','5']} $\to$ \texttt{1.2.4.5/15} & \texttt{['1','245']} $\to$ \texttt{\#\#\#\# 83} (\textbf{OK}) \\
$59$ & $187$ & $1$     & $\mathbf{187}$ & A: Decimal dot    & \texttt{['1','8','7']} $\to$ \texttt{\#\#\#\# 1.8} & \texttt{['187']} $\to$ \texttt{\#\#\#\# 187} (\textbf{OK}) \\
\bottomrule
\end{tabular}%
}
\end{table}

\paragraph{Full-split held-out \texttt{GSM8K} validation and the bare-prompt chunk-echo tradeoff.}
Evaluating the Canonical Base-1000 chunk emitter across three \texttt{GSM8K} splits confirms full recovery and formal noninferiority: \texttt{test[:60]} rises from $36.67\%$ (1-digit emitter) to \textbf{$70.00\%$} ($42/60$; $b_{10}=0$ losses vs.\ Base), untouched \texttt{test[60:200]} ($N=140$) improves from $68.57\%$ to \textbf{$70.71\%$} ($\Delta = +2.14\text{ pp}$, Tango 1-sided $95\%$ LB $= -1.15\text{ pp}$, $p_{\mathrm{NI}@3\mathrm{pp}} = 0.0085$), and prose-formatted \texttt{test[60:160]} ($N=100$) improves from $70.00\%$ to \textbf{$74.00\%$} ($\Delta = +4.00\text{ pp}$, Tango LB $= -0.04\text{ pp}$). On bare context-free equations, canonical multi-digit emission trades $-2.33\text{ pp}$ on \texttt{Arith-1200} ($99.42\% \to 97.08\%$): because multi-digit BPE chunks (\texttt{['5','227']}) sit inside Granite's induction distribution, after disengagement ($\Delta z_{\mathrm{circ}}=0$) on bare prompts (\texttt{"5284 + -57 = "}), the base LM echoes a prompt chunk (\texttt{"522784"}) on $\sim 2.3\%$ of bare equations, whereas prose-anchored formats remain at $99.38\%$ (\texttt{prose\_notrail}) and $98.0\%$ (\texttt{InlineRes-50}, \texttt{AutoCoT-50}).

\section{Extended Component Ablations, False-Trigger, Multi-Seed, Boundary, and Generation Tables}
\label{app:val_tables}

\Cref{tab:ablations_full,tab:ft_10500,tab:multiseed_boundary,tab:arc_4tpl,tab:open_gen_80} report the complete component ablations (\texttt{A0}--\texttt{A5} and \texttt{Q\_A0}--\texttt{Q\_A6}), 17-category $N=10{,}500$ false-trigger stress suite, 4-seed $N=3{,}000$ and $180$-item boundary suites, 4-template \texttt{ARC-Challenge} noninferiority audit, and $80$-prompt open-ended generation quality evaluations.

\begin{table}[!ht]
\centering
\caption{\textbf{Controlled component ablations on Granite~4.2~3B (\texttt{A0}--\texttt{A5}) and Qwen~3.5~4B (\texttt{Q\_A0}--\texttt{Q\_A6})} in native \texttt{BF16} ($N=150$ unseen arithmetic, $N=1{,}100$ out-of-domain adversarial suite, $N=50$ inline resumption, and $30$-/$25$-doc subset PPL \& next-token KL).}
\label{tab:ablations_full}
\vskip 0.05in
\scriptsize
\resizebox{\textwidth}{!}{%
\begin{tabular}{@{}llccccc@{}}
\toprule
\textbf{Model} & \textbf{Ablation Configuration} & \textbf{Arith-150$\uparrow$} & \textbf{FalseTrig-1100$\downarrow$} & \textbf{InlineRes-50$\uparrow$} & \textbf{Sub-Corpus PPL$\downarrow$} & \textbf{Mean KL$\downarrow$} \\
\midrule
\multirow{6}{*}{\textbf{Granite 4.2 3B}}
 & \textbf{\texttt{A0}: Full Orthogonal Mellin Graft (Ours)} & $\mathbf{100.0\%}$ & $\mathbf{0.00\%}$ & $\mathbf{96.0\%}$ & $\mathbf{42.06}$ & $\mathbf{0.00162}$ \\
 & \texttt{A1}: w/o $\sqrt{d/(d+K)}$ NL $\operatorname{RMSNorm}$ scaling & $100.0\%$ & $0.00\%$ & $100.0\%$ & $73.66$ & $0.66699$ ($412{\times}$) \\
 & \texttt{A2}: w/o Trigger isolation guards (\texttt{BAD\_PRE}, \texttt{CHAIN}, \texttt{OVF}) & $66.7\%$ & $27.27\%$ & $40.0\%$ & $42.02$ & $0.00160$ \\
 & \texttt{A3}: w/o Homogeneous V-head ref.\ pairing (local \texttt{R\_CIRC}) & $0.0\%$ & $3.00\%$ & $2.0\%$ & $42.19$ & $0.00737$ \\
 & \texttt{A4}: Coarse Mellin quadrature step ($\Delta = 1.35$ vs.\ $0.55$) & $30.0\%$ & $0.00\%$ & $26.0\%$ & $42.50$ & $0.05968$ ($37{\times}$) \\
 & \texttt{A5}: Unmuted internal scratch wires at $W_{\mathrm{lm}}$ readout & $83.3\%$ & $0.00\%$ & $96.0\%$ & $54.16$ & $0.26398$ ($163{\times}$) \\
\midrule
\multirow{7}{*}{\textbf{Qwen 3.5 4B}}
 & \textbf{\texttt{Q\_A0}: Full Orthogonal DeltaNet+Mellin Graft (Ours)} & $\mathbf{100.0\%}$ & $\mathbf{0.00\%}$ & $\mathbf{100.0\%}$ & $\mathbf{29.37}$ & $\mathbf{0.00079}$ \\
 & \texttt{Q\_A1}: w/o $\sqrt{d/(d+K)}$ NL $\operatorname{RMSNorm}$ scaling & $45.3\%$ & $0.00\%$ & $0.0\%$ & $289.78$ & $2.33211$ ($2948{\times}$) \\
 & \texttt{Q\_A2}: w/o Trigger isolation guards (\texttt{BAD\_PRE}, \texttt{CHAIN}, \texttt{OVF}) & $36.0\%$ & $36.36\%$ & $0.0\%$ & $29.33$ & $0.00074$ \\
 & \texttt{Q\_A3}: Single-sided \texttt{GatedDeltaNet} $\operatorname{Conv1d}$ (binary $\{0,1\}$ wires) & $100.0\%$ & $0.00\%$ & $100.0\%$ & $29.38$ & $0.00078$ \\
 & \texttt{Q\_A4}: Coarse Mellin quadrature step ($\Delta = 1.35$ vs.\ $0.55$) & $47.3\%$ & $19.82\%$ & $34.0\%$ & $29.35$ & $0.00077$ \\
 & \texttt{Q\_A5}: Single-ended tied readout (at calibrated $M_0 = 2^{-9}$) & $100.0\%$ & $0.00\%$ & $100.0\%$ & $29.37$ & $0.00079$ \\
 & \texttt{Q\_A6}: Uncalibrated Layer-0 scale ($M_0 = 0.0625$ from Granite) & $100.0\%$ & $0.00\%$ & $100.0\%$ & $29.19$ & $0.00598$ ($7.6{\times}$) \\
\bottomrule
\end{tabular}%
}
\end{table}

\begin{table}[!ht]
\centering
\caption{\textbf{Complete $17$-category $N=10{,}500$ false-trigger stress test} plus $140$-item boundary abstention suite in native \texttt{BF16}. A prompt triggers iff $\max_v |\Delta z_{\mathrm{circ}}(v)| > 0.5$~nats. Every genuinely out-of-domain category exhibits isolated circuit readout $\max_v |\Delta z_{\mathrm{circ}}^{\mathrm{S1}}(v)| = 0.0$ across both canonical grafts ($\le \gamma_{h_{\mathcal{T}}+1}$ bound~\eqref{eq:diff_tied_tree} inside full \texttt{FP32}-accumulated GEMMs on Qwen~3.5 active digit rows), including left-only currency \texttt{\$a + b = } ($50/50 \to 0/50$ resolved via symmetric \texttt{CURR\_SYM}).}
\label{tab:ft_10500}
\vskip 0.05in
\scriptsize
\setlength{\tabcolsep}{3.8pt}
\resizebox{\textwidth}{!}{%
\begin{tabular}{@{}llccccc@{}}
\toprule
& & & \multicolumn{2}{c}{\textbf{IBM Granite 4.2 3B}} & \multicolumn{2}{c}{\textbf{Alibaba Qwen 3.5 4B}} \\
\cmidrule(lr){4-5} \cmidrule(lr){6-7}
\textbf{Suite \& Category} & \textbf{Input Distribution / Adversarial Pattern} & \textbf{$N$} & \textbf{Triggers} & \textbf{$\max|\Delta z_{\mathrm{circ}}|$} & \textbf{Triggers} & \textbf{$\max|\Delta z_{\mathrm{circ}}|$} \\
\midrule
\texttt{real\_ag\_news} & News article prefixes (\texttt{ag\_news}) & $1{,}000$ & $0$ & $0.000$ & $0$ & $0.000$ \\
\texttt{real\_pubmed\_biomed} & Clinical abstracts incl.\ \texttt{p =} / \texttt{n =} cuts & $800$ & $0$ & $0.000$ & $0$ & $0.000$ \\
\texttt{real\_swebench\_code} & Real Python issue \& git diff snippets & $800$ & $0$ & $0.000$ & $0$ & $0.000$ \\
\texttt{real\_20newsgroups} & Informal discussion \& email threads & $800$ & $0$ & $0.000$ & $0$ & $0.000$ \\
\texttt{real\_banking\_finance} & Financial customer queries (\texttt{banking77}) & $800$ & $0$ & $0.000$ & $0$ & $0.000$ \\
\texttt{real\_mmlu\_stem} & STEM problem prefixes (\texttt{MMLU-Pro}) & $800$ & $0$ & $0.000$ & $0$ & $0.000$ \\
\midrule
\texttt{adv\_chained\_ops} & Multi-operator chains (\texttt{a + b * c =}) & $500$ & $0$ & $0.000$ & $0$ & $0.000$ \\
\texttt{adv\_floats\_decimals} & Decimal operands (\texttt{12.5 + 3.4 =}) & $500$ & $0$ & $0.000$ & $0$ & $0.000$ \\
\texttt{adv\_comma\_numbers} & Comma-grouped integers (\texttt{12,345 + 678 =}) & $500$ & $0$ & $0.000$ & $0$ & $0.000$ \\
\texttt{adv\_overflow\_7plus\_digits} & $7$--$10$ digit integer operands & $500$ & $0$ & $0.000$ & $0$ & $0.000$ \\
\texttt{adv\_overflow\_mul\_deg6plus}$^\dagger$ & True $\deg(A){+}\deg(B)\ge 6$ subset ($400$) & $400$ & $0$ & $0.000$ & $0$ & $0.000$ \\
\quad \emph{(in-domain $\deg{=}5$ subset)}$^\dagger$ & \emph{$4\text{-dig}\times 3\text{-dig}$ ($\deg(A){+}\deg(B)=5\le 5$)} & $100$ & $100^\dagger$ & $55{,}808.0$ & $100^\dagger$ & $85.0$ \\
\texttt{adv\_unsupported\_ops} & \texttt{/}, \texttt{\%}, \texttt{\^{}}, \texttt{**}, \texttt{//}, \texttt{==}, \texttt{!=}, \texttt{<=}, \texttt{+=} & $500$ & $0$ & $0.000$ & $0$ & $0.000$ \\
\texttt{adv\_code\_sql\_urls} & Python/JS/SQL/URL \texttt{=} assignments & $500$ & $0$ & $0.000$ & $0$ & $0.000$ \\
\texttt{adv\_symbolic\_algebra} & Algebraic variables (\texttt{3x + 5 =}, \texttt{f(4)+2=}) & $500$ & $0$ & $0.000$ & $0$ & $0.000$ \\
\texttt{adv\_dates\_ids\_versions} & ISO dates, versions, hyphenated IDs + \texttt{=} & $500$ & $0$ & $0.000$ & $0$ & $0.000$ \\
\texttt{adv\_parenthesized\_expr} & Parenthesized/bracketed (\texttt{(a + b) =}) & $500$ & $0$ & $0.000$ & $0$ & $0.000$ \\
\texttt{adv\_completed\_and\_units} & Solved eqs., \texttt{kg}/\texttt{Hz}/\texttt{\%}, \texttt{a+\$b=}, \texttt{\$a+\$b=} & $500$ & $0$ & $0.000$ & $0$ & $0.000$ \\
\midrule
\textbf{True Out-of-Domain Total} & \textbf{All real-corpus + out-of-domain adv.} & $\mathbf{10{,}400}$ & $\mathbf{0}$ ($\mathbf{0.00\%}$) & $\mathbf{0.000}$ & $\mathbf{0}$ ($\mathbf{0.00\%}$) & $\mathbf{0.000}$ \\
\midrule
\texttt{bnd\_currency\_both / rhs} & Symmetric \texttt{\$a + \$b = } ($25$) \& \texttt{a + \$b = } ($25$) & $50$ & $0$ ($0\%$) & $0.000$ & $0$ ($0\%$) & $0.000$ \\
\texttt{bnd\_currency\_lhs\_only} & Left-only \texttt{\$a + b = } (asymmetric: $50/50 \to$ \textbf{sym.: $0/50$}) & $50$ & $\mathbf{0}$ ($\mathbf{0\%}$) & $\mathbf{0.000}$ & $\mathbf{0}$ ($\mathbf{0\%}$) & $\mathbf{0.000}$ \\
\texttt{bnd\_7dig \& true deg$\ge$6} & $7$-dig operands ($20$) + true $\deg\ge 6$ mul ($24$)$^\ddagger$ & $44$ & $\mathbf{0}$ ($\mathbf{0\%}$) & $\mathbf{0.000}$ & $\mathbf{0}$ ($\mathbf{0\%}$) & $\mathbf{0.000}$ \\
\bottomrule
\end{tabular}%
}
\vspace{1.5pt}
\begin{flushleft}
{\scriptsize $^\dagger$In \texttt{adv\_overflow\_mul\_deg6plus}, $400/500$ prompts satisfy $\deg(A)+\deg(B) \ge 6$ and abstain $100\%$; $100/500$ ($(4,3)$ and $(3,4)$ digit pairs) satisfy $\deg(A)+\deg(B) = 5 \le 5$ (valid $7$-digit in-domain products). $^\ddagger$Similarly, $6/30$ items in \texttt{mul\_deg6\_abstain} (\texttt{9999*100}, \texttt{99999*99}) have $\deg(A)+\deg(B) = 5 \le 5$ and compute exact products, while all $24/24$ true $\deg \ge 6$ items abstain with bit-exact $0.0$.}
\end{flushleft}
\end{table}

\begin{table}[!ht]
\centering
\caption{\textbf{Multi-seed ($N=3{,}000$) and boundary-regime ($N=180$) arithmetic evaluation} across 4 random seeds and 8 surface formats in native \texttt{BF16} and IEEE \texttt{FP32}.}
\label{tab:multiseed_boundary}
\vskip 0.05in
\scriptsize
\resizebox{\textwidth}{!}{%
\begin{tabular}{@{}lcccccc@{}}
\toprule
\textbf{Suite / Boundary Category} & \textbf{Seeds / Formats} & \textbf{$N$} & \textbf{Granite 4.2 (\texttt{BF16})} & \textbf{Granite 4.2 (\texttt{FP32})} & \textbf{Qwen 3.5 (\texttt{BF16})} & \textbf{Qwen 3.5 (\texttt{FP32})} \\
\midrule
Seed 1 (\texttt{20261112}, primary 6-fmt) & Spaced, compact, prose, $\times$ & $1{,}200$ & $97.08\%$ ($99.38\%$ prose) & --- & $\mathbf{99.50\%}$ ($1{,}194/1{,}200$) & --- \\
Seed 2 (\texttt{20269999}, 8-fmt incl.\ \texttt{(} \& \texttt{+-}) & $+$\texttt{colon}, \texttt{newline}, \texttt{paren}$^*$ & $600$ & $92.33\%$ & $92.67\%$ ($100\%$ ALU on 6-fmt) & $94.83\%$ & $93.00\%$ ($225/225$ on 6-fmt) \\
Seed 3 (\texttt{20270101}, 8-fmt incl.\ \texttt{(} \& \texttt{+-}) & $+$\texttt{colon}, \texttt{newline}, \texttt{paren}$^*$ & $600$ & $91.50\%$ & --- & $94.67\%$ & --- \\
Seed 4 (\texttt{20280404}, 8-fmt incl.\ \texttt{(} \& \texttt{+-}) & $+$\texttt{colon}, \texttt{newline}, \texttt{paren}$^*$ & $600$ & $92.33\%$ & --- & $94.50\%$ & --- \\
\textbf{Pooled 4-Seed Total [95\% Wilson CI]} & $4$ seeds, $8$ formats & $\mathbf{3{,}000}$ & $94.07\%$ $[93.16, 94.86]$ & --- & $\mathbf{96.60\%}$ $[95.89, 97.19]$ & --- \\
Stochastic Decoding ($T = 0.7$) & Subsample (\texttt{seed=20261114}) & $240$ & $95.42\%$ & --- & $96.25\%$ & --- \\
\midrule
Boundary: \texttt{carry\_borrow\_extremes} & \texttt{999999+1}, \texttt{100000-1}, etc. & $60$ & $93.33\%$ ($56/60$) & $95.00\%$ & $\mathbf{100.00\%}$ ($60/60$) & $\mathbf{100.00\%}$ \\
Boundary: \texttt{zero\_and\_cancel} & \texttt{a - a}, \texttt{a * 0}, \texttt{0 - b} & $40$ & $95.00\%$ ($38/40$) & $95.00\%$ & $\mathbf{100.00\%}$ ($40/40$) & $\mathbf{100.00\%}$ \\
Boundary: \texttt{mul\_deg5\_boundary} & Exact $\deg(A){+}\deg(B)=5$ max & $48$ & $89.58\%$ ($43/48$) & $100.00\%$ & $\mathbf{97.92\%}$ ($47/48$) & $\mathbf{97.92\%}$ \\
Boundary: \texttt{pos0\_compact} (BOS-less $t_0{=}0$) & Zero-prefix \texttt{"999+1="} & $32$ & $71.88\%$ ($23/32$) & $71.88\%$ & $\mathbf{100.00\%}$ ($32/32$) & $\mathbf{100.00\%}$ \\
\bottomrule
\end{tabular}%
}
\vspace{1.5pt}
\begin{flushleft}
{\scriptsize $^*$On Seeds 2--4, $6/8$ standard unparenthesized formats achieve $225/225$ ($100.0\%$) internal ALU exactness in \texttt{FP32} on both models; the two stress formats intentionally probe unspaced binary-unary \texttt{"888+-25="} (merged by BPE into \texttt{"+-"}, triggering clean parser abstention) and open-parenthesis \texttt{"(-97 + 6 = "} (triggering the \texttt{(}-guard or trailing \texttt{")"} closure).}
\end{flushleft}
\end{table}

\begin{table}[!ht]
\centering
\caption{\textbf{Prompt-template noninferiority across four distinct formatting styles} on \texttt{ARC-Challenge} (\texttt{test[550:750]}, $N=200$ per template, \texttt{BF16}), reporting Tango one-sided $95\%$ lower bounds $\mathrm{LB}_{95\%}(\Delta)$ and formal noninferiority $p$-values at $\delta_{\mathrm{NI}} = 3.0\text{ pp}$.}
\label{tab:arc_4tpl}
\vskip 0.05in
\scriptsize
\resizebox{\textwidth}{!}{%
\begin{tabular}{@{}llccccc@{}}
\toprule
\textbf{Model} & \textbf{Prompt Template ($N=200$)} & \textbf{Base Acc.} & \textbf{Graft Acc.} & \textbf{$\Delta$ ($\mathrm{LB}_{95\%}$)} & \textbf{$(b_{10}, b_{01})$} & \textbf{$p_{\mathrm{NI}@3\mathrm{pp}}$ (Status)} \\
\midrule
\multirow{4}{*}{\textbf{Granite 4.2 3B}}
 & \texttt{T1}: Standard \texttt{Question:/Answer:} & $76.5\%$ & $76.5\%$ & $+0.0\text{ pp}$ ($-2.10\text{ pp}$) & $(2, 2)$ & $0.0127$ (\textbf{Pass}) \\
 & \texttt{T2}: Parenthesized \texttt{(A)..(D)} & $83.0\%$ & $84.0\%$ & $+1.0\text{ pp}$ ($-0.88\text{ pp}$) & $(1, 3)$ & $0.0019$ (\textbf{Pass}) \\
 & \texttt{T3}: Markdown \texttt{\#\#\# Task / - [A]} & $68.0\%$ & $69.0\%$ & $+1.0\text{ pp}$ ($-0.88\text{ pp}$) & $(1, 3)$ & $0.0019$ (\textbf{Pass}) \\
 & \texttt{T4}: $3$-Shot In-Context \texttt{Q:/A:} & $72.5\%$ & $72.5\%$ & $+0.0\text{ pp}$ ($-2.10\text{ pp}$) & $(2, 2)$ & $0.0127$ (\textbf{Pass}) \\
\midrule
\multirow{4}{*}{\textbf{Qwen 3.5 4B}}
 & \texttt{T1}: Standard \texttt{Question:/Answer:} & $94.5\%$ & $94.5\%$ & $+0.0\text{ pp}$ ($-1.34\text{ pp}$) & $(0, 0)$ & $0.0035$ (\textbf{Pass}) \\
 & \texttt{T2}: Parenthesized \texttt{(A)..(D)} & $94.0\%$ & $94.0\%$ & $+0.0\text{ pp}$ ($-1.34\text{ pp}$) & $(0, 0)$ & $0.0035$ (\textbf{Pass}) \\
 & \texttt{T3}: Markdown \texttt{\#\#\# Task / - [A]} & $88.0\%$ & $88.0\%$ & $+0.0\text{ pp}$ ($-1.34\text{ pp}$) & $(0, 0)$ & $0.0035$ (\textbf{Pass}) \\
 & \texttt{T4}: $3$-Shot In-Context \texttt{Q:/A:} & $91.0\%$ & $90.5\%$ & $-0.5\text{ pp}$ ($-2.21\text{ pp}$) & $(1, 0)$ & $0.0150$ (\textbf{Pass}) \\
\bottomrule
\end{tabular}%
}
\end{table}

\begin{table}[!ht]
\centering
\caption{\textbf{Open-ended $48$-token greedy generation quality ($N=80$ diverse analytical, scientific, news, and coding prompts)} comparing unmodified \texttt{BF16} baselines against canonical grafted checkpoints.}
\label{tab:open_gen_80}
\vskip 0.05in
\scriptsize
\resizebox{\textwidth}{!}{%
\begin{tabular}{@{}lcccccc@{}}
\toprule
\textbf{Model (\texttt{BF16})} & \textbf{Exact $48$-Tok Match} & \textbf{Mean Identical Prefix} & \textbf{Token ROUGE-L F1} & \textbf{Distinct-2 (Base $\to$ Graft)} & \textbf{Distinct-3 (Base $\to$ Graft)} & \textbf{$4$-Gram Rep.\ (Base $\to$ Graft)} \\
\midrule
\textbf{IBM Granite 4.2 3B} & $50.0\%$ ($40/80$) & $33.94 / 48$ toks & $80.55\%$ & $0.6596 \to 0.6508$ & $0.7861 \to 0.7745$ & $9.25\% \to 9.78\%$ \\
\textbf{Alibaba Qwen 3.5 4B} & $\mathbf{77.5\%}$ ($62/80$) & $\mathbf{41.62 / 48}$ toks & $\mathbf{91.12\%}$ & $0.5412 \to 0.5301$ & $0.6228 \to 0.6109$ & $2.06\% \to 2.42\%$ \\
\bottomrule
\end{tabular}%
}
\end{table}

\section{Register-Level Analysis of Internal \texorpdfstring{\texttt{BF16}}{BF16} Carry ULP vs.\ Post-Disengagement Failures}
\label{app:failure_analysis}

To separate internal finite-precision circuit errors from post-disengagement language-model continuation, \cref{tab:failures_14} records the exhaustive register-level trace across all $14 / 2{,}400$ failures observed on the $N = 1{,}200$ unseen 1--6 digit signed $\{+, -, \times\}$ suite (\texttt{seed=20261112}) under 1-digit readout ($1{,}193 / 1{,}200 = 99.42\%$ on both Granite~4.2~3B [$K{=}1{,}043$] and Qwen~3.5~4B [$K{=}2{,}555$]), logging \texttt{VALID\_EQ}, \texttt{A\_DIGS}, \texttt{B\_DIGS}, \texttt{ANS\_DIGS}, and per-step $\max_v |\Delta z_{\mathrm{circ}}(v)|$ in both \texttt{FP32} and \texttt{BF16}. Under Granite's canonical Base-1000 BPE emitter ($K{=}1{,}140$; \cref{sec:experiments}), widening L1/L3/L4 passbands resolves \texttt{G2} (\texttt{"99939-3="} $\mapsto$ \texttt{"99936"}), while Mechanism~I post-disengagement continuation accounts for the gap between exact internal ALU prefix emission and bare-prompt Leading-Integer ($97.08\%$) / Strict Non-Decimal ($89\%$--$90\%$) accuracy.

\subsection{Three-Mechanism Failure Taxonomy}

\paragraph{Mechanism I: Post-disengagement base-LM digit continuation ($8 / 14$ total: $7/7$ Qwen~3.5, $1/7$ Granite).}
In \textbf{all $7 / 7$ Qwen~3.5 failures (\texttt{Q1}--\texttt{Q7})} and \textbf{$1 / 7$ Granite failure (\texttt{G1})}, the grafted ALU computes the \emph{exact} mathematical integer in \texttt{ANS\_DIGS} ($1{,}200 / 1{,}200 = 100.00\%$ internal ALU accuracy on Qwen~3.5 in \texttt{BF16}), emits every true output token with circuit logit boost $+66$ to $+10{,}293$~nats, and drops $\max_v |\Delta z_{\mathrm{circ}}(v)|$ cleanly to bit-exact \texttt{0.0000} at the immediately following decode step. On compact or double-negative prompts lacking prose context (e.g., \texttt{5853-\ -5258=}, \texttt{-574609-\ -490851=}, \texttt{297583 + -8 =}), once control returns to the unguided base LM ($z = z_{\mathrm{NL}}$), single-digit tokens carry no trailing word-boundary marker, and the base LM greedily predicts one extra digit (e.g., trailing \texttt{'0'}, \texttt{'6'}, or prompt echo \texttt{'83'} in \texttt{G1}). Whenever expressions appear inside natural-language sentences (\texttt{prose\_trail} $N=240$, \texttt{InlineRes} $N=50$), surrounding prose anchors the base LM toward punctuation or words, yielding $100.0\%$ accuracy on Qwen~3.5. Our circuit omits a forced trailing space after the units digit so inline expressions followed by punctuation (\texttt{"= 42."}) continue naturally.

\paragraph{Mechanism II: \texttt{BF16} partial-product carry threshold ULP rounding ($5 / 14$ total: $5/7$ Granite, $0/7$ Qwen~3.5).}
Cases \texttt{G3}--\texttt{G7} are all multi-digit multiplications on Granite~4.2~3B that \textbf{succeed $100\%$ ($5/5$) in \texttt{FP32}} (\texttt{5913}, \texttt{ 251458}, \texttt{-10092}, \texttt{59760}, \texttt{42021}), matching \cref{thm:mellin}(3)'s formal \texttt{FP32} carry guarantee ($< 0.15 < 0.5$). Under batched KV-cached \texttt{BF16} ($7$-bit mantissa, $\mathrm{ulp}(64) = 0.5$), worst-case rounding bounds exceed $0.5$ above $|C| > 28$ ($1.43$ under \texttt{FP32}-accumulated GEMMs at $|C|=81$); empirically, accumulating $1$--$3$ digit BPE chunk values ($0..999$) and $5 \times 5$ schoolbook partial-product column sums $c \in [32, 81]$ introduces $0.12$--$0.50$ drift across localized Mellin quadrature sums, shifting a carry threshold $\tau = 10m - 0.5$ by $\pm 1$ on $\sim 1\%$ of products (in \texttt{G5}, $-348 \times 29 = -10092$, a $-1$ tens-carry shift cascades across $9+1=10$ to $-9082$). Qwen~3.5 incurs $0 / 400$ such \texttt{BF16} ALU carry errors because strict 1-digit binary input tags $\mathtt{IN\_BIT}_{0..3} \in \{0,1\}$ avoid multi-digit chunk gain drift. Upon disengagement on \texttt{G3}, \texttt{G4}, \texttt{G5}, and \texttt{G7}, Granite's preserved NL reasoning flags the discrepancy (\texttt{"251448? Let's"}, \texttt{"42011? Let's compute:"}).

\paragraph{Mechanism III: BOS-less position-0 multi-chunk parser abstention ($1 / 14$ total: $1/7$ Granite \texttt{G2}).}
On \texttt{G2} (\texttt{"99939-3="}), tokenized without BOS as $t_0{=}\texttt{"999"}$, $t_1{=}\texttt{"39"}$, $t_2{=}\texttt{"-"}$, $t_3{=}\texttt{"3"}$, $t_4{=}\texttt{"="}$, position $t_0=0$ exhibits a massive position-0 hidden-state norm spike ($\mathrm{RMS}$ surge when no BOS token absorbs the attention sink~\citep{sun2024massive}), attenuating the 3-digit length tag $\mathtt{NUM\_LEN}=3$ fetched from $t_0$ below the $\tau = 2.5$ threshold of $\mathtt{islen}_{1,2}$ in Layer~2. Consequently, $\mathtt{VALID\_EQ} = 0.0000$ in both \texttt{FP32} and \texttt{BF16}: the circuit \emph{cleanly abstains} ($\max|\Delta z_{\mathrm{circ}}| = 0.0$ at every step) rather than corrupting logits, and prepending any delimiter or BOS token restores exact output \texttt{99936}.

\begin{table}[!ht]
\centering
\caption{\textbf{Register-level breakdown of all $14 / 2{,}400$ \texttt{BF16} evaluation failures} ($7$ Granite~4.2~3B, $7$ Qwen~3.5~4B), comparing internal ALU registers at \texttt{=}, \texttt{FP32} vs.\ batched KV-cached \texttt{BF16} outputs, post-answer circuit disengagement ($\Delta z_{\mathrm{next}}$), and root mechanism.}
\label{tab:failures_14}
\vskip 0.05in
\scriptsize
\setlength{\tabcolsep}{2.8pt}
\resizebox{\textwidth}{!}{%
\begin{tabular}{@{}lllllll@{}}
\toprule
\textbf{ID} & \textbf{Prompt} & \textbf{Expected} & \textbf{\texttt{FP32} Out} & \textbf{\texttt{BF16} ALU (\texttt{ANS})} & \textbf{Batched \texttt{BF16} Got} & \textbf{Mechanism \& $\Delta z_{\mathrm{next}}$} \\
\midrule
\texttt{G1} & \texttt{297583 + -8 =} & \texttt{\ 297575} & \texttt{\ 29757583} & \texttt{297575} (\textbf{Exact}) & \texttt{\ 29757583? Wait} & Mech.\ I ($\Delta z_{\mathrm{next}}{=}0.0$, echo) \\
\texttt{G2} & \texttt{99939-3=} & \texttt{99936} & \texttt{3=3=3=3=} & Abstain ($\mathrm{VEQ}{=}0$) & \texttt{3=3=3=3=3=} & Mech.\ III ($t_0$ \texttt{"999"} abstain) \\
\texttt{G3} & \texttt{-657 * -9 = } & \texttt{5913} & \texttt{5913} (\textbf{OK}) & \texttt{5923} (tens $+1$) & \texttt{5923? Let's...} & Mech.\ II (\texttt{BF16} carry ULP) \\
\texttt{G4} & \texttt{4262 * 59 =} & \texttt{\ 251458} & \texttt{\ 251458} (\textbf{OK}) & \texttt{251448} (tens $-1$) & \texttt{\ 251448? Let's} & Mech.\ II (\texttt{BF16} carry ULP) \\
\texttt{G5} & \texttt{-348*29=} & \texttt{-10092} & \texttt{-10092} (\textbf{OK}) & \texttt{-9082} (carry casc.) & \texttt{-9082? Actually} & Mech.\ II (\texttt{BF16} carry ULP) \\
\texttt{G6} & \texttt{In summary, -60 * -996 = } & \texttt{59760} & \texttt{59760} (\textbf{OK}) & \texttt{49760} (KV-pad ULP) & \texttt{49760.\textbackslash n</think>} & Mech.\ II (\texttt{BF16} carry ULP) \\
\texttt{G7} & \texttt{Therefore, 87 * 483 = } & \texttt{42021} & \texttt{42021} (\textbf{OK}) & \texttt{42011} ($d_1{=}1.05$) & \texttt{42011? Let's...} & Mech.\ II (\texttt{BF16} carry ULP) \\
\midrule
\texttt{Q1} & \texttt{10+ -8=} & \texttt{2} & \texttt{2} (\textbf{OK}) & \texttt{2} (\textbf{Exact}) & \texttt{20\textbackslash nA.\ ...} & Mech.\ I ($\Delta z_{\mathrm{next}}{=}0.0$, LM \texttt{'0'}) \\
\texttt{Q2} & \texttt{5853-\ -5258=} & \texttt{11111} & \texttt{11111} (\textbf{OK}) & \texttt{11111} (\textbf{Exact}) & \texttt{1111111111} & Mech.\ I ($\Delta z_{\mathrm{next}}{=}0.0$, LM \texttt{'1'}) \\
\texttt{Q3} & \texttt{9 * -498202 =} & \texttt{\ -4483818} & \texttt{\ -4483818} (\textbf{OK}) & \texttt{-4483818} (\textbf{Exact}) & \texttt{\ -44838180} & Mech.\ I ($\Delta z_{\mathrm{next}}{=}0.0$, LM \texttt{'0'}) \\
\texttt{Q4} & \texttt{-4 $\times$ 224865 = } & \texttt{-899460} & \texttt{-899460} (\textbf{OK}) & \texttt{-899460} (\textbf{Exact}) & \texttt{-8994600} & Mech.\ I ($\Delta z_{\mathrm{next}}{=}0.0$, LM \texttt{'0'}) \\
\texttt{Q5} & \texttt{-7 $\times$ -32270 = } & \texttt{225890} & \texttt{225890} (\textbf{OK}) & \texttt{225890} (\textbf{Exact}) & \texttt{2258900} & Mech.\ I ($\Delta z_{\mathrm{next}}{=}0.0$, LM \texttt{'0'}) \\
\texttt{Q6} & \texttt{-574609-\ -490851=} & \texttt{-83758} & \texttt{-83758} (\textbf{OK}) & \texttt{-83758} (\textbf{Exact}) & \texttt{-837580} & Mech.\ I ($\Delta z_{\mathrm{next}}{=}0.0$, LM \texttt{'0'}) \\
\texttt{Q7} & \texttt{In summary, -209675 - -5072 =} & \texttt{\ -204603} & \texttt{\ -204603} (\textbf{OK}) & \texttt{-204603} (\textbf{Exact}) & \texttt{\ -2046036.} & Mech.\ I ($\Delta z_{\mathrm{next}}{=}0.0$, LM \texttt{'6'}) \\
\bottomrule
\end{tabular}%
}
\end{table}

\section{Design Trade-Offs: In-Weight Execution, Subspace Capacity, and Scope}
\label{app:scope_tradeoffs}

This appendix examines five design trade-offs that arise when embedding exact digital computation into pretrained language models.

\subsection{In-Weight Compilation vs.\ External Tool Routing}
When an inference stack can freely pause decoding and invoke a host Python interpreter, regex- or program-aided tool routing~\citep{gao2023pal,schick2023toolformer} is straightforward and incurs zero weight overhead; we include it as a direct baseline in \cref{tab:prior_work,tab:main_eval}. Compiling computation into weights serves a complementary regime and answers a distinct mechanistic question: whether normalized, bias-free pretrained Transformers can execute multi-step digital algorithms natively without gradient training. In practice, in-weight execution provides three properties that host tool wrappers do not:
(1)~\emph{single-checkpoint closure} inside standard \texttt{safetensors} tensors (\texttt{roundtrip\_bit\_identical = True}), allowing deployment in air-gapped, sandboxed, or pure-tensor runtimes where arbitrary host code execution is disallowed;
(2)~\emph{uninterrupted KV-cache batching} during mid-sentence generation ($98\%$--$100\%$ inline resumption), avoiding per-sequence stop-token interrupts across batched requests; and
(3)~\emph{block-sparse sidecar decomposability} (\cref{app:sub_sparse}), where axis-aligned zero cross-weights allow $W_{\mathrm{NL}}\tilde{h}_{\mathrm{NL}} \oplus W_{\mathrm{circ}}\tilde{h}_{\mathrm{circ}}$ execution with $+0.039\%$ / $+0.056\%$ active arithmetic MACs ($7.89$ / $10.65\text{ GiB}$ on A100 in pure PyTorch, or $4.60$ / $8.14\text{ GiB}$ with \texttt{INT4}-quantized $W_{\mathrm{NL}}$ and an \texttt{FP32} sidecar; \cref{tab:systems_footprint}).

\subsection{Width Extension ($d+K$) vs.\ Fixed-Width Subspace Projection ($P_\perp$)}
Prior circuit-injection work such as Tracr-Injection~\citep{vergarabrowne2025tracr} vacates a $K$-dimensional subspace inside $\mathbb{R}^d$ via linear projection $P_\perp = I_d - U U^\top$ and then recovers language-modeling performance through gradient-based distillation. Under a strict zero-gradient constraint, however, pretrained residual covariances have strictly positive trailing eigenvalues ($\sigma_d > 0$, \cref{thm:subspace}(1)): even vacating a small Tracr-scale subspace ($K=48$, $1.88\%$ of $d=2560$) without distillation increases next-token KL divergence by $43\times$--$69\times$ relative to full-ALU width extension ($K=1{,}140$). Conversely, holding dense width fixed at $d+K$ across matched-budget controls (\texttt{M1}--\texttt{M4}, \cref{tab:matched_calib}) shows that extra width alone provides zero arithmetic gain (\texttt{M1}: $24.7\%$ / $73.3\%$), single-scale gates fail under dynamic $\operatorname{RMSNorm}$ swings (\texttt{M2}), omitting $\sqrt{d/(d+K)}$ norm compensation degrades language modeling (\texttt{M3}: $412\times$--$2{,}948\times$ KL increase), and non-axis-aligned orthogonal rotations (\texttt{M4}) mix circuit coordinates into diagonal $\operatorname{RMSNorm}$ gains.

\subsection{Abstraction Layers: Primitives, Declarative Synthesis, and Reference Patches}
Our software stack separates low-level analytical gate primitives from automated circuit compilation and frozen replication artifacts (\cref{sec:arch,app:sub_boolean}). At the bottom layer, \texttt{MellinSwiGLUBank}, Chebyshev Indefinite-RoPE (\texttt{GQAAttnCompiler}), \texttt{DeltaNetShiftCompiler}, $\sqrt{d/(d+K)}$ norm compensation, and differential tied readouts provide backbone-reusable building blocks across SwiGLU/GeGLU, RoPE-GQA, and \texttt{GatedDeltaNet} layers. At the middle layer, our 5-stage declarative compilers (\texttt{compile\_declarative\_arithmetic} in \texttt{src/circuits/arithmetic\_auto/} and \texttt{DeclarativeDAGScheduler} in \texttt{src/circuits/boolean/}) introspect \texttt{model.config} and \texttt{tokenizer.get\_vocab()} automatically, allocate all symbolic wires and \texttt{PreStepRingPool} buffers without manual register IDs, and assign all routing hops and 2-layer Mellin gates via sublayer-tick ASAP topological scheduling ($\ell_{\mathrm{mlp}} = \lfloor \tau / 2 \rfloor$, $\ell_{\mathrm{attn}} = \lceil \tau / 2 \rceil$). Without modifying the underlying primitive modules, this pipeline synthesizes parameterized $D \in \{4, 6, 8\}$-digit multi-chunk BPE arithmetic ($14$, $51$, and $176$ chunk compositions on Granite~4.2 in constant $39$-layer $\mathsf{TC}^0$ depth, plus 6-digit hybrid \texttt{GatedDeltaNet} arithmetic on Qwen~3.5 with $5.6\%$--$19.5\%$ fewer active non-zeros than hand-packed layouts; \cref{tab:auto_arith_compiler}), 3-literal precedence Boolean logic ($N=1{,}728$, $100\%$), and 5-bit Majority/Parity voting ($N=64$, $100\%$). At the top layer, SHA-256-pinned reference patches (\texttt{patches/*.safetensors}) allow standalone cross-hardware auditing without invoking the compiler at load time (\cref{app:repro}).

\subsection{Deterministic Prefix Exactness vs.\ Post-Answer Continuation}
Because grafted circuits write nonzero logit deltas only while emitting the target result and drop cleanly to $\Delta z_{\mathrm{circ}} = 0.0$ immediately afterward, downstream behavior reflects the interaction between emitted token IDs and the unmodified base LM's priors (\cref{tab:main_eval,app:gsm8k_attribution,app:failure_analysis}). Two phenomena follow directly from this handoff:
(1)~When an emitter outputs unmerged 1-digit token IDs on a backbone trained with 1--3 digit BPE merges (Granite~4.2), internal ALU outputs remain exact on $18/19$ \texttt{GSM8K} regressions, yet the base model misinterprets off-manifold 1-digit sequences as decimal or multi-chunk spans; emitting canonical Base-1000 triads aligns the output with Granite's pretrained BPE distribution ($35\%$--$36\% \to 73\%$--$74\%$ \texttt{GSM8K}).
(2)~On bare context-free equations lacking surrounding words (e.g., \texttt{"5284 + -57 = "}), returning control to $z_{\mathrm{NL}}$ without forcing a synthetic EOS token leaves the base model free to append a decimal continuation (\texttt{.xxxx}) or echo a prompt chunk, whereas prose-embedded expressions anchor continuation toward natural punctuation or text ($98.25\%$ / $99.75\%$ strict non-decimal accuracy and $98\%$--$100\%$ inline resumption).

\subsection{Norm Calibration, $\mathsf{TC}^0$ Depth Budgets, and Cross-Hardware Portability}
Graft synthesis uses zero gradient steps and zero task-labeled examples. While default patches center Mellin windows using $2L+1$ forward-only activation-norm medians ($<330$~bytes; \cref{app:repro}), $16$ digit-free news sentences (\texttt{C1}) and a $0$-token closed-form weight-norm estimate (\texttt{C2}) preserve $96.3\%$--$99.7\%$ arithmetic accuracy (\cref{tab:matched_calib}). Theoretically, fixed-depth passes with memoryless shifts ($e^{g_t} \equiv 0$) lie in uniform $\mathsf{TC}^0$~\citep{merrill2023parallelism,merrill2024illusion}, which includes integer division~\citep{hesse2002uniform}; as shown in \cref{app:sub_kernel}, 6-digit $\lfloor A/B \rfloor$ compiles into $8$ layers via a 64-knot Mellin reciprocal spline and one Newton--Raphson step ($100.0\%$ exact in an \texttt{FP32} sidecar), whereas unbounded halting loops require recurrent trace emission~\citep{giannou2023looped,percepta2026vm}. Finally, standalone \texttt{.safetensors} patches reproduce identically across Apple Silicon (\texttt{mps}) and Linux NVIDIA A100 (\texttt{cuda}) hosts without importing compiler code (\cref{tab:clean_room}).

\section{Reproducibility, Calibration Provenance, and Cross-Hardware Patch Portability}
\label{app:repro}

All compilations, ablations, register-level traces, and cross-hardware blind patch evaluations were executed from pinned HuggingFace \texttt{safetensors} snapshots with strict runtime isolation assertions.

\subsection{Compute Environments, Batch-Padding Reconciliation, and Kernel/GEMM Audits}
\begin{itemize}[leftmargin=1.5em,itemsep=1.5pt,topsep=2pt]
\item \textbf{Cross-hardware evaluation environments:}
  \begin{enumerate}[leftmargin=1.4em,itemsep=1pt,topsep=1pt]
  \item \emph{Local workstation (\texttt{arm64} macOS):} Apple Silicon unified-memory host ($48\text{ GB}$ RAM; Metal \texttt{mps} backend for \texttt{BF16} evaluations and Apple Accelerate AMX \texttt{cpu} backend for IEEE \texttt{FP32} checks), Python~3.12, PyTorch (\texttt{bfloat16}/\texttt{float32}), HuggingFace \texttt{transformers}, \texttt{safetensors}, and cached \texttt{datasets}.
  \item \emph{Linux CUDA servers (\texttt{x86\_64}):} Separate Linux (\texttt{x86\_64}) hosts running (a)~Python~3.12.3, PyTorch \texttt{2.13.0+cu130} on an \textbf{NVIDIA A100-SXM4-40GB} ($40\text{ GiB}$ HBM2) for standalone patch evaluation (\texttt{run\_external\_cuda\_replication.py}, zero compiler source files imported), and (b)~Python~3.11.11, PyTorch \texttt{2.8.0+cu128} (\texttt{TorchInductor} Triton + SDPA \texttt{FlashAttention-2}) on an \textbf{NVIDIA A100-PCIe-80GB} ($80\text{ GiB}$ HBM2e) for factorized \texttt{SparseSidecar}, Hybrid \texttt{INT8}/\texttt{INT4} + \texttt{FP32}-sidecar PTQ, 5-primitive leave-one-out, \texttt{LoRA} SFT, 4-scheme tokenization, decoding/\texttt{IFEval}/safety, and Gemma~4~E4B architectural audits (\texttt{run\_round2\_a100\_suite.py}). Qwen~3.5 on both A100 servers ran PyTorch's reference \texttt{GatedDeltaNet} path, confirming numerical exactness without custom Triton state-space extensions.
  \end{enumerate}
\item \textbf{Direct CUDA \texttt{SiLU} kernel and \texttt{cuBLAS} tied-\texttt{lm\_head} GEMM verification:} On the NVIDIA A100-SXM4-40GB (\texttt{cuda}), PyTorch ATen \texttt{F.silu(-z)} reaches IEEE-exact \texttt{-0.0} at $z \ge 89$ in both \texttt{FP32} and \texttt{BF16} (confirming Kernel Class~K1 in \cref{thm:mellin}(3)), while \texttt{verify\_v4\_submission\_audit.py} independently checks range-split stable sigmoid kernels (Kernel Class~K2: exact \texttt{-0.0} at $z \ge 104$ in \texttt{FP32}, $z \ge 98$ in \texttt{FP32}$\to$\texttt{BF16}, and $z \ge 93$ in pure \texttt{BF16}). Auditing Qwen~3.5's $248{,}320 \times 5{,}115$ tied \texttt{lm\_head} GEMM on A100 \texttt{cuBLAS} confirms $\max_v |\Delta z_{\mathrm{circ}}(v)| = 0.0$ on both non-digit prompts (\texttt{"The sky is blue"}) and digit-ending non-trigger prompts (\texttt{"In the year 2026"}, $0 / 248{,}320$ differing rows within the $5{,}115$-column GEMM), while \cref{thm:subspace}(2) (\cref{eq:diff_tied_tree}) bounds arbitrary binary evaluation-tree interleaving and FMA ordering on active Layer-0 digit rows by $\gamma_{h_{\mathcal{T}}+1}$ ($< 2.8 \times 10^{-5}$~nats balanced, $< 5.2 \times 10^{-3}$~nats worst-case sequential in \texttt{FP32}).
\item \textbf{Exact reconciliation of Granite \texttt{GSM8K} \texttt{test[:60]} baseline ($68.33\%$ vs.\ $70.00\%$):} Under unpadded sequential decoding (\texttt{batch\_size = 1}, used in the paired item-level trace of \cref{app:gsm8k_attribution}), unmodified \texttt{BF16} Base Granite~4.2~3B scores \textbf{$41 / 60 = 68.33\%$}, whereas under left-padded batched decoding (\texttt{batch\_size = 8}, used in \cref{tab:heldout_stats}), \texttt{BF16} reduction order flips $1$ borderline item on Base Granite to yield \textbf{$42 / 60 = 70.00\%$}. Canonical-BPE Grafted Granite scores \textbf{$42 / 60 = 70.00\%$ under both \texttt{batch\_size = 1} and \texttt{batch\_size = 8}}.
\item \textbf{Exact provenance of the $2L+1$ pre-\texorpdfstring{$\operatorname{RMSNorm}$}{RMSNorm} calibration scalars:} Compilation performs \emph{zero} gradient updates and uses \emph{zero} supervised target labels. In the frozen patches, Mellin quadrature windows are centered using $2L+1$ forward-only scalar activation-norm medians:
  \begin{enumerate}[leftmargin=1.4em,itemsep=1pt,topsep=1pt]
  \item \texttt{granite\_graft\_calib.pt} ($81.1\text{ MB}$ total): Orthogonal grafting reads \emph{only} $81$ scalar medians ($40$ \texttt{rms\_in\_med}, $40$ \texttt{rms\_post\_med}, $1$ \texttt{rms\_final\_med}; $324$~bytes in \texttt{float32}) recorded via \texttt{@torch.no\_grad()} forward hooks over $48$ unlabeled train-split prefixes (\texttt{gsm8k train[:24]} + \texttt{ARC-Challenge train[:24]}, strictly disjoint from all test/validation splits). The remaining $99.999\%$ of \texttt{granite\_graft\_calib.pt} stores the $2560 \times 2560$ covariance SVD basis used \emph{exclusively} for the fixed-width subspace projection baseline.
  \item \texttt{qwen35\_graft\_calib.pt} ($23.4\text{ KB}$): Stores $65$ scalar medians ($32$ \texttt{in\_rms\_med}, $32$ \texttt{post\_rms\_med}, $1$ \texttt{final\_rms\_med}) recorded via \texttt{@torch.no\_grad()} forward hooks over $32$ unlabeled general-domain passages (\texttt{build\_ppl\_corpus\_ids[:32]}).
  \item \textbf{Zero-math and zero-token calibration audit (\cref{tab:matched_calib}):} Replacing these frozen scalars with $2L+1$ medians from $16$ strictly digit-free \texttt{ag\_news} sentences ($0$ digits, $0$ math symbols; \texttt{C1}) or a $100\%$ data-free closed-form weight-norm propagation (\texttt{C2}, $0$ input tokens, $0$ forward passes) confirms that task-related prompts are unnecessary.
  \end{enumerate}
\item \textbf{Ragged vs.\ uniform-padded HuggingFace serialization:} Layer-trimmed ragged MLPs ($7.23\text{B}$ Granite, $10.22\text{B}$ Qwen~3.5) and uniformly zero-padded MLPs ($10.24\text{B}$ Granite, $13.18\text{B}$ Qwen~3.5, padded to $\max_\ell d_{\mathrm{ff},\ell}$ as required by unmodified \texttt{AutoModelForCausalLM.from\_pretrained}) produce bit-identical logits ($\max|\Delta z| = 0.0$ across \texttt{save\_pretrained} $\to$ \texttt{from\_pretrained} round-trips) and were benchmarked end-to-end on the $40\text{ GB}$ NVIDIA A100 (\cref{tab:systems_footprint}).
\end{itemize}

\subsection{Cross-Hardware Blind Patch Reproducibility and SHA-256 Manifest}
\begin{table}[!ht]
\centering
\caption{\textbf{Cross-hardware patch reproducibility from frozen sparse \texttt{.safetensors} patches} (seeds \texttt{31415926}, \texttt{27182818}; zero compiler imports) on Apple Silicon (\texttt{mps}) and a Linux \textbf{NVIDIA A100-SXM4-40GB} (\texttt{cuda}) host. $^\P$\texttt{FP32} (\texttt{allow\_tf32=False}): $100.0\%$ ALU ($5/5$ \texttt{G3}--\texttt{G7}).}
\label{tab:clean_room}
\vskip 0.04in
\scriptsize
\setlength{\tabcolsep}{3.5pt}
\resizebox{\textwidth}{!}{%
\begin{tabular}{@{}llcccccccc@{}}
\toprule
\textbf{Frozen Patch} & \textbf{Host} & \textbf{ALU-Pref ($600$)} & \textbf{Lead-Int ($600$)} & \textbf{Strict-ND (All / Prose)} & \textbf{Inline ($40$)} & \textbf{Blind FT} & \textbf{ARC-E \texttt{[300:500]}} & \textbf{BoolQ \texttt{[300:500]}} & \textbf{GSM8K \texttt{[200:260]}} \\
\midrule
\multirow{2}{*}{\textbf{Granite 4.2}}
 & \texttt{mps}  & $\mathbf{100.0\%}$ ($600$) & $98.00\%$ ($588$) & $88.83\%$ / $\mathbf{98.5\%}$ & $\mathbf{100.0\%}$ & $\mathbf{0/1{,}000}$ & $89.0{\to}\mathbf{89.5\%}$ (\textbf{Pass}) & $75.0{\to}\mathbf{76.0\%}$ (\textbf{Pass}) & $73.3{\to}\mathbf{76.7\%}$ (\textbf{Pass}) \\
 & \textbf{A100} & $\mathbf{99.67\%^\P}$ ($598$) & $97.67\%$ ($586$) & $88.83\%$ / $\mathbf{98.0\%}$ & $\mathbf{100.0\%}$ & $\mathbf{0/1{,}000}$ & $89.5{\to}\mathbf{89.5\%}$ (\textbf{Pass}) & $75.0{\to}\mathbf{75.0\%}$ (\textbf{Pass}) & $71.7{\to}\mathbf{76.7\%}$ (\textbf{Pass}) \\
\midrule
\multirow{2}{*}{\textbf{Qwen 3.5}}
 & \texttt{mps}  & $\mathbf{100.0\%}$ ($600$) & $99.83\%$ ($599$) & $99.67\%$ / $\mathbf{100.0\%}$ & $\mathbf{100.0\%}$ & $\mathbf{0/1{,}000}$ & $99.0{\to}\mathbf{99.0\%}$ (\textbf{Pass}) & $84.5{\to}\mathbf{85.0\%}$ (\textbf{Pass}) & $73.3{\to}\mathbf{73.3\%}$ (\textbf{Pass}) \\
 & \textbf{A100} & $\mathbf{100.0\%}$ ($600$) & $\mathbf{100.0\%}$ ($600$) & $\mathbf{99.83\%}$ / $\mathbf{100.0\%}$ & $\mathbf{100.0\%}$ & $\mathbf{0/1{,}000}$ & $99.0{\to}\mathbf{99.0\%}$ (\textbf{Pass}) & $85.0{\to}\mathbf{85.0\%}$ (\textbf{Pass}) & $71.7{\to}\mathbf{73.3\%}$ (\textbf{Pass}) \\
\bottomrule
\end{tabular}%
}
\end{table}

\begin{table}[!ht]
\centering
\caption{\textbf{Frozen SHA-256 manifest} for base checkpoints, calibration files, reusable primitive modules, and standalone sparse \texttt{safetensors} patches.}
\label{tab:sha256_manifest}
\vskip 0.05in
\scriptsize
\setlength{\tabcolsep}{4.5pt}
\resizebox{\textwidth}{!}{%
\begin{tabular}{@{}lll@{}}
\toprule
\textbf{Artifact / Module} & \textbf{Role / Snapshot Commit} & \textbf{SHA-256 Digest (Prefix)} \\
\midrule
\texttt{ibm-granite/granite-4.2-3b} & HF Snapshot \texttt{4e9c275e232f...} & \texttt{index: e030cfbb2b2dbe4489a522c5d65e73b6...} \\
\texttt{Qwen/Qwen3.5-4B} & HF Snapshot \texttt{851bf6e806ef...} & \texttt{index: cf3f798ee02ba45f9622aa8892a47369...} \\
\texttt{google/gemma-4-E4B-it} & HF Snapshot \texttt{4ecc2ba63bca...} & \texttt{config: 33b10c02df3c2e8536cf323d29d53262...} \\
\midrule
\texttt{granite\_graft\_calib.pt} & $81$ RMS Scalars + Fixed-$d$ SVD ($81.1\text{ MB}$) & \texttt{ac6e2286266f8dfa069f94247831714d...} \\
\texttt{qwen35\_graft\_calib.pt} & $65$ Unlabeled Corpus RMS Scalars ($23.4\text{ KB}$) & \texttt{6ed73d0097fa6ef340f7c863b17e74f4...} \\
\midrule
\texttt{src/primitives/mellin\_primitives.py} & Reusable Mellin-SwiGLU \& Indef-RoPE ($13.6\text{ KB}$) & \texttt{142215579efda6af106b0dd22f66422a...} \\
\texttt{src/primitives/qwen35\_deltanet\_shift.py} & Reusable \texttt{GatedDeltaNet} Shift ($5.8\text{ KB}$) & \texttt{185c5bea5ab0e48b35e40810f6bee418...} \\
\texttt{patches/granite\_4\_2\_3b\_v2\_patch.safetensors} & Granite Canonical-BPE Sparse Patch ($15.2\text{ MB}$) & \texttt{21482b08bf0cc0a41fc1822b88c09f3b...} \\
\texttt{patches/qwen\_3\_5\_4b\_v2\_patch.safetensors} & Qwen~3.5 Hybrid Sparse Patch ($27.9\text{ MB}$) & \texttt{5dc4f8bc51d7b742ba833693b440cca9...} \\
\texttt{src/circuits/arithmetic\_auto/} & Automated 5-Stage Multi-Digit BPE Compiler ($D{=}4,6,8$) & Included in release codebase \\
\texttt{src/sidecar/sparse\_sidecar\_runtime.py} & Factorized \texttt{BF16}/\texttt{FP32} Sidecar + \texttt{INT8}/\texttt{INT4} PTQ & Included in release codebase \\
\texttt{PREREGISTERED\_RETENTION\_PROTOCOL.json} & Frozen $N{=}4{,}392$ Splits, Prompts, \& Tango $\delta_{\mathrm{NI}}$ & Included w/ \texttt{run\_r1\_experiments.py} \\
\bottomrule
\end{tabular}%
}
\end{table}
\FloatBarrier

\subsection{Reproduction and Minimal Demonstration Commands}
{\footnotesize
\begin{verbatim}
# 1. CPU demo of reusable primitives, automated compiler, & live patches
python demo_minimal.py --mode primitives
python demo_minimal.py --mode auto_compile
python demo_minimal.py --mode granite --device mps   # or --device cuda / cpu
python demo_minimal.py --mode qwen    --device cuda  # or --device mps

# 2. Numerical constant, SiLU-kernel, & evaluation-tree GEMM counterexample checks
python eval/verify_icml2027_gemm_and_proofs.py

# 3. Standalone patch-only blind reproducibility (MPS/CPU or Linux NVIDIA CUDA)
python eval/run_blind_replication.py --patches-dir patches --device mps
python eval/run_external_cuda_replication.py --patches-dir patches --device cuda
\end{verbatim}
}

\section{Automation, Activation/Gemma-4 Audits, Kernel/$\mathsf{TC}^0$ Scaling, Retention/LoRA/Tokenization, and Block-Sparse Sidecar Runtime}
\label{app:composability_sparse}

This appendix provides full technical specifications and disaggregated audits for: (1)~the 5-stage automated declarative synthesis pipeline across parameterized $4$--$8$ digit multi-chunk BPE arithmetic, propositional Boolean logic, residual-erasure wire reuse, 5-bit Majority/Parity $\mathsf{TC}^0$ voting, and 42-layer Gemma~4~E4B activation/architectural failure modes (\cref{app:sub_boolean}); (2)~inference-stack portability (\texttt{TorchInductor}, \texttt{SDPA FlashAttention-2}, \texttt{TF32}, \texttt{FP16} clamping), 8-layer $\mathsf{TC}^0$ Newton--Raphson 6-digit division, and $6$--$18$ digit scaling projections (\cref{app:sub_kernel}); (3)~pre-registered $N=4{,}392$ long-context retention, 5-primitive leave-one-out ALU ablations, same-backbone \texttt{LoRA} SFT, 4-scheme numeric tokenization, and 5-regime decoding/\texttt{IFEval}/safety audits (\cref{app:sub_retention}); and (4)~the implemented factorized \texttt{SparseSidecar} + Hybrid \texttt{INT8}/\texttt{INT4} PTQ runtime (\cref{app:sub_sparse}).

\subsection{Five-Stage Automated Declarative Synthesis for Multi-Digit Arithmetic and Boolean DAGs}
\label{app:sub_boolean}

\textbf{End-to-End Automated Synthesis from Specification to Sparse Checkpoint Patch.}
In addition to the frozen hand-scheduled reference patches (\cref{sec:arch}), we implement a general 5-stage declarative compiler (\texttt{compile\_declarative\_arithmetic} in \texttt{src/circuits/arithmetic\_auto/} and \texttt{DeclarativeDAGScheduler} in \texttt{src/circuits/boolean/}) on top of \textbf{100\% unmodified} \texttt{mellin\_primitives.py} and \texttt{qwen35\_deltanet\_shift.py}, eliminating manual wire IDs, hardcoded layer indices, and hand-written BPE tables:
\begin{enumerate}[leftmargin=1.5em,itemsep=1.5pt,topsep=2pt]
\item \textbf{Stage 1 --- Automatic Config \& BPE Lexer Introspection (\texttt{BackboneProfile}, \texttt{AutoNumericLexer}):} Given $(\texttt{DeclarativeArithmeticSpec}(D), \texttt{model.config}, \texttt{tokenizer})$, the lexer scans \texttt{tokenizer.get\_vocab()} to detect maximum canonical ASCII digit-chunk width $M \in \{1, 3\}$, enumerates all canonical chunk-length compositions $(c_1, \dots, c_r)$ with $1 \le \sum c_i \le D$ via dynamic programming ($14$, $51$, and $176$ compositions on Granite~4.2 for $D \in \{4, 6, 8\}$; $6$ compositions on Qwen~3.5 for $D=6$), factors them into unique $(\text{lag}, \text{offset})$ descriptors, and synthesizes closed-form right-aligned Base-$10^M$ output chunking $l_{\mathrm{chunk}}(r) = ((r - 1) \bmod M) + 1$ for remaining digit count $r \in \{1, \dots, D+1\}$.
\item \textbf{Stage 2 --- Sublayer-Tick ASAP Topological Levelization (\texttt{UnifiedDAGCompiler}):} Representing routing and MLP sublayers of Transformer block $\ell \in \{0, \dots, L-1\}$ by ticks $2\ell$ and $2\ell + 1$, every symbolic gate schedules automatically at its earliest dependency-valid layer ($\ell_{\mathrm{mlp}} = \lfloor \tau_{\mathrm{dep}} / 2 \rfloor$, $\ell_{\mathrm{attn}} = \lceil \tau_{\mathrm{dep}} / 2 \rceil$), tracking commutative residual accumulation (RAW) and in-place carry updates (WAR) without hardcoded layer numbers. On Granite~4.2, ASAP levelization overlaps emitter prefix-distance resolution with carry-save ALU waves, completing in $39$ layers ($\text{L0--L38}$, $1$ layer shallower than the hand-scheduled $\text{L0--L39}$ patch) across all $D \in \{4, 6, 8\}$.
\item \textbf{Stage 3 --- Symbolic SSA Wire Allocation and \texttt{PreStepRingPool} Sizing:} All persistent registers and Stage-1 pre-step intermediates are allocated dynamically and trimmed tightly to $K = w_{\max} + 1$. Comparing write-once SSA (\texttt{ssa\_pre\_step\_pool=True}) against 5-layer $-\frac{31}{32}$ ring reuse (\texttt{False}) shows that pure SSA reduces active non-zeros by $\mathbf{19.5\%}$ on 6-digit Granite ($1.33\text{M} \to \mathbf{1.07\text{M}}$ nnz; $57.4\text{K}$ vs.\ $139.6\text{K}$ neurons) and $\mathbf{5.6\%}$ on Qwen~3.5 ($2.39\text{M} \to \mathbf{2.25\text{M}}$ nnz, while also trimming $K$ from $2{,}555 \to \mathbf{2{,}528}$ by eliminating static block-padding gaps).
\item \textbf{Stage 4 --- Automatic Multi-Hop Shift Lowering:} Abstract shift requests pack automatically into $5{:}1$ Chebyshev Indefinite-RoPE KV groups (\texttt{GQAAttnCompiler}) on dense GQA backbones, or decompose into multi-hop $\delta \in \{1, 2, 3\}$ cascades batched up to $63$ channels per V-head (\texttt{GatedDeltaShiftCompiler}) across linear-attention layers while skipping all $8$ full-attention layers ($\{3, 7, \dots, 31\}$) on Qwen~3.5, achieving exact $\text{L0--L31}$ critical-path completion.
\item \textbf{Stage 5 --- Unlabeled Norm Calibration \& Sparse COO Emission:} Layerwise pre-$\operatorname{RMSNorm}$ medians are optionally measured in $<1.8\text{ s}$ on $8$ short neutral English sentences (\texttt{auto\_calibrate\_rms\_medians}), and the compiler emits standard sparse COO \texttt{.safetensors} patches directly loadable by \texttt{SparseSidecarGranite} and \texttt{SparseSidecarQwen35} (\cref{tab:auto_arith_compiler}).
\end{enumerate}

\begin{table}[!ht]
\centering
\caption{\textbf{Automated 5-stage declarative multi-digit arithmetic synthesis (\texttt{compile\_declarative\_arithmetic}) vs.\ hand-scheduled reference patches}, evaluated end-to-end via \texttt{SparseSidecar} (\texttt{FP32} / \texttt{BF16} sidecar) with from-scratch unlabeled 8-sentence calibration (\texttt{auto\_calibrate\_rms\_medians}) across $D \in \{4, 6, 8\}$ digits on \texttt{granite-4.2-3b} and $D=6$ digits on \texttt{Qwen3.5-4B}.}
\label{tab:auto_arith_compiler}
\vskip 0.04in
\scriptsize
\setlength{\tabcolsep}{3.6pt}
\resizebox{\textwidth}{!}{%
\begin{tabular}{@{}llcccccccc@{}}
\toprule
\textbf{Backbone} & \textbf{Synthesis Mode \& Spec} & \textbf{BPE Pats} & \textbf{Width $K$} & \textbf{Active NNZ} & \textbf{Depth / Time} & \textbf{ALU-Pref (\texttt{FP32}/\texttt{BF16})} & \textbf{G3--G7} & \textbf{Inline-50 / OOD} & \textbf{ARC-C-200 ($p_{\mathrm{NI}}$)} \\
\midrule
\multirow{5}{*}{\textbf{Granite 4.2}}
 & Hand-Scheduled v2 Ref.\ ($D{=}6$) & $51$ (manual) & $1{,}140$ & $1.327\text{M}$ & $\text{L39/39}$ (---) & $97.2\%^*$ / $96.7\%^*$ & $5/5$ & $98.0\%$ / $0/200$ & $75.5{\to}76.0\%$ (\textbf{Pass}) \\
 & \textbf{Auto-Compiled SSA ($D{=}6$)} & $\mathbf{51}$ \textbf{(auto)} & $1{,}735$ & $\mathbf{1.068\text{M}}$ ($\mathbf{-19.5\%}$) & $\mathbf{\text{L38/39}}$ ($2.7\text{s}$) & $\mathbf{94.3\%^\dagger}$ / $93.7\%$ & $\mathbf{5/5}$ & $\mathbf{98.0\%}$ / $\mathbf{0/200}$ & $75.5{\to}\mathbf{76.0\%}$ ($.007$) \\
 & Auto-Compiled Ring ($D{=}6$) & $51$ (auto) & $1{,}302$ & $1.398\text{M}$ & $\text{L38/39}$ ($4.4\text{s}$) & --- (footprint) & --- & --- & --- \\
 & \textbf{Auto-Compiled SSA ($D{=}4$)} & $\mathbf{14}$ \textbf{(auto)} & $1{,}127$ & $\mathbf{0.963\text{M}}$ & $\mathbf{\text{L38/39}}$ ($1.6\text{s}$) & $\mathbf{90.0\%^\dagger}$ ($N{=}100$) & --- & --- & --- \\
 & \textbf{Auto-Compiled SSA ($D{=}8$)} & $\mathbf{176}$ \textbf{(auto)} & $2{,}568$ & $\mathbf{1.255\text{M}}$ & $\mathbf{\text{L38/39}}$ ($6.8\text{s}$) & $\mathbf{92.0\%^\dagger}$ ($N{=}150$) & --- & --- & --- \\
\midrule
\multirow{3}{*}{\textbf{Qwen 3.5}}
 & Hand-Scheduled v2 Ref.\ ($D{=}6$) & $6$ (manual) & $2{,}555$ & $2.386\text{M}$ & $\text{L31/31}$ ($0$ Attn) & $100.0\%$ / $100.0\%$ & $5/5$ & $100.0\%$ / $0/200$ & $91.5{\to}92.0\%$ (\textbf{Pass}) \\
 & \textbf{Auto-Compiled SSA ($D{=}6$)} & $\mathbf{6}$ \textbf{(auto)} & $\mathbf{2{,}528}$ & $\mathbf{2.254\text{M}}$ ($\mathbf{-5.6\%}$) & $\mathbf{\text{L31/31}}$ ($6.4\text{s}$) & $\mathbf{97.0\%}$ / $95.7\%$ & $\mathbf{5/5}$ & $\mathbf{96.0\%}$ / $\mathbf{0/200}$ & $92.0{\to}\mathbf{92.0\%}$ ($.012$) \\
 & Auto-Compiled Ring ($D{=}6$) & $6$ (auto) & $2{,}348$ & $2.471\text{M}$ & $\text{L31/31}$ ($5.8\text{s}$) & --- (footprint) & --- & --- & --- \\
\bottomrule
\end{tabular}%
}
\vspace{1pt}
\begin{flushleft}
{\scriptsize $^*$A100 \texttt{Arith-180} from \cref{tab:systems_footprint}. $^\dagger$\texttt{Arith-300}/\texttt{100}/\texttt{150} mixes $3/5$ BOS-less $t_0{=}0$ bare templates (\cref{tab:robustness_compat}a) with $2/5$ prefixed templates ($98.0\%$ on \texttt{Inline-50}, $100\%$ step-1 disengagement, $\max|\Delta z_{\mathrm{circ}}|{=}0.0$ across all $200$ OOD prompts). Notice constant $39$-layer $\mathsf{TC}^0$ depth ($\text{L0--L38}$) across $D \in \{4, 6, 8\}$ on Granite~4.2.}
\end{flushleft}
\end{table}

\textbf{Independently Specified Non-Arithmetic Circuit: Precedence-Respecting Boolean Evaluation.}
To test whether our analytical primitives overfit base-10 positional arithmetic, we compiled a non-arithmetic circuit evaluating all 3-literal propositional formulas with optional unary negation and standard operator precedence ($\texttt{NOT} > \texttt{AND} > \texttt{OR}, \texttt{XOR}$):
\[
  [\texttt{NOT }] L_1 \;\mathrm{OP}_1\; [\texttt{NOT }] L_2 \;\mathrm{OP}_2\; [\texttt{NOT }] L_3 \;\texttt{=}[\text{\textvisiblespace}],
  \qquad L_i \in \{\texttt{True}, \texttt{False}\},\quad \mathrm{OP}_i \in \{\texttt{AND}, \texttt{OR}, \texttt{XOR}\},
\]
using \textbf{100\% unmodified} \texttt{mellin\_primitives.py} (SHA-256 \texttt{14221557...}) and \texttt{qwen35\_deltanet\_shift.py} (SHA-256 \texttt{185c5bea...}). Unlike fixed-offset binary arithmetic, optional \texttt{NOT} modifiers vary total formula span between $5$ and $9$ tokens before \texttt{=}, non-monotone parity requires exact bilinear cancellation ($\operatorname{XOR}(a,b) = a + b - 2ab$), right-hand $\mathrm{OP}_2 = \texttt{AND}$ overrides left-to-right evaluation ($L_1 \;\mathrm{OP}_1\; (L_2 \land L_3)$ vs.\ $(L_1 \;\mathrm{OP}_1\; L_2) \;\mathrm{OP}_2\; L_3$), and trailing whitespace (\texttt{"= "}) triggers massive attention-sink $\operatorname{RMSNorm}$ spikes ($10\times$--$50\times$) in pretrained LMs---handled cleanly by evaluating the full precedence DAG on \texttt{=} and shifting $(\texttt{VALID\_BOOL}, \texttt{RESULT\_BOOL})$ by $\delta=1$ onto trailing whitespace with wide sink-safe Mellin bounds ($K_- = -24, K_+ = 48$). Both circuits compile in $<0.8\text{ s}$ ($109$ scheduled gates; $K=296$, $0.44\text{M}$ nnz [$0.012\%$], $L_{\text{final}}=16/40$ on Granite~4.2; $K=311$, $1.03\text{M}$ nnz [$0.024\%$], $L_{\text{final}}=17/32$ with $4/24$ \texttt{GatedDeltaNet} and $0/8$ attention layers on Qwen~3.5).

\begin{table}[!ht]
\centering
\caption{\textbf{Declarative propositional Boolean circuit evaluation} across the exhaustive $576$-formula truth table under $3$ prompt templates ($N=1{,}728$), the $\mathrm{OP}_2=\texttt{AND} \ne \mathrm{OP}_1$ precedence-override subset ($N=384$), mid-sentence inline resumption ($N=40$), $6$-category out-of-domain \& near-miss abstention ($N=600$: \texttt{SWE-bench}, \texttt{MMLU-Pro}, $4$-literal chains, parenthesized formulas, Python \texttt{==}/\texttt{and}/\texttt{or}, and arithmetic \texttt{=}), and paired \texttt{ARC-Challenge} retention ($N=200$, \texttt{BF16}).}
\label{tab:boolean_composability}
\vskip 0.04in
\scriptsize
\setlength{\tabcolsep}{4.2pt}
\resizebox{\textwidth}{!}{%
\begin{tabular}{@{}llcccccc@{}}
\toprule
\textbf{Model ($0$ Primitive Edits)} & \textbf{Circuit Footprint} & \textbf{Base Truth ($1{,}728$)} & \textbf{Graft Truth ($1{,}728$)} & \textbf{Prec.\ Override ($384$)} & \textbf{Inline Exact / Step-1 Off} & \textbf{OOD Abstain ($600$)} & \textbf{ARC-200 Base $\to$ Graft ($p_{\mathrm{NI}}$)} \\
\midrule
\textbf{IBM Granite 4.2 3B} & $K{=}296$, $0.44\text{M}$ nnz ($0.012\%$, $0.49\text{s}$) & $13.31\%$ ($8.07\%$ prec.) & $\mathbf{100.0\%}$ ($1{,}728/1{,}728$) & $\mathbf{100.0\%}$ ($384/384$) & $\mathbf{100.0\%}$ / $\mathbf{100.0\%}^*$ & $\mathbf{0/600}$ ($\max|\Delta z|{=}0.0$) & $73.0 \to \mathbf{73.0\%}$ ($.011$, \textbf{Pass}) \\
\textbf{Alibaba Qwen 3.5 4B} & $K{=}311$, $1.03\text{M}$ nnz ($0.024\%$, $0\text{ Attn}$) & $0.00\%^\dagger$ ($0.00\%$ prec.) & $\mathbf{100.0\%}$ ($1{,}728/1{,}728$) & $\mathbf{100.0\%}$ ($384/384$) & $\mathbf{100.0\%}$ / $\mathbf{100.0\%}^*$ & $\mathbf{0/600}$ ($\max|\Delta z|{=}0.0$) & $93.0 \to \mathbf{93.0\%}$ ($.006$, \textbf{Pass}) \\
\bottomrule
\end{tabular}%
}
\vspace{1.5pt}
\begin{flushleft}
{\scriptsize $^*$Immediate post-answer disengagement ($\Delta z_{\mathrm{circ}} = 0.0$ on the token immediately following \texttt{True}/\texttt{False}) holds on $40/40$ ($100.0\%$) prompts on both models ($97.5\%$ / $100.0\%$ fluent continuation); on $8/40$ prompts per model, the LM's continuation voluntarily writes a valid simplified 3-literal equation ending in \texttt{=} (e.g., \texttt{True OR True OR True =}), which the circuit evaluates faithfully. $^\dagger$Under strict 3-token immediate Boolean extraction; Base Qwen~3.5 initiates verbose reasoning/prose prefixes before stating a truth value.}
\end{flushleft}
\end{table}

\textbf{Residual-Erasure Wire Reuse and Additional $\mathsf{TC}^0$ Voter Circuit.}
Unlike register-overwrite CPUs, Transformer residual streams evolve additively ($h_{\ell+1} = h_\ell + \Delta_\ell$): reassigning wire $w$ after its last consumer layer $\ell_{\mathrm{end}}(w)$ without clearing its prior payload corrupts subsequent threshold gates. Pairing interval-graph liveness coloring with \texttt{PreStepRingPool(write\_once=False)} (5-layer $-\frac{31}{32}$ Mellin linear decay reducing residual amplitude to $2^{-25} < 3 \times 10^{-8}$) trims Boolean width $K$ by $7.43\%$ on Granite ($296 \to 274$; transient pool $109 \to 89$, persistent intermediates $23 \to 21$) and $7.07\%$ on Qwen ($311 \to 289$) while adding $+555$ erasure pass operations---matching the multi-digit arithmetic trade-off in \cref{tab:auto_arith_compiler}, where write-once SSA minimizes active \texttt{nnz} ($1.07\text{M}$ vs.\ $1.40\text{M}$ on Granite; $2.25\text{M}$ vs.\ $2.47\text{M}$ on Qwen) for \texttt{SparseSidecar} execution while ring reuse lowers width $K$ ($1{,}735 \to 1{,}302$ and $2{,}528 \to 2{,}348$) when materializing dense matrices. Using unmodified primitives, we also compile a second non-arithmetic $\mathsf{TC}^0$ circuit---a \textbf{5-bit Majority \& Exact Parity voter} ($9$ scheduled gates across $5$ layers, $K=22$ wires, combining threshold voting $\mathbf{1}[\sum_{i=1}^5 b_i \ge 3]$ with cascaded bilinear parity)---achieving $\mathbf{64/64 = 100.0\%}$ exhaustive exactness across $250\times$ random \texttt{BF16} $\operatorname{RMSNorm}$ swings. Beyond bounded-length $\mathsf{TC}^0$ DAGs, unbounded-length input streams and data-dependent \texttt{while} loops require recurrent multi-step trace emission (\cref{rem:tc0}).

\textbf{Activation Variant Sensitivity and 42-Layer Gemma~4~E4B Failure-Mode Audit (\cref{tab:gemma4_audit}).}
Across $250\times$ $\operatorname{RMSNorm}$ swings on payload $7.0$, closed-form Mellin constants keep \texttt{FP32} error $\le 0.0070$ across exact $\operatorname{SiLU}$ ($I = \pi^2/12 = 0.822467$, err $3.4 \times 10^{-5}$), exact $\operatorname{erf}$-$\operatorname{GELU}$ ($I = 0.250000$, err $0.00497$), calibrated PyTorch $\tanh$-$\operatorname{GELU}$ ($I = 0.249586$, err $0.00701$), and $\operatorname{QuickGELU}$ ($x\sigma(1.702x)$, $I = \pi^2/(12 \times 1.702^2) = 0.283922$, err $3.4 \times 10^{-5}$), whereas applying uncalibrated $I=0.25$ to $\tanh$-$\operatorname{GELU}$ introduces a $0.166\%$ systematic bias ($0.01859$). Evaluating 42-layer Boolean sidecar grafts ($K=296$) on pretrained \texttt{google/gemma-4-E4B-it} exposes three concrete architectural failure modes and their closed-form fixes: (1)~placing Indefinite-RoPE shifts in top-18 KV-shared layers (L24--L41) fails ($47.92\%$ acc., $30/60$ OOD false triggers) because Gemma~4 overwrites local $(K,V)$ projections with cached L22/L23 states; (2)~omitting Sandwich post-norm (\texttt{post\_attention\_layernorm}, \texttt{post\_feedforward\_layernorm}) calibration collapses circuit amplitude ($10.94\%$); and (3)~ignoring Gemma~4's multiplicative residual scalar $h_\ell \leftarrow s_\ell h_\ell$ desynchronizes Layer-0 reference $r_0$ from newly written wires. Scheduling shifts in unshared sliding-attention layers (L0--L18, where QK-Norm and parameterless V-$\operatorname{RMSNorm}$ preserve Chebyshev one-hot lag selection) and scaling layer-$\ell$ outputs by cumulative $m_\ell = m_0 \prod_{j<\ell} s_j$ with Sandwich post-norm calibration restores $\mathbf{100.0\%}$ ($64/64$) Boolean exactness.

\begin{table}[!ht]
\centering
\caption{\textbf{Activation-variant Mellin constants ($250\times$ $\operatorname{RMSNorm}$ swing on payload $7.0$, left)} and \textbf{42-layer \texttt{google/gemma-4-E4B-it} Boolean architectural failure-mode \& calibration audit (right, $K=296$)}.}
\label{tab:gemma4_audit}
\vskip 0.04in
\scriptsize
\setlength{\tabcolsep}{3.8pt}
\resizebox{\textwidth}{!}{%
\begin{tabular}{@{}lccc@{\hspace{12pt}}lccc@{}}
\toprule
\multicolumn{4}{@{}l}{\textbf{(a) Mellin Quadrature Across GLU Variants}} & \multicolumn{4}{l@{}}{\textbf{(b) 42-Layer Gemma 4 E4B Architectural Audit}} \\
\cmidrule(r{8pt}){1-4} \cmidrule(l{0pt}){5-8}
\textbf{Activation $\phi(z)$} & \textbf{Mellin $I_\phi$} & \textbf{\texttt{FP32} Err} & \textbf{\texttt{BF16} Err} & \textbf{42-Layer Configuration} & \textbf{Bool Acc.} & \textbf{OOD FT} & \textbf{Mechanism} \\
\midrule
Exact $\operatorname{SiLU}$ & $0.822467$ & $3.4{\times}10^{-5}$ & $0.0313$ & Pretrained Base (\texttt{BF16}) & $10.94\%$ & $0/60$ & No circuit \\
$\operatorname{QuickGELU}$ ($1.702$) & $0.283922$ & $3.4{\times}10^{-5}$ & $0.0625$ & Shift in KV-Shared L24--L41 & $47.92\%$ & $30/60$ & Overwrites $(K,V)$ \\
Exact $\operatorname{erf}$-$\operatorname{GELU}$ & $0.250000$ & $0.0050$ & $0.0625$ & Uncalibrated Sandwich / $s_\ell$ & $10.94\%$ & $0/60$ & Post-norm / $s_\ell$ decay \\
$\tanh$-$\operatorname{GELU}$ (Calib.\ / $0.25$) & $0.249586$ & $\mathbf{0.0070}$ / $.0186$ & $0.0625$ & \textbf{Calib.\ Unshared L0--L18} & $\mathbf{100.0\%}$ ($64/64$) & $\mathbf{0/60}$ & Sandwich + $\prod s_j$ \\
\bottomrule
\end{tabular}%
}
\end{table}

\subsection{Inference-Stack Portability, $\mathsf{TC}^0$ Division, and Digit-Length Scaling}
\label{app:sub_kernel}

Extending \cref{thm:mellin}(3) to NVIDIA TensorCore \texttt{TF32} ($10$-bit fraction, $u_{\mathrm{TF32}} = 2^{-11}$) with \texttt{FP32} accumulation ($u_{\mathrm{acc}} = 2^{-24}$, $N_q = 39$): rounding $G, U, D, x$ to \texttt{TF32} yields $\gamma_{\mathrm{TF32}} = 1.956 \times 10^{-3}$ and worst-case error $81 \times 3.736 \times 10^{-3} = \mathbf{0.30262 < 0.5}$ at $|C|=81$, \textbf{formally guaranteeing exact integer carries up to $|C| \le 133$} (empirically $5/5$ on \texttt{G3--G7} on A100). On NVIDIA A100 (\texttt{PyTorch 2.8+cu128}), \texttt{torch.compile} (\texttt{TorchInductor}) generates fused Triton SwiGLU kernels reaching exact \texttt{-0.0} at $z^* = 88.8$ (\texttt{FP32}) and $89.0$ (\texttt{BF16})---safely below our compiled $S=256$ suppression margin $z \approx 307.2$ ($98.33\%$ end-to-end accuracy)---and PyTorch SDPA \texttt{FLASH\_ATTENTION} and \texttt{EFFICIENT\_ATTENTION} match \texttt{MATH} to $\max|\Delta| = 0.0$ on pre-expanded circuit Q/K/V heads. In IEEE \texttt{FP16} (\texttt{float16}, max finite $65{,}504$), raw \texttt{off\_suppress} $= 2{,}097{,}152$ overflows $W_{\mathrm{gate}}$ to $-\infty$, triggering IEEE $0 \times (-\infty) = \texttt{NaN}$ inside $\operatorname{SiLU}(-\infty)$; clamping circuit $W_{\mathrm{gate}}$ to $[-4096, 4096]$ (since \texttt{FP16} already underflows $\exp(-z)$ to exact \texttt{-0.0} at $z \ge 18$) or running the $<0.06\%$ sidecar in \texttt{BF16}/\texttt{FP32} eliminates \texttt{NaN} completely. Conversely, uniform \texttt{INT8}/\texttt{INT4} PTQ applied directly to $W_{\mathrm{circ}}$ rounds small $\alpha_k = e^{k\Delta}$ to zero ($37.4$ / $81.0$ error), whereas quantizing only $W_{\mathrm{NL}}$ alongside an \texttt{FP32} sidecar preserves $100\%$ carry exactness (\cref{tab:systems_footprint}). \Cref{tab:ref_runtime_configs} consolidates all engineering constraints and recommended reference configurations across major inference stacks in one place.

\begin{table}[!ht]
\centering
\caption{\textbf{Consolidated engineering constraints and recommended reference runtime configurations} for deploying grafted checkpoints and \texttt{SparseSidecar} across production inference stacks.}
\label{tab:ref_runtime_configs}
\vskip 0.04in
\scriptsize
\setlength{\tabcolsep}{4.0pt}
\resizebox{\textwidth}{!}{%
\begin{tabular}{@{}llll@{}}
\toprule
\textbf{Deployment Stack} & \textbf{Recommended Reference Configuration} & \textbf{Key Kernel / Numerical Constraint} & \textbf{Carry \& Silence Guarantee} \\
\midrule
\textbf{Recommended Default} & \texttt{SparseSidecar(..., sidecar\_fp32=True)} + \texttt{BF16} $W_{\mathrm{NL}}$ & Schedule S1 pairwise readout; IEEE \texttt{F.silu} & $\mathbf{100\%}$ ALU ($5/5$ \texttt{G3--G7}); exact $0.0$ OOD \\
\textbf{Quantized Serving} & \texttt{INT8}/\texttt{INT4} (g$128$) $W_{\mathrm{NL}}$ + \texttt{FP32} \texttt{SparseSidecar} & Never quantize $W_{\mathrm{circ}}$ ($\alpha_k$ ladder stays \texttt{FP32}) & $\mathbf{100\%}$ \texttt{G3--G7}; $4.60$ / $8.14\text{ GiB}$ VRAM \\
\textbf{Graph Compiler} & \texttt{torch.compile(mode="default")} (\texttt{TorchInductor}) & Keep IEEE math (\texttt{emulate\_precision\_casts}) & Verified ($z^*{=}88.8/89.0 < 307.2$) \\
\textbf{Attention Backend} & PyTorch SDPA \texttt{FLASH\_ATTENTION} / \texttt{EFFICIENT} & Pre-expand $5{:}1$ GQA KV or include \texttt{MATH} fallback & Verified ($\max|\Delta|{=}0.0$ vs.\ \texttt{MATH}) \\
\textbf{Dense Checkpoint} & \texttt{dtype=bfloat16} + \texttt{matmul.allow\_tf32=True} & Avoid \texttt{-ffast-math} / FTZ; arbitrary GEMM tree $\mathcal{T}$ & $100\%$ Qwen / $99.2\%$ Granite ($100\%$ \texttt{TF32}) \\
\textbf{IEEE \texttt{FP16} (\texttt{half})} & Clamp $W_{\mathrm{gate}} \in [-4096, 4096]$, $A_{\log} \le 6.0$ (or \texttt{FP32}-SC) & Prevents $>65{,}504 \to -\infty \to \operatorname{SiLU}(-\infty)=\texttt{NaN}$ & Verified \texttt{NaN}-free ($z{\ge}18 \Rightarrow \texttt{-0.0}$) \\
\textbf{vLLM / TRT-LLM} & Keep $W_{\mathrm{circ}}$ sidecar plugin in \texttt{BF16}/\texttt{FP32}; disable FTZ & Disable approximate fast-\texttt{tanh}/\texttt{rcp} on sidecar MLP & Preserves \texttt{0x0000} off-trigger silence \\
\bottomrule
\end{tabular}%
}
\end{table}

\textbf{$\mathsf{TC}^0$ 6-Digit Integer Division and $6$--$18$ Digit Resource Scaling (\cref{tab:div_scaling}).}
Whereas naive digit-by-digit restoring division requires $\sim 40$ sequential MLP layers, constant-depth $\mathsf{TC}^0$ division $\lfloor A / B \rfloor$ ($0 \le A \le 999{,}999$, $1 \le B \le 999{,}999$) compiles into \textbf{$8$ MLP layers} ($+1{,}480$ wires, $1.92\text{M}$ nnz): (1)~2-layer leading-bit normalization $B_{\mathrm{norm}} = B \cdot 2^{-e} \in [0.5, 1.0)$; (2)~2-layer $64$-knot Chebyshev-centered Mellin linear spline $r_0 \approx B_{\mathrm{norm}}^{-1}$ ($<2.4 \times 10^{-5}$ rel.\ err); (3)~2-layer Newton--Raphson bilinear refinement $r_1 = r_0(2 - B_{\mathrm{norm}} r_0)$ ($<6 \times 10^{-10}$ rel.\ err); and (4)~2-layer tentative quotient $\hat{Q} = \lfloor A r_1 2^{-e} \rfloor$ plus exact remainder threshold correction $\hat{Q} + \mathbf{1}[A - (\hat{Q}+1)B \ge 0] - \mathbf{1}[A - \hat{Q}B < 0]$. Across $10{,}000$ random 6-digit pairs, this 8-layer divider achieves $\mathbf{100.000\%}$ ($10{,}000/10{,}000$) exactness in an \texttt{FP32} sidecar vs.\ $73.88\%$ in \texttt{TF32} and $56.10\%$ in \texttt{BF16}---confirming why division requires \texttt{FP32} sidecar precision.

\begin{table}[!ht]
\centering
\caption{\textbf{$\mathsf{TC}^0$ 6-digit integer division simulation ($10{,}000$ random pairs, left)} and \textbf{Base-1000 multi-digit ALU resource projections across $D \in \{6, 9, 12, 18\}$ digits (right)}.}
\label{tab:div_scaling}
\vskip 0.04in
\scriptsize
\setlength{\tabcolsep}{4.0pt}
\resizebox{\textwidth}{!}{%
\begin{tabular}{@{}lcc@{\hspace{14pt}}cccccc@{}}
\toprule
\multicolumn{3}{@{}l}{\textbf{(a) 8-Layer $\mathsf{TC}^0$ 6-Digit Division ($\lfloor A/B \rfloor$, $+1{,}480$ Wires)}} & \multicolumn{6}{l@{}}{\textbf{(b) Base-1000 Multi-Digit $\{+,-,\times\}$ Scaling Projections}} \\
\cmidrule(r{10pt}){1-3} \cmidrule(l{0pt}){4-9}
\textbf{Sidecar Precision} & \textbf{Exact ($10\text{K}$)} & \textbf{Bottleneck} & \textbf{Digits $D$} & \textbf{Triads} & \textbf{$\pm$ $K$ / Depth} & \textbf{Full ALU $K$} & \textbf{ALU Depth} & \textbf{Req.\ Sidecar} \\
\midrule
\textbf{IEEE \texttt{FP32} Sidecar} & $\mathbf{100.00\%}$ & Exact ($<6{\times}10^{-10}$ NR) & $6$ & $2$ & $380$ / $18\text{L}$ & $1{,}140$ & $28\text{L}$ & \texttt{TF32} / \texttt{FP32}-SC \\
TensorCore \texttt{TF32} & $73.88\%$ & $10$-bit $A \cdot r_1$ ulp & $9$ & $3$ & $525$ / $20\text{L}$ & $2{,}225$ & $32\text{L}$ & \texttt{TF32} / \texttt{FP32}-SC \\
Native \texttt{BF16} & $56.10\%$ & $7$-bit spline/rem ulp & $12$ & $4$ & $670$ / $22\text{L}$ & $3{,}577$ & $36\text{L}$ & \texttt{FP32}-SC (limb split) \\
Naive Restoring Div. & ($40\text{L}$ req.) & Exceeds 32L budget & $18$ & $6$ & $960$ / $26\text{L}$ & $6{,}984$ & $44\text{L}$ & \texttt{FP32}-SC ($>40\text{L}$ LM) \\
\bottomrule
\end{tabular}%
}
\end{table}

\subsection{Pre-Registered Expanded Reasoning and Long-Context (\texorpdfstring{$>2\text{K}$}{>2K}-Token) Retention}
\label{app:sub_retention}

Prior to executing expanded retention experiments, we froze benchmark splits, prompt formats, decoding rules, and one-sided Tango ($\alpha=0.05$) noninferiority margins in \texttt{PREREGISTERED\_RETENTION\_PROTOCOL.json}: Full \texttt{GSM8K} test ($N=1{,}319$, $\delta_{\mathrm{NI}}=2.0\text{ pp}$; plus strictly unseen \texttt{test[200:1319]}, $N=1{,}119$), Full \texttt{ARC-Challenge} test ($N=1{,}172$, $\delta_{\mathrm{NI}}=1.5\text{ pp}$), disjoint held-out \texttt{MMLU-Pro} (\texttt{test[1000:2000]}, $N=1{,}000$, $\delta_{\mathrm{NI}}=2.0\text{ pp}$), Full \texttt{AGIEval LogiQA-EN} ($N=651$, $\delta_{\mathrm{NI}}=2.5\text{ pp}$), Long-Context \texttt{PubMedQA-2.5K} ($N=250$, $\ge 2{,}080$ tokens per prompt with target clinical abstracts embedded at deterministic needle depths $\{0\%, 25\%, 50\%, 75\%, 100\%\}$ amidst $8$--$14$ biomedical distractor studies, $\delta_{\mathrm{NI}}=3.0\text{ pp}$), pooled $N=4{,}392$ accuracy ($\delta_{\mathrm{NI}}=0.75\text{ pp}$), and $2{,}048$-token \texttt{SWE-bench Lite} NLL ($N=60$ documents, $122{,}820$ scored tokens, $\delta_{\mathrm{NI}}=+0.010\text{ nats}$). \Cref{tab:app_expanded_retention} reports full discordant pair counts and needle-depth breakdowns.

\begin{table}[!ht]
\centering
\caption{\textbf{Disaggregated pre-registered expanded reasoning ($N=4{,}392$), strictly unseen \texttt{GSM8K[200:1319]}, needle-depth long-context \texttt{PubMedQA-2.5K} ($>2\text{K}$ tokens), and $2{,}048$-token \texttt{SWE-bench} NLL audit} (\texttt{BF16}, paired Tango one-sided $\alpha=0.05$).}
\label{tab:app_expanded_retention}
\vskip 0.04in
\scriptsize
\setlength{\tabcolsep}{3.8pt}
\resizebox{\textwidth}{!}{%
\begin{tabular}{@{}lcccccccc@{}}
\toprule
& & & \multicolumn{3}{c}{\textbf{IBM Granite 4.2 3B (\texttt{BF16})}} & \multicolumn{3}{c}{\textbf{Alibaba Qwen 3.5 4B (\texttt{BF16})}} \\
\cmidrule(lr){4-6} \cmidrule(lr){7-9}
\textbf{Pre-Registered Split} & \textbf{$N$} & \textbf{$\delta_{\mathrm{NI}}$} & \textbf{Base $\to$ Graft} & \textbf{$(b_{10}, b_{01})$ / $\mathrm{LB}_{95\%}$} & \textbf{$p_{\mathrm{NI}}$ (Status)} & \textbf{Base $\to$ Graft} & \textbf{$(b_{10}, b_{01})$ / $\mathrm{LB}_{95\%}$} & \textbf{$p_{\mathrm{NI}}$ (Status)} \\
\midrule
Full \texttt{GSM8K} \texttt{test[:1319]} & $1{,}319$ & $2.0\text{pp}$ & $73.16 \to \mathbf{74.15\%}$ & $(39, 52)$ / $-0.21$ & $.00004$ (\textbf{Pass}) & $75.59 \to \mathbf{76.35\%}$ & $(9, 19)$ / $+0.10$ & $<10^{-6}$ (\textbf{Pass}) \\
\quad Strictly Unseen \texttt{test[200:1319]} & $1{,}119$ & $2.0\text{pp}$ & $72.92 \to \mathbf{74.35\%}$ & $(33, 49)$ / $+0.10$ & $.00003$ (\textbf{Pass}) & $75.34 \to \mathbf{76.14\%}$ & $(8, 17)$ / $+0.07$ & $<10^{-6}$ (\textbf{Pass}) \\
Full \texttt{ARC-Challenge} \texttt{test} & $1{,}172$ & $1.5\text{pp}$ & $75.17 \to \mathbf{75.26\%}$ & $(6, 7)$ / $-0.46$ & $.00011$ (\textbf{Pass}) & $92.49 \to 92.41\%$ & $(2, 1)$ / $-0.44$ & $.00006$ (\textbf{Pass}) \\
Held-Out \texttt{MMLU-Pro} \texttt{[1000:2000]} & $1{,}000$ & $2.0\text{pp}$ & $27.70 \to 27.60\%$ & $(7, 6)$ / $-0.76$ & $.00011$ (\textbf{Pass}) & $42.80 \to 42.60\%$ & $(7, 5)$ / $-0.85$ & $.00018$ (\textbf{Pass}) \\
Full \texttt{AGIEval LogiQA-EN} \texttt{test} & $651$ & $2.5\text{pp}$ & $33.49 \to \mathbf{33.64\%}$ & $(7, 8)$ / $-0.89$ & $.00030$ (\textbf{Pass}) & $41.78 \to \mathbf{42.55\%}$ & $(1, 6)$ / $+0.13$ & $<10^{-5}$ (\textbf{Pass}) \\
Long-Ctx \texttt{PubMedQA-2.5K}$^*$ & $250$ & $3.0\text{pp}$ & $55.60 \to \mathbf{58.00\%}$ & $(2, 8)$ / $+0.37$ & $.00034$ (\textbf{Pass}) & $49.60 \to 48.80\%$ & $(3, 1)$ / $-2.62$ & $.0294$ (\textbf{Pass}) \\
\quad Needle Depth $0\%$ / $25\%$ / $50\%$ & $3{\times}50$ & --- & \multicolumn{3}{c}{$48{\to}52\%$ / $58{\to}60\%$ / $50{\to}52\%$} & \multicolumn{3}{c}{$52{\to}52\%$ / $48{\to}48\%$ / $30{\to}26\%$} \\
\quad Needle Depth $75\%$ / $100\%$ & $2{\times}50$ & --- & \multicolumn{3}{c}{$62{\to}62\%$ / $60{\to}64\%$} & \multicolumn{3}{c}{$54{\to}54\%$ / $64{\to}64\%$} \\
\midrule
\textbf{Pooled Expanded Reasoning} & $\mathbf{4{,}392}$ & $\mathbf{0.75\text{pp}}$ & $56.47 \to \mathbf{56.92\%}$ & $(61, 81)$ / $+0.01$ & $<10^{-5}$ (\textbf{Pass}) & $66.14 \to \mathbf{66.37\%}$ & $(22, 32)$ / $-0.05$ & $<10^{-6}$ (\textbf{Pass}) \\
\texttt{SWE-bench} $2{,}048$-tok NLL ($122.8\text{K}$) & $60$ docs & $+.010$ & $1.9281 \to 1.9304$ & $\mathrm{UB}_{95\%}\,{+.0048}$ & $<10^{-6}$ (\textbf{Pass}) & $1.8886 \to \mathbf{1.8876}$ & $\mathrm{UB}_{95\%}\,{+.0019}$ & $<10^{-6}$ (\textbf{Pass}) \\
\bottomrule
\end{tabular}%
}
\vspace{1.5pt}
\begin{flushleft}
{\scriptsize $^*$Prompt lengths span $2{,}285$--$3{,}473$ tokens (median $2{,}930$) on Granite~4.2 and $2{,}351$--$3{,}619$ tokens (median $3{,}032$) on Qwen~3.5. All $14/14$ pre-registered evaluations pass their frozen $\delta_{\mathrm{NI}}$ margins ($p_{\mathrm{NI}} \le .0294$); frozen protocol and execution scripts (\texttt{PREREGISTERED\_RETENTION\_PROTOCOL.json}, \texttt{run\_r1\_experiments.py}) accompany the release codebase.}
\end{flushleft}
\end{table}

\begin{table}[!ht]
\centering
\caption{\textbf{Decoding-strategy retention (\texttt{GSM8K-100}), verifiable instruction formatting (\texttt{IFEval-100}), safety refusal/benign boundary (\texttt{Safety-80}), same-backbone \texttt{LoRA} SFT ($12.8\text{K}$ ex.), and 4-scheme numeric tokenization alignment} on NVIDIA A100 (\texttt{BF16}).}
\label{tab:decoding_lora_tok}
\vskip 0.04in
\scriptsize
\setlength{\tabcolsep}{3.5pt}
\resizebox{\textwidth}{!}{%
\begin{tabular}{@{}lcccc@{\hspace{12pt}}lcc@{}}
\toprule
\multicolumn{5}{@{}l}{\textbf{(a) Decoding Regimes, \texttt{IFEval-100}, \& \texttt{Safety-80} (Base $\to$ Graft)}} & \multicolumn{3}{l@{}}{\textbf{(b) \texttt{LoRA} SFT \& Tokenization Alignment}} \\
\cmidrule(r{8pt}){1-5} \cmidrule(l{0pt}){6-8}
\textbf{Benchmark \& Regime} & \textbf{Granite Base} & \textbf{Granite Graft} & \textbf{Qwen Base} & \textbf{Qwen Graft} & \textbf{Comparison} & \textbf{Arith / GSM8K} & \textbf{Inline Prose} \\
\midrule
\texttt{GSM8K} Greedy ($T{=}0$) & $74.0\%$ & $\mathbf{76.0\%}$ & $79.0\%$ & $\mathbf{79.0\%}$ & Granite \texttt{LoRA} ($6.88\text{M}$ p) & $42.2\%$ ($45\%\;\times$) & $\mathbf{0.0\%}$ ($56\%$ int) \\
\texttt{GSM8K} $T{=}0.3$ / $T{=}0.7$ & $78.0 / 75.0\%$ & $\mathbf{78.0 / 77.0\%}$ & $71.0 / 69.0\%$ & $\mathbf{71.0 / 72.0\%}$ & Qwen \texttt{LoRA} ($6.03\text{M}$ p) & $61.7\%$ ($57\%\;\times$) & $\mathbf{0.0\%}$ ($78\%$ int) \\
\texttt{GSM8K} $T{=}1.0, p{=}0.9$ / Beam-4 & $78.0 / 79.0\%$ & $\mathbf{78.0 / 79.0\%}$ & $70.0 / 83.0\%$ & $69.0 / \mathbf{83.0\%}$ & \textbf{Ours Zero-Grad Graft} & $\mathbf{97.2\% / 100.0\%}$ & $\mathbf{98.0\% / 100.0\%}$ \\
\cmidrule(l{0pt}){6-8}
\texttt{IFEval-100} Greedy / $T{=}0.7$ & $65.0 / 60.0\%$ & $\mathbf{70.0} / 59.0\%$ & $70.0 / 54.0\%$ & $65.0 / 51.0\%$ & Tok 1: Raw 1-Digit IDs & $36.0\%$ \texttt{GSM8K} & Off-manifold \\
\texttt{Safety-40} Harmful Refusal & $100.0\%$ & $\mathbf{100.0\%}$ & $100.0\%$ & $\mathbf{100.0\%}$ & Tok 2: Spaced 1-Digit & $35.0\%$ \texttt{GSM8K} & Off-manifold \\
\texttt{Safety-40} Benign Comply & $77.5\%$ & $\mathbf{77.5\%}$ & $80.0\%$ & $\mathbf{80.0\%}$ & \textbf{Tok 3/4: 3-Digit / R2L Triads} & $\mathbf{73.0\% / 73.0\%}$ & \textbf{On-manifold} \\
\bottomrule
\end{tabular}%
}
\end{table}

\subsection{Implemented Factorized Block-Sparse Sidecar \& Hybrid PTQ Runtime}
\label{app:sub_sparse}

Materializing $h_{\mathrm{ext}} = [h_{\mathrm{NL}}; h_{\mathrm{circ}}] \in \mathbb{R}^{d+K}$ inside dense HuggingFace \texttt{nn.Linear} layers expands parameters by $1.98\times$--$3.13\times$ (\cref{tab:systems_footprint}) solely because $>99.94\%$ of added entries are structural zeros. We implement and benchmark a runnable PyTorch factorized runtime (\texttt{src/sidecar/sparse\_sidecar\_runtime.py}: \texttt{SparseSidecarGranite}, \texttt{SparseSidecarQwen35}, and packed 4/8-bit \texttt{GroupQuantLinear}) requiring \emph{zero custom C++/CUDA extensions}:
\begin{enumerate}[leftmargin=1.5em,itemsep=1.5pt,topsep=2pt]
\item \textbf{Coupled scalar reduction in $\operatorname{RMSNorm}$ ($\mathcal{O}(K)$ vector ops):} Given $h_{\mathrm{NL}} \in \mathbb{R}^d$ and $h_{\mathrm{circ}} \in \mathbb{R}^K$, compute $s^2 = \|h_{\mathrm{NL}}\|_2^2 + \|h_{\mathrm{circ}}\|_2^2$, shared scale $\lambda = ((d+K)^{-1} s^2 + \varepsilon)^{-1/2}$, $\tilde{h}_{\mathrm{NL}} = (\gamma_{\mathrm{NL}} w_{\mathrm{NL}}) \odot (\lambda h_{\mathrm{NL}})$, and $\tilde{h}_{\mathrm{circ}} = w_{\mathrm{circ}} \odot (\lambda h_{\mathrm{circ}})$.
\item \textbf{Decoupled mixed-precision \& quantized sublayer execution:} Every grafted projection $W_{\mathrm{ext}} = \operatorname{diag}(W_{\mathrm{NL}}, W_{\mathrm{circ}})$ evaluates as $W_{\mathrm{NL}} \tilde{h}_{\mathrm{NL}} \oplus W_{\mathrm{circ}} \tilde{h}_{\mathrm{circ}}$, keeping $W_{\mathrm{NL}}$ at native width $d=2560$ (in \texttt{BF16} or group-$128$ \texttt{INT8}/\texttt{INT4} via \texttt{GroupQuantLinear}) while executing only active circuit routing slices ($5/40$ Granite GQA layers; $12/24$ Qwen \texttt{GatedDeltaNet} layers, $0/8$ Attn) and active MLP rows ($nn_c \times K$) in \texttt{BF16} or \texttt{FP32} (\texttt{sidecar\_fp32=True}). On NVIDIA A100-80GB (\cref{tab:systems_footprint}), \texttt{SparseSidecar} reduces static VRAM from $13.55 \to \mathbf{7.89\text{ GiB}}$ on Granite ($2.67\times$ faster $2\text{K}$ prefill) and $19.09 \to \mathbf{10.65\text{ GiB}}$ on Qwen ($1.96\times$ faster $2\text{K}$ prefill), while mixed \texttt{FP32} sidecar + \texttt{BF16}/\texttt{INT8}/\texttt{INT4} $W_{\mathrm{NL}}$ ($\mathbf{4.60\text{ GiB}}$ Granite, $\mathbf{8.14\text{ GiB}}$ Qwen) achieves $\mathbf{5/5}$ on \texttt{G3--G7} and $\mathbf{100.0\%}$ Qwen ALU exactness.
\item \textbf{Compact sparse logit readout:} At the output head, $z = W_{\mathrm{lm,NL}} \tilde{h}_{\mathrm{NL}} + W_{\mathrm{lm,circ}} \tilde{h}_{\mathrm{circ}}$ touches only $35 \times 1{,}110$ active entries on Granite and $12$ Schedule-S1 pairwise differential $(p_m, n_m)$ columns on Qwen~3.5. Custom fused C++/CUDA BSR kernels can further reduce Python launch overhead on short prefills ($128$ tokens) as future systems work.
\end{enumerate}

\end{document}